\documentclass[10pt,twocolumn,letterpaper]{article}

\usepackage[pagenumbers]{cvpr} 

\usepackage{cuted} 

\usepackage{currfile} 

\usepackage{caption} 

\usepackage{booktabs}
\usepackage{multirow}
\usepackage{xspace}
\usepackage{amssymb}

\usepackage{tcolorbox}
\tcbuselibrary{breakable}

\newcommand{\ours}{EEG-Video-R1\xspace} 
\newcommand{\base}{\model{Qwen3-VL-4B}\xspace}

\usepackage{placeins}

\usepackage{tikz}
\usetikzlibrary{arrows.meta,positioning,calc,fit,backgrounds}
\usepackage{amsmath}

\DeclareMathOperator{\BA}{BA}

\newcounter{finding}
\definecolor{findingBg}{HTML}{F7F4FC}
\definecolor{findingBorder}{HTML}{D8CBEA}
\definecolor{findingText}{HTML}{66508A}

\newcommand{\finding}[1]{%
    \refstepcounter{finding}%
    \begin{tcolorbox}[
        colback=findingBg,
        colframe=findingBorder,
        coltext=black,
        boxrule=0.5pt,
        arc=2mm,
        left=6pt,
        right=6pt,
        top=2pt,
        bottom=2pt,
        before skip=6pt,
        after skip=6pt,
        parbox=false
    ]%
    \textcolor{findingText}{\textbf{Finding~\thefinding.}}\enspace#1%
    \end{tcolorbox}%
}

\definecolor{baseRow}{HTML}{FFF4D6}   
\definecolor{meanCol}{HTML}{EAF1FB}   
\definecolor{gainTxt}{HTML}{1B7F3B}   
\newcommand{\bR}{\cellcolor{baseRow}}
\newcommand{\gup}[1]{\,{\tiny\textcolor{gainTxt}{+#1}}}

\usepackage{makecell}

\usepackage{pifont}
\definecolor{cmarkgreen}{HTML}{1A9E5C}
\definecolor{xmarkred}{HTML}{D9534F}
\newcommand{\cmark}{\textcolor{cmarkgreen}{\ding{51}}}
\newcommand{\xmark}{\textcolor{xmarkred}{\scalebox{0.85}{\ding{55}}}}

\usepackage{multirow,graphicx,tabularx,booktabs}
\usepackage[table]{xcolor}
\newcolumntype{Y}{>{\centering\arraybackslash}X}
\definecolor{mutedtxt}{gray}{0.5}
\definecolor{rulegray}{gray}{0.72}
\newcommand{\inp}[1]{\textcolor{mutedtxt}{\footnotesize #1}}
\newcommand{\lightrule}{\arrayrulecolor{rulegray}\specialrule{0.3pt}{1.5pt}{1.5pt}\arrayrulecolor{black}}

\usepackage{multirow,graphicx,tabularx,booktabs}
\usepackage[table]{xcolor}
\newcolumntype{Y}{>{\centering\arraybackslash}X}
\definecolor{gainL}{HTML}{E3F4E6}   
\definecolor{gainM}{HTML}{BFE5C7}   
\definecolor{gainH}{HTML}{8FD3A0}   
\definecolor{mutedtxt}{gray}{0.35}
\definecolor{rulegray}{gray}{0.55}
\newcommand{\best}[1]{\textbf{#1}}

\newcommand{\grp}[3]{\multirow{#1}{*}{\rotatebox[origin=c]{90}{\fontsize{6}{6.5}\selectfont\textcolor{mutedtxt}{\shortstack{#2\\#3}}}}}
\newcommand{\mut}[1]{\textcolor{mutedtxt}{#1}}

\definecolor{cvprblue}{rgb}{0.21,0.49,0.74}
\usepackage[pagebackref,breaklinks,colorlinks,allcolors=cvprblue]{hyperref}

\def\confName{CVPR}
\def\confYear{2027}

\title{\ours: Do Spatio-Temporal Grounding and Reasoning in Video-LLMs Transfer to Brain Signals?}

\author{
Negin Baghbanzadeh$^{1,2,3}$ \quad
Pritam Sarkar$^{2,4}$ \quad
Anis Yazidi$^{6,5}$ \quad
Ali Ramezani-Kebrya$^{5}$ \\
Leonid Sigal$^{2,4}$ \quad
Ali Etemad$^{3,7}$ \quad
Elham Dolatabadi$^{1,2,3}$ \\[1ex]
$^{1}$York University \quad
$^{2}$Vector Institute \quad
$^{3}$Connected Minds \quad
$^{4}$University of British Columbia \\
$^{5}$University of Oslo \quad
$^{6}$Simula Research Laboratory \quad
$^{7}$Queen's University
}

\usepackage{booktabs}
\usepackage{amssymb}
\usepackage{xcolor}
\usepackage{colortbl}

\usepackage{booktabs}
\usepackage{xcolor}
\usepackage{tcolorbox}
\usepackage{tabularx}

\usepackage{pifont}
\definecolor{whenfg}{HTML}{D9731A}\definecolor{whenbg}{HTML}{FCEBDA}
\definecolor{wherefg}{HTML}{1B5E20}
\definecolor{wherebg}{HTML}{DEEADF}
\newcommand{\when}[1]{{\setlength{\fboxsep}{1pt}\colorbox{whenbg}{\textcolor{whenfg}{\vphantom{Ag}#1}}}}
\newcommand{\where}[1]{{\setlength{\fboxsep}{1pt}\colorbox{wherebg}{\textcolor{wherefg}{\vphantom{Ag}#1}}}}
\newcommand{\obs}[1]{\ding{\numexpr201+#1\relax}}

\usepackage[normalem]{ulem}

\usepackage{graphicx}
\usepackage{capt-of}

\definecolor{oursPurple}{HTML}{6246A8}
\definecolor{oursLight}{HTML}{F1EDFA}

\usepackage{tabularx}
\newcolumntype{Y}{>{\centering\arraybackslash}X}

\usepackage{enumitem}
\usepackage{wrapfig}

\newcommand{\model}[1]{{\ttfamily\hyphenchar\font=`\-\relax #1}}

\usepackage{booktabs,tabularx}
\newcolumntype{W}[1]{>{\hsize=#1\hsize\centering\arraybackslash}X}

\providecommand{\tblPending}{\textcolor{gray}{\textit{TBD}}}

\newtcolorbox{promptbox}[1][]{%
    colback=gray!8,             
    colframe=gray!60,           
    arc=2mm,                    
    boxrule=0.7pt,
    left=8pt, right=8pt,
    top=8pt, bottom=8pt,
    fonttitle=\bfseries\sffamily,
    fontupper=\ttfamily\small,  
    title=#1,
    breakable
}

\usepackage{tikz}
\usetikzlibrary{patterns}
\definecolor{swsolid}{HTML}{9A9A9A}
\definecolor{swlight}{HTML}{E1E1E1}
\definecolor{swmid}{HTML}{C7C7C7}
\definecolor{swedge}{HTML}{646464}
\newsavebox{\swA}\newsavebox{\swB}\newsavebox{\swC}\newsavebox{\swD}
\AtBeginDocument{%
  \sbox{\swA}{\tikz[baseline=0.1ex]{\fill[swsolid] (0,0) rectangle (1.6ex,1.3ex);}}%
  \sbox{\swB}{\tikz[baseline=0.1ex]{\fill[swlight] (0,0) rectangle (1.6ex,1.3ex);
      \draw[swedge, line width=0.3pt, pattern=north east lines, pattern color=swedge] (0,0) rectangle (1.6ex,1.3ex);}}%
  \sbox{\swC}{\tikz[baseline=0.1ex]{\fill[swmid] (0,0) rectangle (1.6ex,1.3ex);
      \draw[swedge, line width=0.3pt, pattern=dots, pattern color=swedge] (0,0) rectangle (1.6ex,1.3ex);}}%
  \sbox{\swD}{\tikz[baseline=0.1ex]{\draw[black, line width=0.6pt] (0,-0.1ex) -- (0,1.4ex);}}%
}
\DeclareRobustCommand{\swsolidbar}{\usebox{\swA}}
\DeclareRobustCommand{\swhatchbar}{\usebox{\swB}}
\DeclareRobustCommand{\swdotbar}{\usebox{\swC}}
\DeclareRobustCommand{\swtick}{\usebox{\swD}}

\definecolor{bxseizure}{HTML}{6246A8}
\definecolor{bxleft}{HTML}{E84294}
\definecolor{bxright}{HTML}{128F8C}
\DeclareRobustCommand{\inbox}[2]{{\setlength{\fboxsep}{1.2pt}\setlength{\fboxrule}{0.9pt}\fcolorbox{#1}{white}{\vphantom{Ag}#2}}}

\begin{document}
\maketitle
\begin{abstract}
Video large language models (Video-LLMs) excel at reasoning and spatio-temporal grounding in videos, but it is unclear whether these
capabilities extend to brain signals for which no reasoning or
grounding annotations exist.
We study this question for electroencephalography (EEG), a long multichannel recording of brain activity. We render each recording as a video that a pre-trained Video-LLM can process through its existing visual pathway.
We adapt the model through mid-training, reasoning self-training, and
reinforcement learning, without reasoning or grounding annotations.
We introduce \ours a Video-LLM adapted to EEG, trained using 15K EEG recordings which achieves 73.2\% accuracy across seven EEG benchmarks, exceeding the strongest baseline by 6.5 percentage points.
We assess grounded question answering, spatio-temporal localization, and event spotting using expert localization annotations reserved for evaluation.
Grounded question answering accuracy improves from 16.4\% for the
pre-trained Video-LLM to 67.3\% after adaptation, above the best
baseline (58.8\%).
Temporal masking and spatial mirroring further support the model’s reliance on relevant signal evidence. We release the
code and dataset. 
\end{abstract}

\section{Introduction}
\label{sec:intro}



\begin{figure*}[t]
    \centering
    \includegraphics[width=0.99\textwidth]{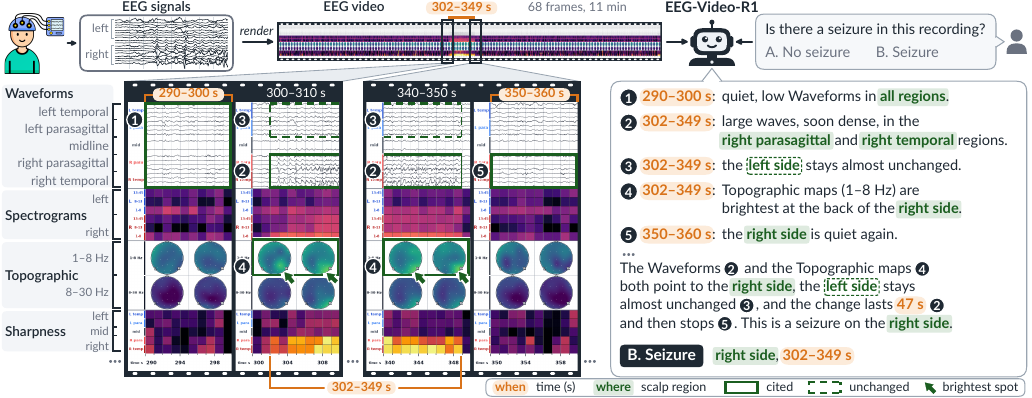}
    \caption{
    \textbf{\ours{} reads EEG as video and reasons about when and where
    events occur.} We render an EEG recording as a video in which each
    frame shows one 10-s window. The figure zooms into two pairs of
    consecutive frames, around the seizure onset (290--310\,s) and its
    end (340--360\,s). Each frame has four panels. \emph{Waveforms} show
    the signals grouped by scalp region, \emph{Spectrograms} show power
    in each frequency band over time for the left and right hemispheres,
    \emph{Topographic} maps show where power lies on the scalp, and
    \emph{Sharpness} highlights abrupt changes. Given the video and a
    question, \ours{} reasons before it answers. Each observation
    states \when{when} it occurs, as an interval in seconds (temporal
    grounding), and \where{where}, as a scalp region (spatial
    grounding). The markers \obs{1}--\obs{5} link each observation to
    its evidence in the frames, where solid boxes mark cited regions and
    dashed boxes mark regions cited as unchanged. Also \ours{} concludes
    that the recording contains a seizure on the \where{right side}
    from \when{302 to 349\,s}.
}
    \label{fig:pipeline}
\end{figure*}

Recent advances in video-large language models (Video-LLMs) have expanded video understanding beyond event recognition to reasoning about \textit{when} and \textit{where} events occur, demonstrating strong performance in long-video understanding and spatio-temporal video grounding~\cite{zhang2020does, bai2025qwen3,ren2024timechat,huang2024vtimellm,he2024ma,li2025llava,munasinghe2025videoglamm,sarkar2026self}.
These capabilities, however, rely heavily on large-scale video-text pretraining, which is difficult to replicate in specialized domains with limited paired annotations.
A representative example is electroencephalography (EEG), which records the brain’s electrical activity as a multichannel time series from electrodes placed across the scalp. Like video, EEG contains spatial and temporal structure: multichannel recordings capture evolving brain activity across scalp regions, and interpretation requires identifying brief events and their spatial distribution within recordings spanning minutes to hours. 
Unlike video, EEG data cannot be collected at web scale, and producing paired text requires clinical expertise, making large-scale annotation costly and difficult to scale. Despite these structural similarities, whether Video-LLMs can transfer their spatio-temporal grounding and reasoning capabilities to non-visual signals such as EEG with limited training data and no explicit grounding supervision remains an open question.

EEG foundation models such as BIOT~\cite{yang2023biot}, LaBraM~\cite{jiang2024large}, EEGPT~\cite{wang2024eegpt}, CBraMod~\cite{wang2025cbramod}, REVE~\cite{el2026reve}, and CodeBrain~\cite{ma2026codebrain} have advanced self-supervised EEG representation learning, primarily through temporal patch tokenization and masked modeling objectives. While effective for downstream classification through linear probing or fine-tuning, these models do not explicitly reason about the spatio-temporal evidence underlying their predictions, with interpretability largely relying on post-hoc attribution methods~\cite{selvaraju2017grad,abnar2020quantifying}. Moving beyond classification requires identifying whether a recording contains a seizure, when it begins and ends, and which scalp regions exhibit the relevant activity. Such evidence-grounded interpretation could also support more interpretable neural decoding in emerging brain–computer interfaces (BCI).

Recent EEG–language models, \cite{gu2026cerebragloss,wang2026neuronarrator,jin2026eeg,jiang2025neurolm, zeng2025wavemind,zhao2025eegagent,pradeepkumar2026neural} have begun to address some of these limitations by 
expanding to
instruction following and natural-language interpretation of EEG recordings. However, these approaches are constrained by short recording durations \cite{gu2026cerebragloss,wang2026neuronarrator,jin2026eeg}, reliance on external signal-analysis tools \cite{zhao2025eegagent}, or coarse recording-level supervision \cite{pradeepkumar2026neural}. Reasoning over long EEG recordings with fine-grained spatio-temporal grounding, therefore, remains largely unexplored. We provide an extended discussion of related work
in Appendix~A.
In this work, we investigate whether a pre-trained Video-LLM can serve as a foundation for spatio-temporal grounding and reasoning in EEG, a non-visual modality recorded as a long multichannel 1D time series. 
To begin with, we render each EEG recording as a video in which each frame shows one 10-second window through four views that expose the shape, frequency content, scalp location, and abrupt events of the signal (Fig.~\ref{fig:pipeline}).
Building on this visual rendering of EEG, we develop \ours, a Video-LLM adapted for EEG-related tasks through weak supervision.

Training proceeds in two stages. In mid-training, we train the visual
projector to answer questions about the rendered EEG. In post-training,
the model is fine-tuned on its own reasoning traces that reach the
correct answer, and then optimized with reinforcement learning (RL),
which rewards correct final answers.
Neither stage requires externally annotated reasoning traces or
grounding labels.
We evaluate \ours across seven benchmarks. We assess its spatio-temporal grounding through evidence-based question answering, seizure onset detection and event localization, and identification of affected scalp regions. Further, we examine its ability to pinpoint precise event boundaries in hour-long EEG recordings even beyond our training distribution and verify whether its predictions rely on the underlying EEG evidence by testing how they change when relevant temporal segments are masked or spatial patterns are mirrored.

\noindent In summary, our contributions are as follows.
\begin{enumerate}[label=\textbf{\arabic*.}, leftmargin=1.4em, itemsep=2pt, topsep=3pt, parsep=0pt]

    \item We introduce \ours and show that \textbf{Video-LLMs can serve
    as a foundation for spatio-temporal reasoning over non-visual
    signals} such as EEG, rendered as video.

    \item \ours achieves \textbf{73.2\%} mean balanced accuracy across
    seven EEG evaluations, outperforming the strongest task-specific EEG
    encoder by \textbf{6.5 points}, and surpasses all baselines in
    grounded question answering accuracy and spatio-temporal grounding. It also outperforms the best prior model by \textbf{19.4 points} on
external data unseen in training, and generalizes to recordings longer
than those seen in training.

    \item We conduct an \textbf{in-depth analysis of what enables
    grounded reasoning over EEG}, examining the effects of reasoning
    supervision, visual rendering, recording length, and input modality.
\end{enumerate}

\section{\ours}
\label{sec:method}



\subsection{Overview}
We consider an off-the-shelf Video-LLM $\pi_\theta$ equipped with video reasoning and spatio-temporal grounding capabilities. Our goal is to adapt $\pi_\theta$ to to non-visual signals, such as EEG, without explicit domain-specific reasoning or grounding supervision, by rendering them as videos processed through the model's existing visual pathway. We first mid-train $\pi_\theta$ on answer labels, which teaches it to read the signal while preserving its ability to reason. We then post-train it through supervised fine tuning (SFT) and further optimizing it with RL. Further details on rendering, training, and inference are provided in Appendix~B.


\subsection{EEG as Video} 
\label{sec:eeg_video}
Given an EEG recording $e\in\mathbb{R}^{C_e\times T}$, a question $q$, and its corresponding answer $a$, where $C_e$ denotes the number of EEG channels and $T$ denotes the number of time steps, we partition $e$ into non-overlapping windows of duration $\Delta=10$ seconds. Each window $e_{l\Delta:(l+1)\Delta}$ is rendered as a video frame $v_l$, preserving the temporal order of the original recording. This yields a video sequence $v\in\mathbb{R}^{C_v\times H\times W\times L}$, where $C_v$ denotes the number of visual channels (e.g., RGB), $H$ and $W$ denote the frame height and width, respectively, and $L=\lfloor T/\Delta\rfloor$ is the total number of frames. To adapt a video-LLM to the rich and complex structure of EEG recordings, we compose each frame $v_l$ from four complementary visualizations, each exposing a distinct aspect of neural activity. Together, they preserve the temporal dynamics, frequency characteristics, spatial distributions, and transient events in EEG, providing a more comprehensive visual account of the underlying brain activity (Fig.~\ref{fig:pipeline}):

(\textit{a}) \textbf{Waveforms} depict the raw EEG signals from individual channels in pixel space, preserving fine-grained signal morphology and channel-wise temporal dynamics. The waveforms are organized into five groups corresponding to distinct scalp regions: left temporal, right temporal, left parasagittal, right parasagittal, and midline.

(\textit{b}) \textbf{Spectrograms} visualize signal power as heatmaps over time and frequency, with one row per frequency band (labeled in Hz) for each hemisphere. They expose frequency-specific activity and its temporal evolution.

(\textit{c}) \textbf{Topographic} maps show scalp top views colored by signal power, with one row per frequency band and one column per 5-second half-window. They reveal the spatial distribution of neural activity across the scalp, highlighting regional concentration and hemispheric asymmetries.

(\textit{d}) \textbf{Sharpness} maps visualize abrupt signal changes as heatmaps, with one row per scalp region. They highlight temporally localized transients, making brief events such as epileptic spikes more salient.

Although Video-LLMs may not be familiar with EEG-specific visualizations, their exposure to diverse spatial structures and ability to model visual dependencies across time provid  e a natural starting point for interpreting visually rendered EEG recordings~\cite{lan2026gem,jin2026ecg,zhao2025mmvu,he2024mmworld,li2024synthesize,lee2023pix2struct}.

\subsection{Adaptation of Video-LLMs for Spatio-temporal Reasoning}
\label{sec:training}

We adopt a two-stage training strategy comprising mid-training and post-training. The Video-LLM can respond in two modes. In thinking mode, it generates a reasoning trace $r$ before the answer $\hat{a}$, whereas in non-thinking mode it outputs $\hat{a}$ directly. In the mid-training stage, we use the non-thinking mode and train only the visual projector of $\pi_\theta$ to predict the correct answer from the rendered EEG video, so that the model learns to recognize EEG patterns. In the post-training stage, we switch to the thinking mode. For each training example, the mid-trained model generates several rollouts, each consisting of a reasoning trace and an answer. The model is first fine-tuned on the rollouts that reach the correct answer, and then further optimized with RL, in which it continually generates new rollouts that are rewarded by the correctness of their answers (Sec.~\ref{sec:posttraining}).

\subsubsection{Mid-training: Adapting EEG as Video}
\label{sec:answer_alignment}

The goal of this stage is for the model to recognize EEG patterns in
the rendered video. To this end, we train only the visual projection modules, which map
vision-encoder features into the language model, and keep the vision
encoder and the language model frozen. Let
$\phi \subset \theta$ denote the projector parameters. In this stage, the training target for each input $(v, q)$
is the answer $a$ alone, with no reasoning trace preceding it. We
train the projector by minimizing the negative log-likelihood of this
target
\begin{equation}
    \mathcal{L}_{\mathrm{proj}}(\phi)
    = -
    \log \pi_\theta\left(a \mid v, q\right),
\end{equation}
which is minimized with respect to $\phi$ only. The resulting model serves as the starting point of post-training.

\begin{table}[t]
\centering
\caption{\textbf{Evaluation tasks.} Duration is the length of the model
input per test recording. $^\dagger$External evaluation dataset of 41 continuous
recordings, 141\,h in total. For
the final-answer evaluation in Sec.~\ref{sec:main_results}, we use
30-minute excerpts.}
\label{tab:datasets}
\vspace{1mm}
\footnotesize
\setlength{\tabcolsep}{0pt}
\renewcommand{\arraystretch}{0.95}
\begin{tabular*}{\columnwidth}{@{\extracolsep{\fill}}llrrr@{}}
\toprule
\multirow{2}{*}[-0.5ex]{Dataset} & \multirow{2}{*}[-0.5ex]{Task} & \multicolumn{3}{c@{}}{Duration (min)} \\
\cmidrule(l){3-5}
 &  & Min & Median & Max \\
\midrule
TUAB~\cite{obeid2016temple}     & Abnormal EEG detection       & 0.3 & 20.7 & 59.7 \\
TUSZ~\cite{obeid2016temple}     & Seizure detection            & 0.3 & 10.0 & 30.0 \\
TUEP~\cite{obeid2016temple}     & Epilepsy diagnosis           & 1.0 & 10.0 & 30.0 \\
TUEV~\cite{obeid2016temple}     & Event type classification    & \multicolumn{3}{c@{}}{fixed 20\,s} \\
TUSL~\cite{obeid2016temple}     & Slowing event classification & \multicolumn{3}{c@{}}{fixed 20\,s} \\
EEGMAT~\cite{zyma2019electroencephalograms} & Mental workload detection & \multicolumn{3}{c@{}}{fixed 20\,s} \\
Siena$^\dagger$~\cite{detti2020eeg} & Seizure detection        & \multicolumn{3}{c@{}}{fixed 30\,min} \\
\bottomrule
\end{tabular*}
\end{table}

\subsubsection{Post-training: Reasoning over EEG as Video}
\label{sec:posttraining}


After mid-training, the model recognizes EEG patterns in the rendered
video, while its frozen language model retains the reasoning and
spatio-temporal grounding capabilities that the Video-LLM acquired in
pre-training on such tasks~\cite{bai2025qwen3}. Post-training brings these
capabilities together. We prompt the mid-trained model to reason over
the rendered EEG before answering and to cite the time interval and
scalp region of each observation, which its pre-trained grounding
capability allows it to do. The model is then trained on its own
reasoning. We use the answer labels to select the reasoning traces
that reach the correct answer for supervised
fine-tuning~\cite{zelikman2022star,zohar2025video}, and as the reward for the
subsequent RL~\cite{shao2024deepseekmath}. This improves its
EEG-specific reasoning and grounding without externally annotated
explanations. We detail both steps below.
\\

\noindent \textbf{Self-training with SFT.}
To construct training samples with reasoning traces $r$, we sample $K$ rollouts $(r^{(k)}, \hat{a}^{(k)}) \sim \pi_{\theta}(\cdot \mid v, q)$, where $k = 1, \dots, K$ and $q$ explicitly probes the model for spatio-temporal grounding in its reasoning traces. We retain the complete traces whose answers are correct, $\hat{a}^{(k)} = a$, applying only minimal structural edits while using answer correctness as a proxy for reasoning quality~\cite{zelikman2022star}. 

The refined traces $\{(v, q, \tilde{r}, a)\}$ are then used to fine-tune the mid-trained model $\pi_{\theta}$ to generate the reasoning traces followed by the answer, by minimizing
\begin{equation}
    \mathcal{L}_{\mathrm{SFT}}(\phi, \psi)
    = -
    \log \pi_\theta\left(\tilde{r}, a \mid
    v, q\right),
\end{equation}
where $\psi$ denotes the parameters of LoRA adapters~\cite{hu2021lora}
added to the language model, trained jointly with the projector
parameters $\phi$. The vision encoder and the original language-model
weights remain frozen.

\noindent \textbf{Optimization with RL.} For each example, we sample a group of
$G$ rollouts $(r^{(g)}, \hat{a}^{(g)}) \sim \pi_\theta(\cdot \mid v, q)$
and give each a binary reward $R^{(g)} = \mathbb{1}[\hat{a}^{(g)} = a]$.
Following GRPO~\cite{shao2024deepseekmath}, we normalize the rewards
within the group to obtain the advantages $A^{(g)}$, and update the
policy by minimizing
\begin{equation}
\begin{aligned}
    \mathcal{L}_{\mathrm{RL}}(\phi, \psi)
    = {} & -\frac{1}{G} \sum_{g=1}^{G} A^{(g)}
      \log \pi_\theta\big(r^{(g)}, \hat{a}^{(g)} \mid v, q\big) \\
    & + \beta\, D_{\mathrm{KL}}\big(\pi_\theta \,\|\, \pi_{\mathrm{SFT}}\big),
\end{aligned}
\label{eq:rl}
\end{equation}
where $\pi_{\mathrm{SFT}}$ is the frozen model after SFT and $\beta$
weights the KL penalty. As we take a single update per group, the
clipped ratio of GRPO reduces to this form.

\providecommand{\tblPending}{\textcolor{gray}{\textit{TBD}}}

\section{Experiments and Results}
\label{sec:exp}

\begin{table*}[t]
    \centering
    \caption{\textbf{Final-answer accuracy.}
    Balanced accuracy (\%) on the held-out test sets of six tasks and on
    Siena, an external seizure-detection dataset unseen by all models.
    EEG encoders use one linear probe per task, whereas all other models
    are single models evaluated on every task. Prior EEG-LLMs are published models, and our raw-EEG variants replace
    the vision encoder of \ours{} with a pre-trained EEG encoder under the
    same training. Mean is the unweighted
    average over the seven evaluations. The best result in each column
    is in \textbf{bold}. The \colorbox{baseRow}{highlighted row} is the base
    model of \ours{}, and small \textcolor{gainTxt}{+x.x} values give the gain
    of \ours{} over it. $^{\ddagger}$Results reported
    in the original paper, shown in italics and not ranked.}
    \label{tab:main_results}
    \vspace{1mm}
    \small
    \setlength{\tabcolsep}{2pt}
    \renewcommand{\arraystretch}{1.02}
    \begin{tabularx}{\textwidth}{@{}c@{\hspace{5pt}}l l *{7}{Y} >{\columncolor{meanCol}}Y@{}}
        \toprule
        & \textbf{Model} & \textbf{Input} & TUAB & TUSZ & TUEP & TUEV & EEGMAT & TUSL & Siena & \textbf{Mean} \\
        \midrule
        & \textcolor{mutedtxt}{Random} & \inp{--}
            & \textcolor{mutedtxt}{50.0} & \textcolor{mutedtxt}{50.0} & \textcolor{mutedtxt}{50.0}
            & \textcolor{mutedtxt}{16.7} & \textcolor{mutedtxt}{50.0} & \textcolor{mutedtxt}{33.3}
            & \textcolor{mutedtxt}{50.0} & \textcolor{mutedtxt}{42.9} \\ \lightrule
        \grp{6}{EEG encoders}{(linear probes)}
        & BIOT~\cite{yang2023biot}      & \inp{Raw EEG} & 82.4 & 52.2 & 84.7 & 36.3 & 51.9 & 46.4 & 51.1 & 57.9 \\
        & EEGPT~\cite{wang2024eegpt}     & \inp{Raw EEG} & \best{86.1} & 53.9 & 70.1 & 40.7 & 64.8 & 54.5 & 48.9 & 59.9 \\
        & LaBraM~\cite{jiang2024large}    & \inp{Raw EEG} & 81.8 & 55.6 & 71.5 & 44.3 & \best{81.5} & 59.6 & 48.9 & 63.3 \\
        & CBraMod~\cite{wang2025cbramod}   & \inp{Raw EEG} & 81.3 & 60.2 & \best{86.8} & 44.2 & 53.7 & 63.6 & 55.3 & 63.6 \\
        & REVE~\cite{el2026reve}      & \inp{Raw EEG} & 84.3 & 59.2 & 73.6 & 41.2 & 66.7 & 72.4 & 48.9 & 63.8 \\
        & CodeBrain~\cite{ma2026codebrain} & \inp{Raw EEG} & 84.3 & 58.5 & 84.0 & 52.6 & 72.2 & 58.1 & 57.4 & 66.7 \\
        \lightrule
        \grp{3}{Prior}{EEG-LLMs}
        & NeuroLM-B~\cite{jiang2025neurolm}       & \inp{Raw EEG} & 82.9 & 51.7 & 54.9 & 40.0 & 66.7 & 59.9 & 51.1 & 58.2 \\
        & CerebraGloss-3B~\cite{gu2026cerebragloss} & \inp{Image}  & 58.8 & 60.5 & 70.8 & 30.8 & 50.0 & 61.3 & 51.1 & 54.8 \\
        & NeuroNarrator$^{\ddagger}$~\cite{wang2026neuronarrator} & \inp{Signal+Image}
            & \textit{70.8} & -- & -- & \textit{19.2} & \textit{58.4} & \textit{31.5} & \textit{62.5} & -- \\
        \lightrule
        \grp{3}{Our raw-}{EEG variants}
        & REVE + Qwen      & \inp{Raw EEG} & 68.0 & 69.7 & 75.0 & 47.7 & 50.0 & 62.0 & 61.7 & 62.0 \\
        & CodeBrain + Qwen & \inp{Raw EEG} & 62.4 & 73.6 & 65.3 & 42.7 & 53.7 & 54.7 & 53.2 & 58.0 \\
        & LaBraM + Qwen    & \inp{Raw EEG} & 80.1 & 74.2 & 50.0 & 32.2 & 53.7 & 58.2 & 51.1 & 57.1 \\
        \lightrule
        \grp{4}{Zero-shot}{VLMs}
        & OctoMed-7B~\cite{ossowski2026octomed}           & \inp{EEG as Video}  & 48.6 & 54.0 & 51.4 & 18.3 & 55.6 & 28.9 & 51.1 & 44.0 \\
        & MedGemma-1.5-4B~\cite{sellergren2026medgemma}      & \inp{Image} & 48.3 & 55.8 & 55.6 & 15.8 & 50.0 & 30.9 & 50.0 & 43.8 \\
        & MedGemma-27B~\cite{sellergren2025medgemma}         & \inp{Image} & 59.1 & 49.4 & 41.7 & 17.5 & 35.2 & 30.3 & 51.1 & 40.6 \\
        & \bR \base~\cite{bai2025qwen3} & \bR\inp{EEG as Video} & \bR 51.9 & \bR 47.8 & \bR 51.4 & \bR 18.6 & \bR 37.0 & \bR 29.6 & \bR 48.9 & 40.7 \\
        \lightrule

        & \bR \textbf{\ours{} (SFT)} & \bR \inp{EEG as Video}
            & \bR 77.2\gup{25.3} & \bR 79.0\gup{31.2} & \bR 56.9\gup{5.5} & \bR 55.2\gup{36.6}
            & \bR 75.9\gup{38.9} & \bR 62.8\gup{33.2} & \bR \best{81.9}\gup{33.0} &  69.8\gup{29.1} \\
        & \bR \textbf{\ours{} (RL)} & \bR \inp{EEG as Video}
            & \bR 81.0\gup{29.1} & \bR \best{86.2}\gup{38.4} & \bR 70.8\gup{19.4} & \bR \best{55.9}\gup{37.3}
            & \bR 63.0\gup{26.0} & \bR \best{74.0}\gup{44.4} & \bR \best{81.9}\gup{33.0} & \best{73.2}\gup{32.5} \\
        \bottomrule
    \end{tabularx}
\end{table*}

We first examine whether \ours answers clinical EEG questions correctly
(Sec.~\ref{sec:main_results}). We then turn to the reasoning behind its
answers, assessing whether it is spatio-temporally grounded in the
correct time intervals and scalp regions (Sec.~\ref{sec:exp_grounding})
and whether the predictions of the model depend on this evidence
(Sec.~\ref{sec:exp_interventions}). Further details on datasets, baselines, metrics, and evaluation
protocols are provided in Appendix~C.

\subsection{Experimental Setup}
\label{sec:experimental_setup}

\paragraph{Datasets and evaluation.}
We use six EEG datasets, each defining one task
(Table~\ref{tab:datasets}). Each recording is paired with a multiple-choice question derived from
its classification label, and these labels are the only annotations
used for training. Each dataset is split into training and test sets, giving
14{,}979 recordings for training and 3{,}875 held-out ones for
evaluation. We train only on videos of at most 30 minutes (180 frames), which
excludes 24 recordings (0.2\% of the training data). We further evaluate on Siena, an external seizure dataset
excluded from training, using 94 balanced 30-minute excerpts. We report
balanced accuracy (BA) on the final answer of each response, for each
task and averaged over the seven evaluations.\\
\\
\noindent \textbf{Implementation details.}
Our base model is \model{Qwen3-VL-4B-Thinking}~\cite{bai2025qwen3}, which we refer to
as \base. 
The visual projection modules comprise the main vision–language merger and three auxiliary DeepStack mergers \cite{meng2024deepstack}, which routes multi-level intermediate ViT features into the corresponding LLM layers to preserve fine-grained visual details; the vision encoder itself remains frozen throughout.
Mid-training trains these modules for 2{,}499 steps. In post-training,
we sample $K = 10$ rollouts per example, keep up to $M = 5$ that reach
the correct answer, and refine them with \model{Qwen3-32B}~\cite{yang2025qwen3}.
A refined trace is kept only if it preserves the original answer,
times and scalp locations and introduces no new numbers. We then fine-tune the projection modules together with LoRA adapters
(rank 64, $\alpha = 128$) for 1{,}200 steps. Finally, we train the same
modules with GRPO on 1{,}418 training recordings, with $G = 8$ responses
per recording and a KL weight of $\beta = 0.04$.\\
\\
\noindent \textbf{Baselines.}
We compare against four groups of baselines discussed in Appendix~A: (i) EEG foundation models with task-specific linear probes (BIOT, EEGPT, LaBraM, CBraMod, REVE, and CodeBrain); (ii) EEG--LLMs (NeuroLM-B, NeuroNarrator and CerebraGloss-3B); (iii) general-purpose and medical multimodal LLMs applied zero-shot to our rendered EEG videos; and (iv) raw-EEG variants of our approach, which replace the video encoder with EEG-specific encoders. The fourth group enables a controlled comparison with our video-based approach, allowing us to critically examine the benefits of video-based EEG representations. Specifically, we replace the vision encoder of the pre-trained \base backbone with one of three EEG-specific encoders (REVE, CodeBrain, and LaBraM), connecting each to language backbone through a trainable projector. This design retains the language backbone and training framework of \ours while replacing visual processing with direct EEG encoding. During mid-training, we train the projector on answer labels, following the same procedure as our video model. During post-training, we fine-tune these models on reasoning traces generated by \ours, rather than their own, for the reasons analyzed in Sec.~\ref{sec:analysis}~(RQ4). Before training, we revise these traces to remove references to visual features while preserving the EEG observations, their temporal information, and scalp locations.

\begin{table*}[t]
\centering
\caption{\textbf{Spatio-temporal grounding and event spotting.}
On TUSZ, we report grounded QA on seizure recordings, spatio-temporal
seizure grounding, and seizure-onset spotting. On TUEV, we report event
spotting and, for focal events, scalp-region overlap (sIoU) and joint
spatio-temporal accuracy (Acc@ST). All values are
percentages. The best result in each column is in \textbf{bold}. The
\colorbox{baseRow}{highlighted rows} are \ours{} and its base model, and
small \textcolor{gainTxt}{+x.x} values give the gain of \ours{} over it.}
\label{tab:grounding}
\vspace{1mm}
\footnotesize
\setlength{\tabcolsep}{3pt}
\renewcommand{\arraystretch}{1.05}
\begin{tabular}{@{}l *{10}{c}@{}}
\toprule
& \multicolumn{3}{c}{\makecell{Grounded QA\\(TUSZ)}}
  & \multicolumn{3}{c}{\makecell{Spatio-temporal\\grounding (TUSZ)}}
  & \multicolumn{2}{c}{\makecell{Event\\spotting}}
  & \multicolumn{2}{c@{}}{\makecell{Space and\\time (TUEV)}} \\
\cmidrule(lr){2-4}\cmidrule(lr){5-7}\cmidrule(lr){8-9}\cmidrule(l){10-11}
\textbf{Model}
  & Acc@QA & mIoP & Acc@GQA
  & m\_tIoU & m\_vIoU & vIoU@0.5
  & Onset@10s & Event@2s
  & sIoU & Acc@ST \\
\midrule
\mut{Random} & \mut{50.0} & \mut{21.0} & \mut{10.5} & \mut{13.2} & \mut{6.0} & \mut{0.6} & \mut{3.8} & \mut{24.9} & \mut{33.8} & \mut{8.9} \\
\lightrule
CerebraGloss-3B      & 69.7 & \best{81.8} & 58.8 & 21.8 & 18.1 & 6.7  & 22.9 & 27.0 & \best{56.1} & 23.0 \\
\lightrule
REVE + Qwen          & 63.6 & 29.9 & 23.6 & 25.5 & 18.1 & 12.1 & 18.6 & 16.4 & 30.5 & 8.2 \\
CodeBrain + Qwen     & 65.5 & 41.3 & 36.4 & 29.1 & 22.2 & 13.3 & 32.1 & 41.0 & 28.0 & 14.8 \\
LaBraM + Qwen        & 72.1 & 36.7 & 30.3 & 19.8 & 14.6 & 7.3  & 20.7 & 36.1 & 24.3 & 16.4 \\
\lightrule
\bR \base & \bR 84.8 & \bR 22.4 & \bR 16.4 & \bR 15.1 & \bR 9.1 & \bR 3.0
  & \bR 14.3 & \bR 21.3 & \bR 26.4 & \bR 11.5 \\
\lightrule
\bR \textbf{\ours{} (SFT)}
  & \bR 88.5\gup{3.7} & \bR 70.0\gup{47.6} & \bR 66.1\gup{49.7}
  & \bR \best{30.7}\gup{15.6} & \bR 23.3\gup{14.2} & \bR 13.3\gup{10.3}
  & \bR \best{42.1}\gup{27.8} & \bR 55.7\gup{34.4} & \bR 48.0\gup{21.6} & \bR 41.0\gup{29.5} \\
\bR \textbf{\ours{} (RL)}
  & \bR \best{93.3}\gup{8.5} & \bR 69.7\gup{47.3} & \bR \best{67.3}\gup{50.9}
  & \bR 30.6\gup{15.5} & \bR \best{25.5}\gup{16.4} & \bR \best{17.6}\gup{14.6}
  & \bR \best{42.1}\gup{27.8} & \bR \best{57.4}\gup{36.1} & \bR 53.8\gup{27.4} & \bR \best{49.2}\gup{37.7} \\
\bottomrule
\end{tabular}
\end{table*}


\subsection{Final-answer Accuracy}
\label{sec:main_results}
Table~\ref{tab:main_results} reports balanced accuracy on the seven
evaluations. The encoder baselines use one linear probe per task,
whereas \ours and the other language models are single models
evaluated on all tasks. \ours achieves the highest mean balanced accuracy both after SFT
(69.8) and after RL (73.2), outperforming the strongest encoder probe,
CodeBrain (66.7), by 3.1 and 6.5 points, respectively. The largest performance margin is observed on Siena, an out-of-domain test set unseen by all models, where it outperforms every baseline by at least 19.4 percentage points.
RL improves five of the seven evaluations, most on TUEP and TUSL, where SFT was weakest.

\begin{figure*}[t]
    \centering
    \includegraphics[width=1\textwidth]{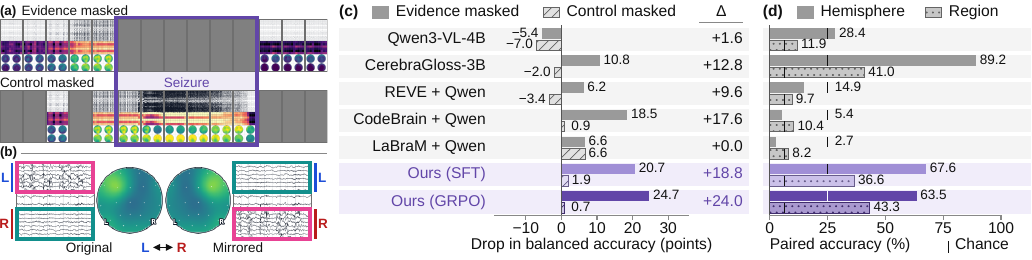}
    \caption{
        \textbf{Counterfactual interventions on evidence.}
        \textbf{(a)} Temporal masking grays out either the
        \inbox{bxseizure}{seizure frames} (evidence) or an equal number of
        seizure-free frames (control).
        \textbf{(b)} Spatial mirroring swaps the \inbox{bxleft}{left} and
        \inbox{bxright}{right} electrode signals, moving the event to the
        opposite hemisphere.
        \textbf{(c)} Drop in balanced accuracy when the evidence (\swsolidbar) or
        control (\swhatchbar) frames are masked, where $\Delta$ is the
        evidence drop minus the control drop. A model that relies on the
        seizure has a large $\Delta$. Bars left of zero mean accuracy rose.
        \textbf{(d)} Paired accuracy under mirroring. A prediction counts as
        correct only if the model names the correct hemisphere (\swsolidbar) or
        scalp region (\swdotbar) on both the original and the mirrored input.
        Chance is marked by \swtick\,.
    }
    \label{fig:evidence_grounding}
\end{figure*}

\subsection{Spatio-Temporal Grounding}
\label{sec:exp_grounding}

We evaluate the spatio-temporal grounding of \ours against expert annotations from TUSZ and TUEV, which specify the onset (start time), offset (end time), and channels of each seizure or event. We group the channels into five scalp regions, two in each hemisphere (the left and right halves of the head) and one along the midline (Sec.~\ref{sec:eeg_video}). These annotations are used only for evaluation.\\
\\
\noindent \textbf{Tasks and metrics.}
As EEG, like video, unfolds over time and
across space, we adapt three video grounding tasks. In \emph{grounded
question answering}~\cite{xiao2024can}, the model answers a clinical
question about a TUSZ seizure recording, and we score the intervals
cited in its reasoning with Acc@QA, mIoP, and Acc@GQA, where Acc@QA is
computed on seizure recordings only. In \emph{spatio-temporal
grounding}~\cite{zhang2020does,yang2022tubedetr}, it states when a seizure occurs and
which scalp regions it involves, and we report m\_tIoU, m\_vIoU, and
vIoU@0.5 with seconds as frames and region sets as boxes. Following
\emph{action spotting}~\cite{giancola2018soccernet,hong2022spotting}, we define Onset@10s
and Event@2s as the fraction of predicted times within 10\,s of a TUSZ
seizure onset or 2\,s of a TUEV event. For focal TUEV events, we
further define sIoU, the overlap between predicted and annotated scalp
regions, and Acc@ST, which requires both the timing threshold and
$\mathrm{sIoU} \geq 0.5$, following the joint criterion of Acc@GQA.
\\

\noindent \textbf{Results.} As shown in Table~\ref{tab:grounding}, the evidence that \ours cites in
grounded question answering is well aligned with the expert
annotations. On average, 70.0\% of each cited interval falls within the
annotated seizure, against 22.4\% for \base. Moreover, 66.1\% of
recordings are both answered correctly and supported by such evidence,
exceeding all baselines, including CerebraGloss (58.8\%). In
spatio-temporal grounding, where the model is asked for the interval
and scalp regions of a seizure, \ours localizes seizures in time with
an m\_tIoU of 30.7\%, against 15.1\% for \base and 21.8\% for
CerebraGloss. In event spotting, it outperforms CerebraGloss by 19.2
points in locating seizure onsets, and on focal events it leads by
18.0 points on Acc@ST, which requires both the time and the scalp regions to be correct. RL preserves or improves these results.

\subsection{Counterfactual Interventions on Evidence}
\label{sec:exp_interventions}

Following intervention-based evaluations of visual
grounding~\cite{xiao2024can,petsiuk2018rise,li2022equivariant}, we perturb the evidence in the
input and observe how the predictions change
(Fig.~\ref{fig:evidence_grounding}(a,b)). A grounded model should be
sensitive to changes in the evidence and invariant to changes
elsewhere.\\
\\
\noindent \textbf{Temporal masking.}
On TUSZ, we mask the \emph{evidence frames}, which contain an annotated
seizure, and, as a control, an equal number of seizure-free
\emph{control frames} from the same recording
(Fig.~\ref{fig:evidence_grounding}(a)). A temporally grounded model
should be sensitive to the former and largely invariant to the latter.
As shown in Fig.~\ref{fig:evidence_grounding}(c), masking the evidence
frames lowers the balanced accuracy of \ours by 20.7 points, whereas
masking the control frames lowers it by only 1.9 points. This
difference of 18.8 points exceeds that of CerebraGloss (12.8 points),
while \base responds similarly to both interventions. The predictions
of \ours thus depend specifically on the frames that contain the
seizure, and this dependence strengthens after RL, where the difference
reaches 24.0 points.\\
\\
\noindent \textbf{Spatial mirroring.}
On TUEV, we exchange the left and right electrode signals before
rendering, which moves each event to the opposite hemisphere, as shown
in Fig.~\ref{fig:evidence_grounding}(b). A spatially grounded
prediction should be equivariant to this flip. We report paired
accuracy, which counts an event only if the model is correct on both
the original and the mirrored input. \ours reaches a paired accuracy of
67.6\% for the hemisphere and 36.6\% for the scalp region, against
28.4\% and 11.9\% for \base (Fig.~\ref{fig:evidence_grounding}(d)). 

\section{Findings}
\label{sec:analysis}

We next investigate \textit{where} the grounding of \ours comes from and \textit{how}
far it extends. We ask whether it depends on reasoning supervision
(RQ1), which visual renderings it relies on (RQ2), whether it holds on
recordings longer than those seen in training (RQ3), and whether it
extends to models that read the raw signal (RQ4). Details are provided
in Appendix~D.


\subsection*{RQ1. What does reasoning self-training add?}
\label{sec:reasoning_supervision}

\ours is trained without externally annotated spatio-temporal events, 
and its only
supervision on where and when events occur comes from the reasoning of
the mid-trained model. To isolate the role of this reasoning, we
compare it with a variant post-trained with SFT from the same
mid-trained checkpoint, on the same recordings, labels, and number of
steps, but trained to answer directly without reasoning. As shown in Table~\ref{tab:reasoning_supervision}, reasoning improves final answer performance, with larger gains in spatio-temporal grounding. With the same labels,
the reasoning model localizes more than twice as many seizure onsets
within 10\,s, and its spatial answers follow the signal far more
reliably when the recording is mirrored. 

\begin{table}[t]
    \centering
    \caption{
    \textbf{Effect of reasoning supervision.} Mean BA is averaged over the seven evaluations, grounding metrics follow Table~\ref{tab:grounding}, and mirroring reports paired accuracy (Fig.~\ref{fig:evidence_grounding}(d)). All values are percentages.
    }
    \label{tab:reasoning_supervision}
    \footnotesize
    \setlength{\tabcolsep}{0pt}
    \renewcommand{\arraystretch}{1.08}
    \begin{tabular*}{\columnwidth}{@{\extracolsep{\fill}}l*{6}{c}@{}}
        \toprule
        & & \multicolumn{3}{c}{Spatio-temporal grounding}
        & \multicolumn{2}{c@{}}{Mirroring} \\
        \cmidrule(lr){3-5}\cmidrule(l){6-7}
        \textbf{Supervision}
        & \makecell{Mean\\BA}
        & Onset@10s & m\_tIoU & m\_vIoU
        & \makecell{Hemi-\\sphere} & Region \\
        \midrule
        Answer-only    & 64.4 & 17.9 & \textbf{31.1} & 20.5 & 52.7 & 26.9 \\
        With reasoning & \textbf{69.8} & \textbf{42.1} & 30.7 & \textbf{23.3} & \textbf{67.6} & \textbf{36.6} \\
        \bottomrule
    \end{tabular*}
\end{table}



\finding{
Reasoning \textit{self-training} improves \ours{}'s spatio-temporal grounding and final answer accuracy without requiring reasoning or localization annotations.
}


\subsection*{RQ2. How does each visual rendering contribute to performance?}
\label{sec:visual_representation}

We ablate the visual renderings in our EEG videos to assess their contribution to final answer performance and spatio-temporal grounding (Sec.~\ref{sec:eeg_video}). 
To this end,
we retrain the model with individual renderings removed or selected subsets retained, using the rest of the setup intact. 
As shown in Table~\ref{tab:panel_grounding}, the complete four-view representation achieves the highest joint spatio-temporal grounding accuracy for focal events (41.0\% Acc@ST, compared with at most 24.6\% for any other variant), alongside the best mean balanced accuracy. Removing the Waveform rendering reduces every evaluated metric, indicating its importance for both classification and grounding. Our results also indicate that different event types rely on different visual cues for grounding. Seizure grounding relies primarily on Waveforms, whereas focal-event grounding benefits from combining multiple cues: Sharpness improves temporal localization, while Topographic and Spectrogram features improve spatial localization.

\finding{
Different visual EEG renderings provide complementary cues that boost performance across diverse EEG tasks.}

\subsection*{RQ3. Does training on short EEG recordings generalize to longer ones?}
\label{sec:temporal_generalization}

\begin{figure}[t]
    \centering
    \setlength{\abovecaptionskip}{3pt}
    \includegraphics[width=0.95\columnwidth]{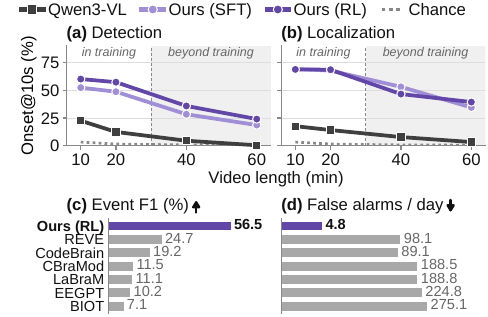}
    \caption{
        \textbf{Seizure localization in long recordings.}
        \textbf{(a, b)} Needle-in-a-haystack evaluation, where a single
        seizure is placed in a video of 10 to 60 minutes. We report Onset@10s as a function of video length.
        In (a), the model is asked whether the video contains a seizure and, if so, when it starts. In (b), it is told the video contains a seizure and asked only when it starts.
        The shaded region marks lengths beyond training.
        \textbf{(c, d)} Event F1 (c) and false alarms per day (d) for seizure
        detection on full-length Siena recordings, for \ours{} and EEG encoders.
        Each recording is several hours long and is processed with 20-minute
        sliding windows. False alarms in (d) are shown on a log scale.
    }
    \label{fig:needle}
\end{figure}

While \ours is trained on recordings of at most 30 minutes, real-world clinical EEG recordings routinely last several hours. We evaluate long-horizon generalization on 
Siena data under two settings: (i) a needle-in-a-haystack protocol~\cite{zhang2024long,lu2025needle} that embeds an annotated seizure inside 10- to 60-minute excerpts, and (ii) full-recording inference using overlapping 20-minute windows across multi-hour sessions.
\ours achieves superior seizure localization performance across all evaluated recording lengths, including 60-minute inputs, twice the maximum training duration. In contrast, \base approaches chance-level performance at longer EEG durations (Fig.~\ref{fig:needle}(a, b)). RL further improves
detection, which requires determining whether a seizure is present before localization (Fig.~\ref{fig:needle}(a)). Over complete multi-hour
recordings, \ours achieves more than twice the event F1 of pre-trained EEG
encoders (56.5\% against at most 24.7\%) while generating nearly 18 times
fewer false alarms (
Fig.~\ref{fig:needle}(c, d)).

\finding{
\ours generalizes spatio-temporal localization to multi-hour EEG recordings well beyond its training context.
}

\begin{table}[t]
    \centering
    \caption{
        \textbf{Visual cues for grounding.} Each variant keeps the feature
        marked \cmark{} in the frame and drops those marked \xmark{}, and is
        retrained on the same recordings with the same recipe. A feature
        matters more when removing it lowers a metric more. Wave.:
        waveforms, Spec.: spectrograms, Topo.: topographic maps, Sharp.:
        sharpness. Mean BA is averaged over the seven evaluations, and
        grounding metrics follow Table~\ref{tab:grounding}. All values are
        percentages, and the best result in each column is in \textbf{bold}.
    }
    \label{tab:panel_grounding}
    \vspace{1mm}
    \small
    \setlength{\tabcolsep}{2pt}
    \renewcommand{\arraystretch}{1.1}
    \resizebox{0.48\textwidth}{!}{
    \begin{tabular}{@{}cccc@{\hspace{8pt}}ccccc@{\hspace{6pt}}>{\columncolor{meanCol}}c}
        \toprule
        \multicolumn{4}{c}{EEG Renderings} & \multicolumn{2}{c}{TUSZ seizures}
          & \multicolumn{3}{c}{TUEV focal events} & \\
        \cmidrule(r){1-4}\cmidrule(lr){5-6}\cmidrule(lr){7-9}
        Wave. & Spec. & Topo. & Sharp. & Acc@GQA & m\_tIoU & Event@2s & sIoU & Acc@ST & Mean BA \\
        \midrule
        \cmark & \cmark & \cmark & \cmark & 66.1 & 30.7 & \textbf{55.7} & \textbf{48.0} & \textbf{41.0} & \textbf{69.8} \\
        \midrule
        \xmark & \cmark & \cmark & \cmark & 37.0 & 23.2 & 11.5 & 23.4 & 6.6  & 56.9 \\
        \cmark & \xmark & \cmark & \cmark & 61.8 & 32.2 & 41.0 & 42.1 & 24.6 & 64.2 \\
        \cmark & \cmark & \xmark & \cmark & 65.5 & \textbf{41.6} & 36.1 & 35.7 & 18.0 & 62.7 \\
        \cmark & \cmark & \cmark & \xmark & \textbf{70.3} & 34.7 & 18.0 & 34.2 & 9.8  & 67.7 \\
        \bottomrule
    \end{tabular}
    }
\end{table}

\subsection*{RQ4. Does reasoning self-training extend to raw-EEG LLMs?}
\label{sec:reasoning_transfer}

\begin{wraptable}{r}{0.50\columnwidth}
    \centering
    \scriptsize
    \setlength{\tabcolsep}{0pt}
    \renewcommand{\arraystretch}{1.05}
    \begin{tabular*}{0.50\columnwidth}{@{\extracolsep{\fill}}lccc@{}}
        \toprule
        Model & BA$\uparrow$ & Grnd.$\uparrow$ & Corr.$\downarrow$ \\
        \midrule
        REVE + Qwen      & 51.0 & 3.7  & 12.8 \\
        CodeBrain + Qwen & 50.7 & 2.0  & 13.4 \\
        LaBraM + Qwen    & 51.3 & 3.0  & 42.1 \\
        \midrule
        \ours{}          & \textbf{60.4} & \textbf{74.3} & \textbf{5.2} \\
        \bottomrule
    \end{tabular*}
    \vspace{-4pt}
    \caption{\textbf{Reasoning after mid-training.} Raw-EEG LLMs produce largely non-grounded (Grnd.) reasoning, while (\ours{}) yields grounded traces with low corruption (Corr.), in percentage.}
    \label{tab:visual_supervision}
    \vspace{-8pt}
\end{wraptable}
We replace the vision encoder of \base with a pre-trained EEG encoder
and train the resulting EEG LLMs using the same data and full training
recipe as \ours. After mid-training, the raw-EEG variants achieve
mean balanced accuracies above chance
(Table~\ref{tab:visual_supervision}). However, under the same prompt,
at most 3.7\% of their generated traces state a specific observation,
compared with 74.3\% for \ours. The scarcity of such descriptions
limits the supervision available for reasoning self-training.
Post-training on \ours's reasoning traces after mid-training improves
performance across all raw-EEG variants, increasing mean balanced
accuracy by up to 11.0 percentage points, to at most 62.0\% (Table~\ref{tab:main_results}). These
models also achieve grounded QA accuracies of up to 36.4\%, above the
chance level of 10.5\% (Table~\ref{tab:grounding}). Nevertheless, both classification and grounding
remain below \ours, which achieves 69.8\% mean balanced accuracy and
66.1\% grounded QA accuracy with rendered-video input.


\finding{
Self-training is ineffective for raw-EEG LLMs, whereas distilling reasoning traces from \ours{} improves their EEG reasoning.
}

\FloatBarrier
\section{Expert Review}
To complement the grounding metrics, an EEG expert evaluated the reasoning generated by \ours for 25 recordings totaling 8,820 seconds of EEG, sampled across six tasks and restricted to cases with correct final answers (see Appendix E). Each case was rated on a 5-point scale from 1 (wrong or invented) to 5 (accurate), with a mean reasoning score of 3.72/5. 
Most cases (64\%) were rated accurate (4 or 5), with a further 24\% judged partially correct (3), totaling 88\% with at least partially correct reasoning. Only
three cases were rated mostly inaccurate or wrong, for example due to
a mislocalized artifact.


\section{Conclusion}
\label{sec:conclusion}
\textbf{Summary}. We introduced \ours, demonstrating, to our knowledge, the first adaptation of a Video-LLM for spatio-temporal reasoning over EEG. Our analyses highlight the benefits of reasoning self-training over answer-only fine-tuning and of combining multiple rendered views to support long-video understanding. \ours also outperforms signal-input counterparts following self-training and generalizes to recordings longer than those seen during training.\\
\noindent \textbf{Limitation}. \ours may inherit biases from the patient populations, clinical labels, and recording protocols represented in the EEG datasets used. Expert evaluation is limited in scale, and the reported results do not establish clinical readiness for deployment.\\
\noindent \textbf{Broader Impact and Societal Considerations}. Using open-weight models and task-level supervision, our approach supports research on transferring Video-LLM reasoning and grounding to non-visual physiological signals, with potential applications in BCIs.



{
    \small
    \bibliographystyle{ieeenat_fullname}
    \bibliography{supplementary}
}
\appendix
\clearpage\setcounter{page}{1}\maketitlesupplementary
\setcounter{table}{0}\setcounter{figure}{0}
\renewcommand{\thetable}{S\arabic{table}}
\renewcommand{\thefigure}{S\arabic{figure}}

\section{Related Works}
\label{app:related}
This appendix extends the related-work discussion of the main paper. We first review Video-LLMs and spatio-temporal grounding in video, on which our approach builds. We then discuss EEG foundation models, EEG-language models, and methods for learning to reason without reasoning annotations.

\paragraph{Video-LLMs and spatio-temporal grounding.}
Video-LLMs connect a pre-trained visual encoder to a large language model and are tuned on large video-instruction corpora, extending video understanding from recognition to open-ended question answering~\cite{Maaz2024_VideoChatGPT,Zhang2023_VideoLLaMA}.
Subsequent work scales these models from short clips to hour-long videos by compressing frame tokens with query transformers~\cite{Zhang2023_VideoLLaMA,Ren2024_TimeChat}, maintaining an online memory bank~\cite{He2024_MALMM}, or extending the context of the language backbone~\cite{Zhang2024_LongVA}. The Qwen-VL line, on which \base{} builds, processes images and videos in one pathway with dynamic resolution and multimodal rotary position embeddings~\cite{Wang2024_Qwen2VL}, adds absolute time encoding for videos up to hours long~\cite{Bai2025_Qwen25VL}, and aligns frames with textual timestamps~\cite{Bai2025_Qwen3VL}. Long-video benchmarks~\cite{Mangalam2023_EgoSchema,Fu2025_VideoMME,Zhou2025_MLVU,Wu2024_LongVideoBench} evaluate this capability over minutes to hours of footage, and needle-in-a-haystack protocols evaluate the retrieval of a short target within a long video~\cite{Zhao2025_NeedleInAVideoHaystack,Zhang2024_LongVA}.
Beyond question answering, a large body of work addresses the localization of visual evidence. Temporal grounding localizes the moment described by a sentence~\cite{Gao2017_TALL,Hendricks2017_DiDeMo,Lei2021_QVHighlights}; grounded VideoQA returns the segment that supports each answer, and scores answer and segment jointly~\cite{Xiao2024_CanITrustYourAnswer_NExTGQA}; spatio-temporal video grounding predicts a tube, a time interval with one box per frame, scored by temporal and spatio-temporal IoU~\cite{Zhang2020_WhereDoesItExist_VidSTG,Yang2022_TubeDETR,Jin2022_STCAT}; and action spotting marks the time point of each event within a tolerance~\cite{Giancola2018_SoccerNet,Hong2022_SpottingTemporallyPreciseEvents}. Recent Video-LLMs address these tasks directly within the language model, through timestamp binding~\cite{Ren2024_TimeChat}, boundary-aware training~\cite{Huang2024_VTimeLLM} and direct prediction of coordinates or masks~\cite{Li2025_LLaVAST,Munasinghe2025_VideoGLaMM}.
Complementary studies examine whether predictions are supported by the relevant evidence, through atemporal probes that quantify the temporal understanding required by a benchmark~\cite{Buch2022_RevisitingVideo}, perturbation-based attribution, which masks parts of the input and measures the change in the output~\cite{Petsiuk2018_RISE}, and invariant and equivariant grounding, which requires VideoQA answers to remain stable under changes to question-irrelevant scenes and to change with question-critical ones~\cite{Li2022_InvariantGroundingVideoQA,Li2022_EquivariantInvariantGroundingVideoQA}.
Collectively, these advances demonstrate that pre-trained Video-LLMs can reason about when and where events occur, including in long videos. These capabilities rely on large-scale video-text pre-training, which is difficult to replicate in specialized domains with limited paired annotations. We therefore investigate whether the spatio-temporal grounding and reasoning of Video-LLMs transfer to brain signals, a non-visual modality, without grounding annotations during training. We assess this transfer with the same families of metrics (grounded QA, tIoU and vIoU, event spotting, temporal masking and spatial mirroring).

\paragraph{EEG foundation models.}
EEG foundation models learn general-purpose representations through self-supervised pre-training on large unlabeled corpora such as the Temple University Hospital EEG corpus~\cite{Obeid2016_TUH}.
BENDR~\cite{Kostas2021_BENDR} adapts contrastive speech pre-training to raw EEG; BIOT~\cite{Yang2023_BIOT} tokenizes each channel into fixed-length segments to handle mismatched channel sets; LaBraM~\cite{Jiang2024_LaBraM} predicts the neural codes of masked channel patches; EEGPT~\cite{Wang2024_EEGPT} combines masked reconstruction with spatio-temporal representation alignment; CBraMod~\cite{Wang2025_CBraMod} models spatial and temporal dependencies with parallel criss-cross attention; CSBrain~\cite{Zhou2025_CSBrain} builds cross-scale spatio-temporal tokens; REVE~\cite{ElOuahidi2025_REVE} adds a 4D positional encoding for arbitrary electrode layouts and recording lengths; and CodeBrain~\cite{Ma2025_CodeBrain} decouples temporal and frequency tokenizers and adds a multi-scale state-space encoder.
These models share a common recipe: temporal patch tokenization (a short segment of one channel becomes a token) with masked modeling, followed by linear probing or fine-tuning of a task-specific head on short windows.
While these encoders yield representations that transfer across tasks, they are used as classifiers: they predict one label per window, so the onset and end of a seizure must be recovered by post-processing, and the spatio-temporal evidence behind a prediction is accessible only through post-hoc attribution such as attention flow~\cite{Abnar2020_AttentionFlow} or Grad-CAM~\cite{Selvaraju2017_GradCAM}.

\paragraph{EEG-language models.}
To move beyond classification, recent work couples EEG with large language models (LLMs).
ELM~\cite{Gijsen2025_ELM} aligns EEG with clinical reports through contrastive learning, enabling zero-shot classification and retrieval.
Subsequent models generate language: NeuroLM~\cite{Jiang2025_NeuroLM} treats EEG as a foreign language, feeding tokens from a text-aligned vector-quantized tokenizer to an LLM for multi-task instruction tuning, and WaveMind~\cite{Zeng2025_WaveMind} maps EEG into a space shared with text and images for open-ended conversation. CerebraGloss~\cite{Gu2026_CerebraGloss} instruction-tunes a vision-language model on EEG visualizations, NeuroNarrator~\cite{Wang2026_NeuroNarrator} aligns waveforms with topographic maps (images of scalp voltage) to generate clinical narratives, and EEG-Captioner~\cite{Jin2026_EEGCaptioner} trains on signal-level question answering and feature-grounded captions.
These models, however, are limited to short recordings~\cite{Gu2026_CerebraGloss,Jin2026_EEGCaptioner,Wang2026_NeuroNarrator}: each interprets one short segment at a time, so an event is described within its segment but is not localized within a long recording.
Approaches that handle longer recordings do so at the expense of grounding.
CELM~\cite{Pradeepkumar2026_CELM} generates clinical reports from long recordings under coarse recording-level supervision: a report summarizes the recording as a whole and provides no fine-grained spatio-temporal grounding of the individual events.
EEGAgent~\cite{Zhao2025_EEGAgent} relies on external signal-analysis tools: an LLM plans calls to preprocessing, feature-extraction and event-detection tools, so its analysis is limited to what these tools provide.
A related line of work renders physiological signals as images for multimodal LLMs, building on chart and screenshot understanding~\cite{Lee2023_Pix2Struct,Li2024_SynthesizeStepByStep}, including time-series plots~\cite{Liu2025_VLTime}, ECG images with time series~\cite{Lan2025_GEM}, and EEG or polysomnography (overnight sleep recordings) images for sleep staging~\cite{Qiu2025_EEGVLM,Deng2026_SleepVLM}; these approaches operate on individual images.
Reasoning over long EEG recordings with fine-grained spatio-temporal grounding therefore remains largely unexplored.
In contrast, \ours{} renders the whole recording as a video, with one frame per 10-second window, which allows a pre-trained Video-LLM (\base{}) to apply its temporal and spatial grounding capabilities to EEG.

\paragraph{Reasoning without reasoning annotations.}
Reasoning supervision is costly to obtain; for medical images, it has been derived from scientific articles~\cite{Baghbanzadeh2026_OpenMedReason}.
Self-training removes the need for annotated rationales: STaR~\cite{Zelikman2022_STaR} fine-tunes a model on its own rationales that reach the correct answer, ReST$^{EM}$~\cite{Singh2024_BeyondHumanData_ReSTEM} scales this filtering by a binary correctness signal into an iterative expectation-maximization loop, and Video-STaR~\cite{Zohar2025_VideoSTaR} carries the idea to video instruction tuning by using existing video labels to select the model's own answers.
Reinforcement learning with verifiable rewards offers a complementary route: group relative policy optimization (GRPO), introduced with DeepSeekMath~\cite{Shao2024_DeepSeekMath}, normalizes the rewards of a group of sampled responses instead of learning a value function, and DeepSeek-R1~\cite{Guo2025_DeepSeekR1} showed that large-scale RL with rule-based rewards elicits reasoning in LLMs. Such R1-style training has since been applied to images~\cite{Liu2025_VisualRFT}, video~\cite{Feng2025_VideoR1}, time series rendered as plots~\cite{Zhang2025_TimeMaster}, electrocardiograms~\cite{Jin2026_ECGR1}, multimodal diagnosis~\cite{Zhang2026_MedTVTR1} and medical vision-language models~\cite{Jeddi2026_WhenDoesRLHelpMedicalVLMs}.
\ours{} combines self-training and RL for EEG rendered as video, a setting in which the model must first learn to interpret its input and in which neither reasoning nor grounding annotations are available. We further analyze what this reasoning supervision adds beyond answer labels (RQ1) and whether it extends to raw-EEG LLMs (RQ4).

\section{Additional Method Details}
\label{app:method}

This appendix details the two components of \ours{}: the rendering of an EEG recording as a video (Appendix~\ref{app:eeg_video}; main paper Sec.~2.2) and the adaptation of the Video-LLM to this video (Appendix~\ref{app:adaptation}; main paper Sec.~2.3).

\subsection{EEG as Video Details (main paper Sec.~2.2)}\label{app:eeg_video}

\begin{figure}[t]
\centering
\includegraphics[width=\linewidth]{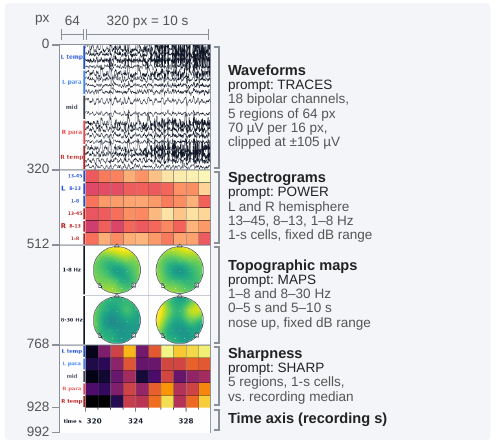}
\caption{\textbf{One frame of the EEG video, as the model receives it.} A 10-second window (320--330\,s) of a TUSZ test recording, during an annotated seizure. As high-frequency, sharp activity builds up in the \textbf{Waveforms}, the 13--45\,Hz \textbf{Spectrogram} cells and the \textbf{Sharpness} cells brighten. The pixel rulers and the notes on the right are added for this figure.}
\label{fig:b1_frame}
\end{figure}

In main paper Sec.~2.2, each recording becomes a video in which every frame shows one 10-second window through four renderings: Waveforms, Spectrograms, Topographic maps and Sharpness (main paper Fig.~1). This part gives the rules that produce these frames: how a recording is prepared and cut into windows, how each rendering is computed and what it shows, how the frames enter the model, and which modified frames the ablations and interventions use. One renderer with fixed settings draws every frame, in training and in evaluation (Fig.~\ref{fig:b1_frame}, Table~\ref{tab:b1_renderings}).

\paragraph{Signal preparation.}
We keep the 19 scalp electrodes of the standard 10--20 placement system, plus Fpz and the two ear electrodes A1 and A2.
If a recording lacks one of these electrodes, its signal is set to zero.
All signals are resampled to 200\,Hz.
We band-pass filter the whole recording to 0.3--75\,Hz and remove power-line noise with a notch filter (60\,Hz for the five TUH datasets, 50\,Hz for EEGMAT and Siena). All band-pass filters, here and in the renderings, are fourth-order zero-phase Butterworth filters.
The frames are drawn from the 19 scalp electrodes.
The same prepared signal, with all 22 electrodes, is the input of the EEG encoder baselines and the raw-EEG variants (Appendices~\ref{app:encoder_setup} and~\ref{app:hparams}).

\paragraph{Montage and renderer filter.}
The Waveforms, Spectrograms and Sharpness are computed from 18 bipolar channels. A bipolar channel is the difference between two neighbouring electrodes, written for example Fp1--F7.
The channels form five scalp regions, drawn top to bottom in this order:
left temporal (Fp1--F7, F7--T3, T3--T5, T5--O1), left parasagittal (Fp1--F3, F3--C3, C3--P3, P3--O1), midline (Fz--Cz, Cz--Pz), right parasagittal (Fp2--F4, F4--C4, C4--P4, P4--O2) and right temporal (Fp2--F8, F8--T4, T4--T6, T6--O2).
The Topographic maps instead use the 19 scalp electrodes, each taken relative to the mean of all 19.
Before drawing, each 10-second window is filtered on its own to keep 0.5--45\,Hz. All four renderings show this filtered window.

\paragraph{Windowing.}
Each recording is split into 10-second windows with no overlap, and each window gives one frame. An incomplete window at the end is discarded.
For TUAB, TUSZ and TUEP, the video covers the whole recording.
For these datasets, windows are kept in consecutive pairs. A pair is dropped if either window is nearly flat, that is, if its standard deviation over time, averaged over electrodes C3, C4, O1 and O2, is at most 3\,$\mu$V.
Flat stretches therefore leave gaps in the video, which stay visible because the time axis and the timestamps give the time in the recording.
A training video has at most 180 frames (30\,min).
At test time, the model sees all windows of a recording in one pass, up to 358 frames (59.7\,min) for TUAB.
TUEV and TUSL examples are 20-second clips (two frames) centred on the annotated event. EEGMAT recordings are cut into 20-second clips with no overlap.

\begin{table*}[t]
\centering
\footnotesize
\caption{\textbf{The four renderings of a frame.} In brackets: the name used in the prompt. Only the \textbf{Sharpness} is scaled per recording, relative to a reference computed from the whole recording. The other renderings use \textbf{fixed scales} shared by all recordings.}
\label{tab:b1_renderings}
\setlength{\tabcolsep}{4pt}
\begin{tabular}{@{}lllll@{}}
\toprule
Rendering & What it shows & Resolution & Scale & Colour map\\
\midrule
Waveforms (TRACES) & bipolar channels as traces, 5 region bands & 1\,s = 32\,px; 0.5--45\,Hz & 70\,$\mu$V per 16\,px, clipped at $\pm$105\,$\mu$V & black lines\\
Spectrograms (POWER) & band power per hemisphere (dB) & 1-s cells; 3 bands & fixed range per band & magma\\
Topographic maps (MAPS) & interpolated electrode band power (dB) & 5-s maps; 2 bands & fixed range per band & viridis\\
Sharpness (SHARP) & peak second difference vs.\ recording median & 1-s cells; 5 regions & $\log_{10}$ ratio on $[0, 0.977]$ & inferno\\
\bottomrule
\end{tabular}
\end{table*}

For each rendering, from top to bottom of the frame, we state what it is, what it tells a reader who is not an EEG expert, and how it is drawn.

\paragraph{Waveforms.}
\emph{What it is.} The Waveforms are the time-domain signal: the voltage of each bipolar channel plotted as a trace over the 10 seconds of the window, as a clinician reads it. Time runs from left to right and amplitude is vertical.
\emph{What it tells.} The morphology (shape) of the traces shows what kind of activity is present, such as low-frequency waves, high-frequency oscillations or sudden spikes, and when it occurs. The traces are grouped by scalp region, so the group in which a pattern appears gives its spatial location.
\emph{How it is drawn.} Each of the five regions has its own horizontal band, and its channels are evenly spaced within it. Each channel is a thin black trace around its own baseline, with positive voltage upward. The amplitude scale is the same for every recording, so a taller wave always means a larger voltage; values beyond the limit are clipped (scale and limit in Table~\ref{tab:b1_renderings}). Thin grey vertical lines mark every second.

\paragraph{Spectrograms.}
\emph{What it is.} An EEG signal is a mixture of oscillations at different frequencies, measured in Hz. The Spectrograms are a time-frequency representation of the signal: a heatmap of the signal power in a low, a middle and a high frequency band, for each second of the window and for each hemisphere.
\emph{What it tells.} A brighter cell means more power in that frequency band during that second. The Spectrograms therefore expose frequency-specific activity and its temporal evolution, which is hard to judge from the traces alone, and show whether the two hemispheres differ. For example, a recording dominated by low-frequency activity is bright in the lowest band.
\emph{How it is drawn.} There are six rows: the bands 13--45, 8--13 and 1--8\,Hz for the left hemisphere, then the same three for the right. The left hemisphere pools the eight left temporal and left parasagittal channels, and the right hemisphere the eight right parasagittal and right temporal channels. Each row has ten cells, one per second. To obtain the power of a band, the window is band-pass filtered to that band and the signal is squared. A cell shows this power averaged over its second and over the eight channels, on a logarithmic (dB) scale. The colour scale (magma) is fixed for each band, so the same colour means the same power in every recording: $[-14.1, 27.7]$\,dB at 13--45\,Hz, $[-6.2, 23.5]$\,dB at 8--13\,Hz and $[-4.3, 34.1]$\,dB at 1--8\,Hz. Values outside the range take the end colour.

\paragraph{Topographic maps.}
\emph{What it is.} A Topographic map is a top view of the scalp, coloured by the signal power at each location.
\emph{What it tells.} The maps show the spatial distribution of the activity: whether it is focal (concentrated in one area), lateralized (stronger in one hemisphere) or generalized (spread over the whole scalp). The Waveforms and Spectrograms encode location only indirectly, through the region or hemisphere of a row.
\emph{How it is drawn.} There are two rows, one for the low-frequency band (1--8\,Hz, top) and one for the higher-frequency band (8--30\,Hz). Each row has two maps, one for each 5-second half of the window, so a spatial change within the window is visible. In each map the nose points up and the left hemisphere is on the left; white dots mark the electrodes, and the letters L and R mark the sides. The value at an electrode is its power in the band over the 5 seconds, in dB, estimated with Welch's method (1-second Hann segments with 50\% overlap, averaged over the frequencies of the band). The 19 electrode values are interpolated across the scalp with a thin-plate spline. The colour scale (viridis) is fixed for each band: $[-4.7, 25.2]$\,dB at 1--8\,Hz and $[-14.3, 16.1]$\,dB at 8--30\,Hz.

\paragraph{Sharpness.}
\emph{What it is.} The Sharpness map is a heatmap of abrupt signal changes: moments where a trace bends sharply, as at the tip of a spike.
\emph{What it tells.} Transient events such as epileptic spikes last a fraction of a second and are easy to miss among the traces. A bright cell localizes such a transient in time (which second) and in space (which region), relative to what is typical for the recording.
\emph{How it is drawn.} There are five rows, one per region in the order of the Waveforms, each with ten 1-second cells. For a channel $x$ of the filtered window, we measure how sharply the trace bends at sample $n$ with the second-order temporal difference $s[n]=|x[n+2]-2x[n+1]+x[n]|$, which is large at sharp peaks. Each channel has a reference, its typical sharpness in the recording: we split the whole recording into 0.5-second segments, take the maximum of $s$ in each segment, and use the median of these maxima. The sharpness of a cell is the maximum of $s$ in the cell divided by the channel's reference, and the cell shows the largest of these ratios among the channels of its region on a logarithmic colour scale (inferno). A cell is black when its sharpest moment is no sharper than the recording's typical value, and brightest at 9.5 times that value.

\paragraph{Frame layout and labels.}
A frame is a colour image on a white background. It stacks the four renderings above a time axis, with a label column on the left and the 10-second window to its right (Fig.~\ref{fig:b1_frame}, which also gives the size of each part in pixels).
The only text in the frame is a set of labels and the time axis.
The region labels L temp, L para, mid, R para and R temp (left temporal, left parasagittal, midline, right parasagittal, right temporal) mark the Waveforms and the Sharpness rows.
The band labels 13-45, 8-13 and 1-8, each with L or R, mark the Spectrogram rows, and 1-8 Hz and 8-30 Hz mark the Topographic maps.
The time axis has a tick every second and prints the time in the recording at 0, 4 and 8\,s into the window.
The prompt (Appendix~\ref{app:adaptation}) names the four renderings TRACES (Waveforms), POWER (Spectrograms), MAPS (Topographic maps) and SHARP (Sharpness).
Model outputs use these short rendering and region names, and the examples in this supplement quote them unchanged.

\paragraph{From frames to model input.}
The frames of a recording form one video, placed before the prompt text.
The model receives every frame of the recording.
\base{} merges two consecutive frames into one temporal patch, so we pass each frame twice, and each 10-second window becomes 372 visual tokens. The longest training video (180 frames) therefore takes 66,960 visual tokens, and the longest test video (358 frames) takes 133,176.
As for any video input, \base{} places a textual timestamp before the visual tokens of each frame (e.g., \texttt{<290.0 seconds>}). We set it to the start time of the window in the recording, in whole seconds.

\paragraph{Modified frames used in the experiments.}
Three experiments of the main paper change the frames that the model sees. Each uses the renderer described above; this paragraph states what changes in the frame.

\emph{Frames with fewer renderings} (main paper Table~5; results in Appendix~\ref{app:rq2}). To measure what each rendering contributes, we train variants of \ours{} whose frames contain only some of the four renderings. Such a frame stacks the chosen renderings in their usual order, above the time axis. The frame description in the prompt names only the renderings that are present. Each variant is trained and tested on its own frames.

\emph{Temporal masking} (main paper Fig.~2(a,\,c); protocol in Appendix~\ref{app:exp_counter}). To test whether an answer depends on the frames that contain the evidence, we hide selected frames. In a masked frame, the four renderings are replaced by a uniform grey area. The region labels, the time axis and the timestamp remain, so the model still knows which moment is hidden.

\emph{Spatial mirroring} (main paper Fig.~2(b,\,d); protocol in Appendix~\ref{app:exp_counter}). To test whether the model follows where an event is on the scalp, we exchange the left and the right side of the head. The signal of each left electrode is swapped with that of the matching right electrode (for example Fp1 with Fp2, and T3 with T4) before the frames are rendered. An event on the left therefore appears on the right, and the reverse. The layout of the frame, its labels and its time axis are the same as for the original input.

\subsection{Adaptation of Video-LLMs for Spatio-temporal Reasoning Details (main paper Sec.~2.3)}\label{app:adaptation}

Main paper Sec.~2.3 adapts the Video-LLM in two stages. Mid-training trains the visual projection modules to answer questions about the rendered EEG. Post-training then makes the model reason before it answers: it is fine-tuned on its own reasoning traces that reach the correct answer (SFT) and further optimized with RL, which rewards correct answers. Both stages use only the class label of each example as supervision.
This part first gives the input format, the prompts, inference and answer extraction, and then the rules of mid-training (Appendix~\ref{app:mid_train}), SFT (Appendix~\ref{app:post_train}) and RL (Appendix~\ref{app:rl}). Prompts are quoted verbatim; square brackets mark the parts filled in for each example. Hyperparameters are in Appendix~\ref{app:hparams} and the data of each stage in Appendix~\ref{app:datasets}.

\paragraph{Model input and output format.}
Each example is a single user message. It holds the video (Appendix~\ref{app:eeg_video}), followed by the text prompt. The prompt starts with the frame description
\begin{promptbox}
The video shows the EEG as one frame per 10-second window. Each frame has four panels, top to bottom: TRACES (EEG by scalp chain), POWER (band power cells for the left and right hemisphere), MAPS (scalp maps for 1-8 Hz and 8-30 Hz, nose up, left on the left), SHARP (sharpness cells by scalp chain, brighter = sharper). The time axis at the bottom gives recording seconds.
\end{promptbox}
then a blank line, the question, the line ``Options:'' and one lettered option per line (Table~\ref{tab:b2_questions}). In the prompts, a panel is one of the four renderings and a chain is one of the five scalp regions. An ending that depends on the stage follows the options. In thinking mode, the model writes its reasoning trace, closes the reasoning block and then writes the answer. In non-thinking mode, the input already contains an empty, closed reasoning block, so the model answers directly. The endings are:
\begin{itemize}
\item \emph{Mid-training} (non-thinking): on the line after the options,
\begin{promptbox}
Answer with the option letter.
\end{promptbox}
\item \emph{Rollouts and SFT} (thinking): after a blank line, the \emph{rollout prompt}
\begin{promptbox}
Think step by step. For each important finding, state the time interval (in recording
seconds) and the scalp region where it occurs, then make your decision. Give the option
letter alone as <answer>X</answer>.
\end{promptbox}
\item \emph{RL and evaluation} (thinking): no ending; the prompt stops after the last option. We call this the \emph{evaluation prompt}.
\end{itemize}

\begin{table*}[t]\centering\footnotesize
\caption{\textbf{Question and options of each task}, used verbatim in all stages and at test time. Letters are assigned in the order shown.}\label{tab:b2_questions}
\setlength{\tabcolsep}{4pt}
\begin{tabular}{@{}p{1.7cm}p{7.9cm}p{7.1cm}@{}}\toprule
Dataset & Question & Options\\\midrule
TUAB & Is the background activity in this EEG recording normal or abnormal? & A. normal, B. abnormal\\
TUSZ, Siena & Does this EEG recording contain an electrographic seizure? & A. no seizure, B. seizure\\
TUEP & Based on this EEG, does this patient have epilepsy? & A. no epilepsy, B. epilepsy\\
TUEV & Which of the following best describes this EEG segment? & A. artifact, B. background, C. eye movement, D. generalized periodic discharges, E. periodic lateralized discharges, F. spike and sharp wave\\
TUSL & Does this EEG segment show normal background, slowing, or a seizure? & A. normal background, B. slowing, C. seizure\\
EEGMAT & Is the subject at rest or performing a mental arithmetic task? & A. rest, B. mental arithmetic\\\bottomrule
\end{tabular}\end{table*}

\paragraph{Inference.}
At test time, the model receives the evaluation prompt with thinking enabled, and it generates its reasoning followed by its answer. We decode with vLLM~\cite{Kwon2023_vLLM} in bf16 precision, using greedy decoding, a repetition penalty of 1.1, a frequency penalty of 0.5 and at most 8,192 new tokens. For the spatio-temporal grounding question (m\,tIoU and m\,vIoU in main paper Table~3), we use no frequency penalty and at most 4,096 new tokens.

\paragraph{Answer extraction.}
The model states its answer as free text after its reasoning, for example ``B. seizure''. To score it, a text-only LLM (\texttt{Qwen3-32B}, greedy decoding) maps this text to one of the options of the question. This extractor sees only the question and the answer of the model, never the recording or the correct label. It receives the system instruction
\begin{promptbox}
You map a model's answer to one of the listed options. Reply with the option letter only (A, B, C, ...). If the answer commits to no single option, or to an option that is not listed, reply NONE.
\end{promptbox}
and a user message that contains the question with its options and the answer of the model (the last 1,500 characters of the text after the reasoning):
\begin{promptbox}
Question and options:

\emph{[the question and its options]}

Model's answer:

\emph{[the answer of the model]}

Which option letter did the model choose?
\end{promptbox}
An answer mapped to NONE counts as wrong. The same extractor scores every model of main paper Table~2 that answers in free text: \ours{}, the raw-EEG variants and the zero-shot VLMs. CerebraGloss-3B and NeuroLM-B are instead scored from their option probabilities (Appendix~\ref{app:exp_base}). For the grounding questions, times and regions are extracted from the answers as described in Appendix~\ref{app:grounding}.
To validate the extractor, we compare it with a rule that reads the option letter directly from the answer text. The two agree on 99.5\% of the test outputs of \ours{} (RL) and on 99.9\% of those of \ours{} (SFT).


\subsubsection{Mid-training Details}\label{app:mid_train}

Mid-training teaches the model to recognize EEG patterns in the rendered video (main paper Sec.~2.3.1). It runs in non-thinking mode and updates only the visual projection modules; the vision encoder and the language model stay frozen.

\paragraph{Data, target and loss.}
Mid-training uses the training splits of the six datasets in main paper Table~1. Each dataset has one multiple-choice question, built from its classification label (Table~\ref{tab:b2_questions}). The options and their order are fixed, and the target is the letter of the label. An example is the whole recording for TUAB, TUSZ and TUEP, and the 20-second clip for TUEV, TUSL and EEGMAT. The data are balanced: every class of every dataset contributes the same number of examples, 2,352, and a smaller class repeats each of its recordings evenly. The loss covers only the answer (main paper Eq.~1).

\subsubsection{Post-training Details (SFT)}\label{app:post_train}

The first post-training step is reasoning self-training (main paper Sec.~2.3.2, ``Self-training with SFT''). The mid-trained model generates rollouts, the rollouts with a correct answer are selected, their wording is refined, and the model is fine-tuned on the refined traces.

\paragraph{Rollouts and selection.}
The mid-trained model generates $K{=}10$ rollouts with the rollout prompt for each of the 3,035 recordings of the SFT set, 30,350 rollouts in total (data in Appendix~\ref{app:datasets}, sampling settings in Table~\ref{tab:c5_ours}). A rollout counts as correct if it ends before the token limit, closes its reasoning block, and then gives the label X, either as \texttt{<answer>X</answer>} or as the first letter of the text after the block. Of all rollouts, 21,644 are correct. From the correct rollouts of a recording, we keep up to $M{=}5$, chosen at random. This gives 12,151 traces.

\paragraph{Trace refinement.}
\texttt{Qwen3-32B}~\cite{Qwen3-32B} then refines each kept trace, applying only minimal structural edits. It acts as an editor: it removes restarts, repetitions and withdrawn statements, and it preserves every observation with its time and scalp location, the steps to the conclusion, and the conclusion. It sees only the text of the trace, and runs greedily with thinking disabled and at most 900 new tokens. Its system instruction is:
\begin{promptbox}
You edit a model's reasoning about an EEG video into a short, to-the-point version. You are an editor, not an interpreter.

KEEP, exactly as the original states them:

- every observation about THIS recording, with its time in recording seconds (single times and ranges such as "1250-1254 s") and its place: chain names (L temp, L para, mid, R para, R temp), left or right hemisphere, the panel (TRACES, POWER, MAPS, SHARP), the frequency band;

- the steps that lead from those observations to the conclusion, and the conclusion.

REMOVE:

- "wait", "hmm", "let me check", restarts and repeated sentences;

- anything the original itself withdrew or corrected (keep only the corrected version);

- talk about the user, the question, the instructions or the video format;

- recaps of textbook definitions, except one short sentence when the decision depends on it;

- guesses about what the video might show that the original never states as seen.

Do not add anything: no new observations, times, places, numbers or interpretations, and do not turn a possibility ("might be", "could indicate") into a fact.

Plain paragraphs, no bullets, headings or bold. Usually 60-200 words.

End with a final line "Answer: X" using the same letter as the original.
\end{promptbox}
The user message contains ``Question:'' with the question and options, ``Reasoning:'' with the reasoning trace, and ``Final answer:'' with the rollout's letter, separated by blank lines. We remove the final ``Answer: X'' line from the output. A refined trace is kept only if it preserves the original answer, times and scalp locations and introduces no new numbers. These conditions are checked automatically:
\begin{itemize}
\item its answer letter equals the original one;
\item it contains neither ``wait'' nor ``hmm'';
\item every number in it occurs in the original;
\item it preserves the original's recording times: at least half of them are retained, each within 1\,s or inside a stated range of seconds;
\item it preserves the original's scalp locations: at least half of them are retained (L temp, L para, mid, R para, R temp, left hemisphere, right hemisphere).
\end{itemize}

\noindent In total, 10,356 refined traces are kept.

\paragraph{SFT target and loss.}
SFT uses the rollout prompt with the same video. The target is the refined trace, followed by the closing of the reasoning block and the answer \texttt{<answer>X</answer>} with the label X. The loss is the cross-entropy over the target tokens only, averaged within each trace (main paper Eq.~2). SFT updates the visual projection modules and the LoRA adapters added to the language model (Table~\ref{tab:c5_ours}).

\subsubsection{Post-training Details (RL)}\label{app:rl}

The second post-training step optimizes \ours{} (SFT) with GRPO on the correctness of its final answers (main paper Sec.~2.3.2, ``Optimization with RL'', Eq.~3).

\paragraph{Rollouts and reward.}
RL rollouts use the evaluation prompt in thinking mode, the same input as at test time. The reward that enters main paper Eq.~3 is binary: 1 if the answer letter of the response equals the label, and 0 otherwise. A fixed rule reads the letter from the text after the reasoning block: it takes the letter in \texttt{<answer>X</answer>} or at the start of that text. Otherwise, it takes the last option letter that stands alone in the final 200 characters of the response. A response that reaches the token limit, does not close its reasoning block or gives no letter receives 0. The remaining settings are in Appendix~\ref{app:hparams}.

\paragraph{Update.}
Each round draws 118 or 119 new training recordings from TUAB, TUSZ, TUEP, TUEV and TUSL, balanced by dataset and class.
The policy gives $G{=}8$ responses to each recording. These responses form one group and are rewarded by the rule above.
The advantage of a response is $A^{(g)}=(R^{(g)}-\bar{R})/\max(\sigma_R,0.1)$, where $\bar{R}$ and $\sigma_R$ are the mean and the standard deviation of the rewards of its group. Groups in which all rewards are equal are skipped.
In main paper Eq.~3, the log-likelihood of a response is averaged over its tokens. The KL term uses the per-token estimate $e^{\rho}-\rho-1$ with $\rho=\log\pi_{\mathrm{SFT}}-\log\pi_\theta$, clamped at 10 and averaged over the response. Both terms are then averaged over the 64 responses of an optimizer step.

\section{Additional Experiments Details}
\label{app:experiments}

This appendix supports Sec.~3 of the main paper. It gives the experimental setup (Appendix~\ref{app:exp_setup}), further results and examples for the final-answer accuracy of Table~2 (Appendix~\ref{app:exp_results}), the details of the spatio-temporal grounding evaluation of Table~3 (Appendix~\ref{app:grounding}) and those of the counterfactual interventions of Fig.~2 (Appendix~\ref{app:exp_counter}).

\subsection{Experimental Setup Details (main paper Sec.~3.1)}\label{app:exp_setup}

Main paper Table~2 compares models that differ in their input, their training and the form of their output. This part gives what is needed to judge that comparison: the data and the metric (Appendix~\ref{app:datasets}), the training settings of \ours{} and of its raw-EEG variants (Appendix~\ref{app:hparams}), and, for every baseline, what it receives, how it is decoded and how its outputs become one answer per recording (Appendix~\ref{app:exp_base}).

\subsubsection{Datasets and Evaluation Details}\label{app:datasets}

We use six EEG datasets, each defining one classification task, and Siena as an external test set for seizure detection (main paper Table~1). One sample is a whole recording (TUAB, TUSZ, TUEP), a 20-second clip (TUEV, TUSL, EEGMAT) or a 30-minute excerpt (Siena). Each sample is rendered as a video with one frame per 10-second window (Appendix~\ref{app:eeg_video}). It has one class label, which is the correct answer to its multiple-choice question (Table~\ref{tab:b2_questions}). Table~\ref{tab:c0_splits} gives the training data of each stage and the composition of every test set.

\begin{table*}[t]\centering\footnotesize
\caption{\textbf{Training and test data of each dataset.} A sample is a recording, a 20-second clip or a 30-minute excerpt, depending on the dataset (see text). The first three columns give the number of training samples used in each training stage of \ours{}. Hours: total length of the test inputs. TUEV classes: BCKG background, GPED generalized periodic discharges, PLED periodic lateralized discharges, ARTF artifact, EYEM eye movement, SPSW spike and sharp wave.}\label{tab:c0_splits}
\setlength{\tabcolsep}{5pt}
\begin{tabular}{@{}lrrrrrp{8.6cm}@{}}\toprule
 & \multicolumn{3}{c}{Training samples} & \multicolumn{2}{c}{Test} & \\\cmidrule(lr){2-4}\cmidrule(lr){5-6}
Dataset & Mid-training & SFT & RL & Samples & Hours & Test classes (number of samples)\\\midrule
TUAB & 2{,}407 & 600 & 194 & 276 & 100.1 & normal 150, abnormal 126\\
TUSZ & 3{,}063 & 800 & 340 & 875 & 127.5 & no seizure 710, seizure 165\\
TUEP & 2{,}188 & 500 & 180 & 96 & 22.1 & epilepsy 72, no epilepsy 24\\
TUEV & 6{,}605 & 593 & 600 & 2{,}533 & 14.1 & BCKG 839, GPED 687, PLED 531, ARTF 319, EYEM 111, SPSW 46\\
TUSL & 207 & 207 & 104 & 59 & 0.3 & slowing 23, seizure 19, background 17\\
EEGMAT & 335 & 335 & -- & 36 & 0.2 & rest 27, mental arithmetic 9\\\midrule
Total & 14{,}805 & 3{,}035 & 1{,}418 & 3{,}875 & 264.3 & \\\midrule
Siena & -- & -- & -- & 94 & 47.0 & seizure 47, no seizure 47\\\bottomrule
\end{tabular}\end{table*}

\paragraph{Tasks and metric.}
Table~\ref{tab:c1_tasks} lists the seven evaluation tasks of main paper Table~1, and Appendix~\ref{app:tasks} shows one worked example per task. Every model is scored on its final answer for each test sample, with balanced accuracy. Balanced accuracy is the recall of each class (the share of its samples answered correctly), averaged over the classes. Unlike plain accuracy, it does not reward always choosing the most frequent class, and several test sets are imbalanced (Table~\ref{tab:c0_splits}). The Mean in main paper Table~2 is the unweighted average over the seven evaluations. We evaluate all 3,875 test samples of the six datasets and all 94 Siena excerpts.

\begin{table}[t]\centering\footnotesize
\caption{\textbf{The seven evaluation tasks}, as in main paper Table~1. Duration: the length of the model input for one test sample. Rec.: recordings. Every task is scored with balanced accuracy.}\label{tab:c1_tasks}
\setlength{\tabcolsep}{2pt}
\begin{tabular}{@{}llrrrl@{}}\toprule
 & & \multicolumn{3}{c}{Duration (min)} & \\\cmidrule(lr){3-5}
Dataset & Task & Min & Median & Max & Test size\\\midrule
TUAB & Abnormal EEG detection & 0.3 & 20.7 & 59.7 & 276 rec.\\
TUSZ & Seizure detection & 0.3 & 10.0 & 30.0 & 875 rec.\\
TUEP & Epilepsy diagnosis & 1.0 & 10.0 & 30.0 & 96 rec.\\
TUEV & Event type classification & \multicolumn{3}{c}{fixed 20\,s} & 2{,}533 clips\\
TUSL & Slowing event classification & \multicolumn{3}{c}{fixed 20\,s} & 59 clips\\
EEGMAT & Mental workload detection & \multicolumn{3}{c}{fixed 20\,s} & 36 clips\\
Siena & Seizure detection & \multicolumn{3}{c}{fixed 30\,min} & 94 excerpts\\\bottomrule
\end{tabular}\end{table}

\paragraph{TUAB.} The TUH Abnormal EEG Corpus (v3.0.1)~\cite{Lopez2017_TUAB,Obeid2016_TUH} labels each recording as normal or abnormal.

\paragraph{TUSZ.} The TUH Seizure Detection Corpus (v2.0.6)~\cite{Shah2018_TUSZ,Obeid2016_TUH}. For every seizure, experts annotate its type, its start and end time, and the channels it involves. A recording is labelled ``seizure'' if at least one seizure of any type is annotated in it, and ``no seizure'' otherwise.

\paragraph{TUEP.} The TUH Epilepsy Corpus (v3.1.0; 200 patients)~\cite{Veloso2017_TUEP,Obeid2016_TUH} labels each recording with the diagnosis of its patient (epilepsy or no epilepsy).

\paragraph{TUEV.} The TUH EEG Event Corpus (v2.0.1)~\cite{Harati2015_TUEV,Obeid2016_TUH}. Experts mark 1-second events of six classes, each with its time and channels. A sample is a 20-second clip centred on one annotated event and labelled with the class of that event. Each distinct annotated event gives one clip, provided the clip fits inside the recording.

\paragraph{TUSL.} The TUH Slowing Corpus (v2.0.1; 38 patients)~\cite{vonWeltin2017_TUSL,Obeid2016_TUH}. Experts mark 10-second segments of slowing, seizure and background. A sample is a 20-second clip centred on one annotated segment and labelled with its class.

\paragraph{EEGMAT.} The mental-arithmetic dataset of~\cite{Zyma2019_EEGMAT}: 36 subjects, each recorded once at rest and once during mental arithmetic (serial subtraction). Each recording is cut into 20-second clips with no overlap, and a clip is labelled with the condition of its recording. The test set holds the recordings of 3 subjects.

\paragraph{Siena.} The Siena Scalp EEG Database~\cite{Detti2020_Siena}: 41 continuous recordings of 14 adult patients, 141\,h in total, with the start and end times of 47 seizures. Siena is used only for testing. For Table~2 we cut 94 excerpts of 30 minutes each (180 frames). Each of the 47 seizures gives one excerpt that contains the whole seizure and no other seizure, with the seizure onset placed near 20\%, 50\% or 80\% of the excerpt. Seizure durations range from 5 to 151\,s. The 47 seizure-free excerpts come from the same recordings, at least 10 minutes away from every seizure. An excerpt is labelled ``seizure'' if it contains a seizure. The complete recordings are used for main paper Fig.~3 (Appendix~\ref{app:rq3}).

\paragraph{Splits.} Each dataset has a training, a validation and a test set. For TUSZ, these are the official training, development and evaluation sets. For TUAB and TUEV, the test set is the official evaluation set, and a random subset of about 10\% of the training patients forms the validation set. For TUEP and TUSL, which have no official split, and for EEGMAT, we divide the patients at random into the three sets. All splits are made by patient, so no patient appears in two splits; the official TUEV evaluation set is disjoint from its training set. Recordings shorter than 20\,s are discarded. \ours{} is trained on the training split only, on videos of at most 30 minutes (180 frames), and test recordings are evaluated at their full length (Table~\ref{tab:c1_tasks}). The encoder baselines fit their probes on the whole training split and use the validation split for model selection (Appendix~\ref{app:encoder_setup}).

\paragraph{Data of each training stage.} Table~\ref{tab:c0_splits} gives the training samples of each stage. \emph{Mid-training} uses 39,984 examples: each of the 17 classes of the six datasets contributes 2,352 (Appendix~\ref{app:mid_train}). Only the seizure-free TUSZ class has more recordings than that, and 2,352 of its 4,206 recordings are used. \emph{SFT} starts from a class-balanced set of 3,035 training recordings, for which the rollouts are generated (Appendix~\ref{app:post_train}): 300 per class for TUAB, 400 per class for TUSZ, 250 per class for TUEP, 100 per class for TUEV (all 93 for spike and sharp wave), and all TUSL and EEGMAT training clips. \emph{RL} uses 1,418 recordings drawn from the mid-training recordings of the five TUH datasets. In total, \ours{} is trained on 14,979 distinct recordings, the number given in the main paper: the 14,805 mid-training recordings and 174 seizure-free TUSZ recordings of the SFT set that are not among them.

\paragraph{Annotations used only for evaluation.} Training uses only the class label of each sample. For TUEV and TUSL, the annotated event also determines where the clip is cut. The finer annotations are used only in the evaluation of grounding and localization (main paper Table~3 and Figs.~2 and~3; Appendices~\ref{app:grounding} and~\ref{app:exp_counter}). These are the start and end times and channels of TUSZ seizures, the times and channels of TUEV events, and the times of Siena seizures.

\paragraph{Data and ethics.} All recordings come from existing, de-identified EEG datasets that are available to the research community: the TUH corpora~\cite{Obeid2016_TUH}, EEGMAT~\cite{Zyma2019_EEGMAT} and Siena~\cite{Detti2020_Siena}. We collected no new data from human subjects. In the expert review (Appendix~\ref{app:expert}), the expert rated model outputs on recordings of these datasets.

\subsubsection{Implementation Details}\label{app:hparams}

This part completes the implementation details of main paper Sec.~3.1. Table~\ref{tab:c5_ours} gives the training settings of \ours{} for each stage, and Table~\ref{tab:c5_raw} those of the three raw-EEG variants of main paper Table~2 and RQ4. Appendix~\ref{app:adaptation} describes what each stage does, and Appendix~\ref{app:datasets} its data.

\paragraph{Base model and shared settings.}
The base model, \base{}~\cite{Bai2025_Qwen3VL}, has 4.44B parameters: 306.2M in the vision encoder, 109.1M in the visual projection modules (the main merger and three DeepStack mergers) and 4.02B in the language model.
All stages train in bf16 precision with AdamW ($\beta_1{=}0.9$, $\beta_2{=}0.95$, no weight decay), and the gradient norm is clipped at 1.0.
Each micro-batch holds one video of at most 180 frames (30\,min, 66,960 visual tokens; Appendix~\ref{app:eeg_video}).

\begin{table*}[t]
\centering
\footnotesize
\setlength{\tabcolsep}{3pt}
\begin{tabular}{@{}lp{4.75cm}p{4.75cm}p{4.75cm}@{}}
\toprule
 & Mid-training & Post-training: SFT & Post-training: RL\\
\midrule
Initialization & \base{} & mid-trained model (step 2,499) with new LoRA adapters, output matrices at zero & \ours{} (SFT), with its LoRA adapters and projection modules\\
Trained & visual projection modules & projection modules + LoRA adapters & as SFT\\
Frozen & vision encoder, language model & vision encoder, language-model weights & vision encoder, language-model weights; reference $\pi_{\mathrm{SFT}}$\\
Trainable / total & 109.1M / 4.44B & 241.2M (LoRA 132.1M + projection 109.1M) / 4.57B & 241.2M / 4.57B\\
LoRA & -- & rank 64, $\alpha{=}128$, dropout 0.05; q, k, v, o, gate, up and down projections of all 36 language-model layers & as SFT, dropout 0\\
Mode and prompt & non-thinking; question, options, ``Answer with the option letter'' & thinking; rollout prompt & thinking; evaluation prompt\\
Supervised target & answer letter and end-of-turn token (main paper Eq.~1) & refined reasoning trace and answer (main paper Eq.~2) & whole response (main paper Eq.~3); reward = answer correct\\
Training data & 39,984 examples, 2,352 per dataset and class (17 classes), from 14,805 training recordings & 10,356 refined traces & 1,418 training recordings (TUAB, TUSZ, TUEP, TUEV, TUSL), 118--119 per round\\
Rollouts & -- & $K{=}10$ per recording, temperature 1.0, top-$p$ 0.95, $\le$4,096 new tokens; up to $M{=}5$ correct kept & $G{=}8$ per recording: 7 at temperature 1.0 (top-$p$ 1.0) + 1 with the inference decoding; $\le$4,096 new tokens\\
Steps & 2,499 (one epoch) & 1,200 (9,600 traces) & 12 rounds, 74 optimizer steps\\
Evaluated checkpoint & step 2,499 & step 1,200 & policy after round 12\\
Batch per step & 16 videos (4 GPUs $\times$ 4 accumulation) & 8 traces & 64 responses (8 groups), one pass\\
Learning rate & $4\times10^{-5}$; 100 warm-up steps, cosine to 0 at step 2,499 & $1\times10^{-5}$; 100 warm-up steps, cosine to 0 at step 1,294 (one epoch) & $2\times10^{-6}$; 5 warm-up steps, then constant\\
KL weight & -- & -- & $\beta{=}0.04$ to $\pi_{\mathrm{SFT}}$\\
Hardware, time & 4 H100, about 22\,h & 1--2 H200, about 19\,h & 1 H200, about 5\,h including sampling\\
\bottomrule
\end{tabular}
\caption{\textbf{Training settings of \ours{}}, one column per stage. \textbf{Trainable / total}: parameter counts. In SFT and RL the total includes the LoRA adapters, which are merged into the language model after training. The SFT learning-rate schedule is defined over one epoch (1,294 steps); the \textbf{evaluated checkpoint} is step 1,200. Times are training wall-clock times.}
\label{tab:c5_ours}
\end{table*}

\begin{table*}[t]
\centering
\footnotesize
\setlength{\tabcolsep}{3pt}
\begin{tabular}{@{}lp{7.3cm}p{7.3cm}@{}}
\toprule
 & Mid-training & Post-training: SFT\\
\midrule
Initialization & language model of \base{}, frozen EEG encoder, projector at random & mid-trained variant (step 2,499); new LoRA adapters\\
Trained & projector & projector + LoRA adapters (rank 64, $\alpha{=}128$, dropout 0.05, targets as \ours{})\\
Frozen & EEG encoder, language model & EEG encoder, language-model weights\\
Mode and target & non-thinking; answer letter and end-of-turn token & thinking; revised reasoning trace and answer, every token weighted equally\\
Training data & the 39,984 examples of \ours{}'s mid-training & 9,652 revised traces from 2,888 training recordings\\
Steps, evaluated & 2,499 (one epoch); step 2,499 & 1,206 (one epoch); step 1,206\\
Batch per step & 16 (1 GPU $\times$ 16 accumulation) & 8 (1 GPU $\times$ 8)\\
Learning rate & $4\times10^{-5}$; 100 warm-up steps, cosine to 0 at step 2,499 & $1\times10^{-5}$; 100 warm-up steps, cosine to 0 at step 1,206\\
\midrule
REVE + Qwen & \multicolumn{2}{@{}p{14.6cm}@{}}{REVE-Large, 390.0M parameters; 242 tokens of width 1,216 per window (22 referential channels $\times$ 11 patches). Trainable / total: 18.4M / 4.43B in mid-training, 150.5M / 4.56B in post-training. 1 H200: about 17\,h and 8\,h.}\\
CodeBrain + Qwen & \multicolumn{2}{@{}p{14.6cm}@{}}{CodeBrain, 15.2M parameters; 160 tokens of width 200 per window (16 bipolar channels $\times$ 10 one-second patches). Trainable / total: 7.1M / 4.04B and 139.2M / 4.18B. 1 H200: about 10\,h and 6\,h.}\\
LaBraM + Qwen & \multicolumn{2}{@{}p{14.6cm}@{}}{LaBraM-Base, 5.8M parameters; 220 tokens of width 200 per window (22 referential channels $\times$ 10 one-second patches). Trainable / total: 7.1M / 4.04B and 139.2M / 4.17B. Mid-training on 1 H200, about 15\,h; post-training on 1 H100.}\\
\bottomrule
\end{tabular}
\caption{\textbf{Training settings of the raw-EEG variants of RQ4.} The upper part holds for all three variants. The lower part gives, for each variant, its \textbf{encoder}, the number of \textbf{tokens per 10-second window} and the \textbf{trainable / total} parameters in each stage. Mid-training follows the recipe of \ours{}. Post-training is SFT on the reasoning traces of \ours{}, after a revision that removes visual vocabulary (see text).}
\label{tab:c5_raw}
\end{table*}

\paragraph{Raw-EEG variants.}
The raw-EEG variants test whether reasoning self-training extends to a model that reads the raw signal through an EEG encoder instead of a video (main paper RQ4; Appendix~\ref{app:rq4}).
In a variant, the vision encoder and the visual projection modules of \base{} are replaced by a frozen pre-trained EEG encoder and a trainable projector. The encoder is REVE-Large~\cite{ElOuahidi2025_REVE}, CodeBrain~\cite{Ma2025_CodeBrain} or LaBraM-Base~\cite{Jiang2024_LaBraM}.
It turns each 10-second window of the signal into tokens, and the projector maps these tokens to the language model's width of 2,560. The projector has the form of the \texttt{Qwen3-VL} merger: LayerNorm, linear layer, GELU, linear layer, with hidden width $\max(4d, 2{,}560)$ for encoder width $d$.
Each window takes the place of a frame: it enters the language model as its start time in seconds, followed by its projected tokens. The prompt is given in Appendix~\ref{app:exp_base}.

Mid-training uses the examples and settings of \ours{} (Table~\ref{tab:c5_raw}).
Post-training starts from the 10,356 refined traces used in the SFT of \ours{}.
These traces describe frames, so \texttt{Qwen3-32B} (greedy decoding) rewrites each of them: every reference to a panel, frame, cell or colour is replaced by the signal property it stands for.
A revision is accepted if four conditions hold: the answer is unchanged; no visual vocabulary remains; it adds no clinical term or number that is absent from the original; and it keeps at least 80\% of the original's recording times and at least 80\% of its scalp regions.
Post-training of the variants consists of SFT on the accepted revisions.
At test time, a variant reads all windows of a recording in one query, with the decoding and the answer extractor of \ours{} (Appendix~\ref{app:adaptation}).

\subsubsection{Baselines Details}\label{app:exp_base}

Main paper Table~2 compares \ours{} with text-generating models and with EEG encoders. This part states, for each of them, what it receives, how it is decoded and how its outputs become one answer per recording. Every model receives all 10-second windows of a test recording, in its own input format. We describe the text-generating models first (Table~\ref{tab:c2_protocol}) and the EEG encoders with linear probes second (Table~\ref{tab:c4_encoders}). Appendix~\ref{app:final_answer} shows examples of the outputs.

\begin{table}[t]\centering\footnotesize
\caption{\textbf{Input and decoding of each text-generating model} of main paper Table~2. The models in the first six rows read the \textbf{whole recording in one query} and answer in free text. The last two read \textbf{one segment per query}, and their answer is derived from the probabilities they give to the option letters. Each baseline uses the decoding settings of its model card ($T$: temperature, $p$: top-$p$, $k$: top-$k$; penalties: repetition / frequency). The prompts are given in the text.}\label{tab:c2_protocol}
\setlength{\tabcolsep}{3pt}
\begin{tabular}{@{}lll@{}}\toprule
Model & Input per query & Decoding\\\midrule
\ours{} & Video & Greedy, penalties 1.1 / 0.5\\
Raw-EEG variants & Encoder tokens & Greedy, penalties 1.1 / 0.5\\
\base{} & Same video & $T$ 1.0, $p$ 0.95, $k$ 20\\
OctoMed-7B & Same video & $T$ 0.6, $p$ 0.95\\
MedGemma-1.5-4B & Same frames, as images & Greedy\\
MedGemma-27B & Same frames, as images & Greedy\\\midrule
CerebraGloss-3B & Own plot, one window & Option probabilities\\
NeuroLM-B & Raw EEG, one segment & Option probabilities\\\bottomrule
\end{tabular}\end{table}

\paragraph{Text-generating models: protocol.} The models are run in one of two ways (Table~\ref{tab:c2_protocol}).

\emph{Whole recording in one query.} \ours{}, the raw-EEG variants (REVE + Qwen, CodeBrain + Qwen, LaBraM + Qwen) and the zero-shot VLMs (\base{}, OctoMed-7B~\cite{Ossowski2026_OctoMed}, MedGemma-1.5-4B~\cite{Sellergren2026_MedGemma15}, MedGemma-27B~\cite{Sellergren2025_MedGemma}) read the whole recording in one query and answer in free text. The extractor of Appendix~\ref{app:adaptation} maps this answer to one option. The zero-shot VLMs are the released models, applied without any training and decoded with the settings of their model cards.

\emph{One segment per query.} CerebraGloss-3B and NeuroLM-B read one segment per query. For each segment, we read the probability that the model gives to each option letter as its next token. The answer for the recording is the option with the highest geometric mean of these probabilities over the segments. This is the same rule as for the encoder probes (Appendix~\ref{app:encoder_setup}). CerebraGloss-3B is the released model. For NeuroLM-B, the instruction-tuned checkpoint is not publicly available. We therefore start from the released NeuroLM-B checkpoint, which precedes instruction tuning, and apply the instruction-tuning recipe of its paper (five epochs, answers as targets) to the 14,805 mid-training recordings of the six training sets.

Siena is evaluated in the same way on its 94 excerpts, using the TUSZ question. The NeuroNarrator row of Table~2 gives the values reported in its paper.

\paragraph{Text-generating models: prompts.}
Each model receives the task question in its recommended prompt format. The prompts below are for the \textbf{TUSZ seizure question}. For the other tasks, only the question and its options change (Table~\ref{tab:b2_questions}).

\textbf{\ours{} and \base{}} receive the video and the evaluation prompt of Appendix~\ref{app:adaptation}: the frame description, the question and its options.

\textbf{OctoMed-7B} receives the same video and prompt. The prompt ends with the sentence that its model card prescribes for multiple-choice questions:
\begin{promptbox}
Please reason step-by-step, and put your final answer within \textbackslash boxed\{\}.
\end{promptbox}

\textbf{MedGemma-1.5-4B and MedGemma-27B} take images as input, so they receive the same frames as separate images, each preceded by its start time. Each image is padded to a square, because MedGemma resizes images to $896\times896$ pixels. The system turn is the one of the model card (``You are a helpful medical assistant.''). The frame description is reworded for images and followed by the question and its options:
\begin{promptbox}
The images show the EEG in time order, one image per 10-second window, each preceded by its start time. Each image has four panels, top to bottom: TRACES (EEG by scalp chain), POWER (band power cells for the left and right hemisphere), MAPS (scalp maps for 1-8 Hz and 8-30 Hz, nose up, left on the left), SHARP (sharpness cells by scalp chain, brighter = sharper). The time axis at the bottom gives recording seconds.
\end{promptbox}

\textbf{The raw-EEG variants} receive, for each 10-second window, its start time followed by the projected encoder tokens (Appendix~\ref{app:hparams}). The frame description is replaced by one sentence, followed by the question and its options. The sentence for REVE is shown below; for each encoder, the bracket gives its own channels and patches:
\begin{promptbox}
The EEG is given as one block of encoder tokens per 10-second window (22 scalp channels x 11 patches), each block preceded by its start time in recording seconds.
\end{promptbox}

\textbf{CerebraGloss-3B} receives one 10-second window per query, drawn as a plot by its released preprocessing and plotting code. The question uses its own multiple-choice format, and for seizures also its own wording:
\begin{promptbox}
Does this EEG recording show evidence of seizure activity?

A) Yes

B) No
\end{promptbox}

\textbf{NeuroLM-B} receives the raw EEG of one segment per query, in its own input format. A segment is one 10-second window for TUAB, TUSZ and TUEP, and the whole 20-second clip for TUEV, TUSL and EEGMAT. The text gives the time span of the segment, then the question and options of Table~\ref{tab:b2_questions} in the format of its instruction tuning:
\begin{promptbox}
Clip from [start] to [end] s of the recording.\\
Does this EEG recording contain an electrographic seizure?\\
Options:\\
A. no seizure\\
B. seizure\\
Answer with the option letter.\\
Answer:
\end{promptbox}

\label{app:encoder_setup}
\paragraph{EEG encoders: overview.} The rows ``EEG encoders (linear probes)'' of main paper Table~2 evaluate six pre-trained EEG foundation models. Each encoder is frozen and used only to extract features. It turns every 10-second window into one feature vector, and one linear probe (a linear classifier) per task is trained on these vectors. All six encoders follow the same protocol; they differ only in their checkpoint, input format and pooling (Table~\ref{tab:c4_encoders}). Each encoder uses its released pre-trained weights, unchanged. For BIOT we use the checkpoint pre-trained without labels (PREST)~\cite{Yang2023_BIOT}, so that no encoder has seen the labels of our evaluation corpora. Appendix~\ref{app:encoder_ft} reports the same encoders fine-tuned end to end.

\begin{table*}[t]\centering\footnotesize
\caption{\textbf{Input and pooling of the six frozen encoders.} Every encoder receives the same 10-second windows sampled at 200\,Hz (EEGPT after resampling to 250\,Hz). It returns one \textbf{feature vector per window}, whose dimension is given in brackets. Bipolar: 16 channels, each the difference between two neighbouring electrodes (the longitudinal bipolar montage, Fp1--F7 \dots\ P4--O2). Referential: the 22 single electrodes of Appendix~\ref{app:eeg_video}; electrodes that were not recorded are set to zero. Clipped: values limited to $\pm$500\,$\mu$V. The last column lists which corpora of our seven evaluations are part of each encoder's \textbf{pre-training data}, as stated in its paper. Pre-training uses no labels. TUEG is the TUH EEG Corpus~\cite{Obeid2016_TUH}; the five TUH corpora are subsets of it.}\label{tab:c4_encoders}
\begin{tabular}{@{}lp{2.3cm}p{2.7cm}p{2.2cm}p{3.7cm}p{2.2cm}@{}}\toprule
Model & \raggedright Checkpoint & \raggedright Channels & \raggedright Scaling & \raggedright Feature per 10-s window & \raggedright Our corpora in pre-training \tabularnewline\midrule
BIOT~\cite{Yang2023_BIOT} & \raggedright PREST, 16 channels (no labels) & \raggedright 16 bipolar & \raggedright each channel divided by its 95th percentile of $|x|$ & \raggedright mean over all tokens (256) & \raggedright none \tabularnewline
EEGPT~\cite{Wang2024_EEGPT} & \raggedright released weights (braindecode port) & \raggedright 20 referential (no A1, A2), average reference & \raggedright clipped; mV & \raggedright two 4-s crops (first 8\,s); per crop, mean over patches of the 4 summary tokens; crops averaged (2,048) & \raggedright none \tabularnewline
LaBraM~\cite{Jiang2024_LaBraM} & \raggedright LaBraM-Base & \raggedright 22 referential & \raggedright clipped; $\mu$V/100 & \raggedright mean of the patch tokens, then LayerNorm, as in LaBraM's classification head (200) & \raggedright TUEP, TUSZ, TUSL, Siena \tabularnewline
CBraMod~\cite{Wang2025_CBraMod} & \raggedright released weights & \raggedright 16 bipolar & \raggedright clipped; $\mu$V/100 & \raggedright mean over channels and 1-s patches (200) & \raggedright TUEG \tabularnewline
REVE~\cite{ElOuahidi2025_REVE} & \raggedright REVE-Large & \raggedright 22 referential, with 3D electrode positions & \raggedright clipped; $\mu$V/100 & \raggedright attention pooling with REVE's pre-trained query token (1,216) & \raggedright TUEG, Siena \tabularnewline
CodeBrain~\cite{Ma2025_CodeBrain} & \raggedright released encoder (EEGSSM) & \raggedright 16 bipolar & \raggedright clipped; $\mu$V/100 & \raggedright mean over channels and 1-s patches (200) & \raggedright TUEG \tabularnewline\bottomrule
\end{tabular}\end{table*}

\paragraph{EEG encoders: input and features.} The encoders receive the same preprocessed signal from which the videos of \ours{} are rendered (22 referential channels at 200\,Hz; Appendix~\ref{app:eeg_video}), cut into the same 10-second windows. On the test sets, the encoders see exactly the windows that \ours{} sees as frames. Each encoder builds its own input from a window and returns one feature vector (Table~\ref{tab:c4_encoders}). A recording of $n$ windows thus gives $n$ vectors, and a 20-second clip of TUEV, TUSL or EEGMAT gives two. Electrodes that a recording lacks are set to zero for the referential encoders: Fpz in every test recording of the TUH corpora and EEGMAT, and Fpz, A1 and A2 in Siena, where all 16 bipolar channels are available.

\paragraph{EEG encoders: linear probe.} For each encoder and task, the probe is one logistic-regression classifier with L2 regularisation (solver L-BFGS, at most 3,000 iterations). Its inputs are the features, standardised with the mean and standard deviation of the training windows, and class weights are inversely proportional to class frequency. The probe is trained on every window of every training recording, and each window takes the label of its recording. The 16,659 training recordings of the six tasks include all recordings used to train \ours{}. The inverse regularisation strength $C\in\{0.01, 0.1, 1, 10\}$ is chosen per encoder and task on the validation split, by recording-level balanced accuracy under the decision rule below. Test recordings are used only for the final score.

\paragraph{EEG encoders: one answer per recording.} The probe gives one output per window, but each task needs one answer per recording. The recording's answer is the class with the highest geometric mean of its window probabilities. Formally, the probe gives class probabilities $p_k(w)$ for each window $w$ of a recording with $n$ windows, and
\begin{equation*}
\hat{y}=\arg\max_k \tfrac{1}{n}\textstyle\sum_{w}\log p_k(w).
\end{equation*}
This rule has no tuned parameter and is the same for every encoder and task. The metric is balanced accuracy (Appendix~\ref{app:datasets}) on the same 3,875 test recordings as for \ours{}. Appendix~\ref{app:decision_rule} adds, for seizure detection, a rule chosen on validation data.

\paragraph{EEG encoders: Siena.} No probe is trained or tuned on Siena. Instead, each encoder's TUSZ probe, with the $C$ chosen on the TUSZ validation split, is applied to all 180 windows of each 30-minute excerpt, and the rule above gives the excerpt's answer. The score is the balanced accuracy over the same 94 excerpts given to \ours{} (Appendix~\ref{app:datasets}).

\paragraph{EEG encoders: pre-training overlap.} Four of the released encoders (LaBraM, CBraMod, REVE and CodeBrain) were pre-trained by their authors, without labels, on data that include corpora of our evaluations (Table~\ref{tab:c4_encoders}, last column); for LaBraM and REVE, these include unlabelled Siena recordings. No label of our evaluations is used in these pre-trainings, and no training performed in this work uses Siena. BIOT (PREST) and EEGPT were pre-trained on other data.

\subsection{Results Details (main paper Sec.~3.2, Table~2)}\label{app:exp_results}

This part supports main paper Table~2 in three steps. Appendix~\ref{app:tasks} shows what \ours{} outputs on each task. Appendix~\ref{app:final_answer} compares these outputs with those of the text-generating baselines on the same input. Appendices~\ref{app:encoder_ft}--\ref{app:metrics} then test whether the comparison with the EEG encoders is robust to three changes: fine-tuning the encoders, a decision rule suited to brief seizures, and additional metrics. The conclusions of Table~2 hold in each case. \ours{} keeps the highest mean balanced accuracy when the encoders are fine-tuned. It stays above every encoder on TUSZ and Siena under the alternative decision rule. After RL, it remains the best model on TUSZ, TUEV and TUSL under the additional metrics.

\subsubsection{Examples of each task}\label{app:tasks}

Main paper Table~2 scores only the final answer. The examples of this part show the complete output of \ours{} (RL) behind such an answer, for one test sample of each task.

\paragraph{Reading the example figures.} All example figures of Appendices~\ref{app:experiments} and~\ref{app:findings} follow the same conventions. Each figure shows the signal, the question and the model outputs, verbatim. The output of \ours{} is complete unless the caption says otherwise. Long outputs of other models are shortened, and the number of omitted words is printed. \emph{In the signal}, the tint and the dark bar mark the annotated event, and the purple bar marks the interval the model states. Blue boxes mark the regions that the model names and that are annotated. \emph{In the outputs}, every statement that the expert annotation can confirm or contradict is coloured: green for a confirmed time, blue for a confirmed scalp region, and red for a contradicted statement. Grey notes give the annotated value. Unmarked statements cannot be checked against the annotation. A time is confirmed if it lies within the tolerance of the matching metric (10\,s of an annotated seizure, 2\,s of an annotated TUEV event), or if the frame it cites overlaps the event. A region is confirmed if it is annotated at that time; regions are compared at the level of the five scalp regions. Siena has no channel annotation, so only times are marked there. Outputs on TUAB, TUEP, TUSL and EEGMAT are not marked; the figures label them ``no event annotation''. The examples are selected, and the aggregate results are those of the main-paper tables. Apart from the failure cases, no checkable statement of \ours{} in these examples is contradicted.

\paragraph{One example per task.} Figures~\ref{fig:c1_c1_1}--\ref{fig:c1_c1_7} each show one correctly answered test sample, in the order of Table~\ref{tab:c1_tasks}. Three are \emph{whole recordings}. On TUAB (Fig.~\ref{fig:c1_c1_1}), the model classifies a 19-minute recording as abnormal from the whole video in one pass. On TUSZ (Fig.~\ref{fig:c1_c1_2}), it answers ``seizure'' and cites a time interval and scalp regions that the annotation confirms. On TUEP (Fig.~\ref{fig:c1_c1_3}), it gives a diagnosis about the patient and points to times and regions with large signal deflections. Three are \emph{20-second clips}. On TUEV (Fig.~\ref{fig:c1_c1_4}), the model recognizes generalized periodic discharges and places them in the frame and the regions where they are annotated. On TUSL (Fig.~\ref{fig:c1_c1_5}), it classifies a segment among three classes. On EEGMAT (Fig.~\ref{fig:c1_c1_6}), where there is no event to point at, it argues from the overall amplitude and the band power. The last example is from Siena, the \emph{dataset never used in training} (Fig.~\ref{fig:c1_c1_7}): the model finds a 70-second seizure in a 30-minute excerpt and cites an interval inside the annotated seizure.

\paragraph{Failure case.} Figure~\ref{fig:c1_c1_8} shows a wrong final answer on TUSZ: the model misses a 12-second seizure in a 17.7-minute recording and answers ``no seizure''.


\subsubsection{Qualitative comparison with the text-generating baselines}\label{app:final_answer}

Balanced accuracy does not show how the answers of the models differ. This part therefore places the output of \ours{} (RL) next to the outputs of the text-generating baselines on the same input. The baselines are run as described in Appendix~\ref{app:exp_base}.

\paragraph{The same question to every model.} Figure~\ref{fig:c2_C2-1} gives the seizure question of main paper Table~2 to \ours{}, to the zero-shot VLMs, and to the prior EEG-LLMs CerebraGloss-3B and NeuroLM-B, on one TUSZ recording with a seizure. \ours{} cites the time and the regions of the annotated seizure. The baselines cite other times or no time, or answer ``no seizure''. Figure~\ref{fig:c2_C2-2} repeats the comparison on a 60-minute Siena video, twice the longest training input, with the localization question of main paper Fig.~3(b) (Appendix~\ref{app:rq3}). \ours{} states the onset and the end of a 14-second seizure within 4 seconds of the annotation. Each baseline states a time unrelated to the seizure.

\paragraph{Per-window models.} CerebraGloss-3B and NeuroLM-B read a recording one 10-second window per query (Table~\ref{tab:c2_protocol}), whereas \ours{} reads it in one pass. Figures~\ref{fig:c2_C2-3} and~\ref{fig:c2_C2-4} show what this difference means for the output. In Fig.~\ref{fig:c2_C2-3}, a seizure-free recording yields one answer per window from each of the two models, and these answers are then combined into an answer for the recording. \ours{} gives a single answer with its reasoning. In Fig.~\ref{fig:c2_C2-4}, CerebraGloss-3B is asked for the onset and the end of a seizure. It states times relative to its own 10-second window, whereas \ours{} states one interval in recording seconds.

\paragraph{Failure case.} Figure~\ref{fig:c2_C2-5} shows a correct ``no seizure'' answer on a Siena excerpt whose reasoning states a wrong time range for the frames it describes.


\subsubsection{Fine-tuned EEG encoders}\label{app:encoder_ft}

In main paper Table~2, each EEG encoder is frozen and evaluated with a linear probe (Appendix~\ref{app:encoder_setup}). To test whether the comparison depends on this choice, we also fine-tune each of the six encoders end to end on each task, with the recipe of its own paper or released code. Table~\ref{tab:c4b_results} reports the results and Table~\ref{tab:c4b_recipes} the recipes.

\begin{table*}[t]\centering\footnotesize
\caption{\textbf{Fine-tuned encoders versus frozen probes and \ours{}.} Balanced accuracy (\%) of each encoder in two settings: \textbf{frozen}, with one linear probe per task (the values of main paper Table~2), and \textbf{fine-tuned} end to end with its own recipe. Scores are on the test sets of Table~2, and on Siena with each encoder's TUSZ model. The \ours{} and base-model rows are from Table~2. \textbf{Mean} is the unweighted average over the seven evaluations; the best result in each column is in bold. The highlighted rows are \ours{} and its base model, \base{}. As in Table~2, the small {\tiny\textcolor[HTML]{2E8B47}{+x.x}} values give the gain of \ours{} over \base{}. On the binary tasks, 50.0 is the chance level.}\label{tab:c4b_results}
\begin{tabular}{@{}llccccccc>{\columncolor[HTML]{EAF1FB}}c@{}}\toprule
\textbf{Model} & \textbf{Encoder} & TUAB & TUSZ & TUEP & TUEV & EEGMAT & TUSL & Siena & \textbf{Mean}\\\midrule
BIOT~\cite{Yang2023_BIOT} & frozen (probe) & 82.4 & 52.2 & 84.7 & 36.3 & 51.9 & 46.4 & 51.1 & 57.9\\
 & fine-tuned & 84.4 & 61.0 & 86.1 & 44.4 & 77.8 & 51.5 & 42.6 & 64.0\\
\rule{0pt}{2.8ex}EEGPT~\cite{Wang2024_EEGPT} & frozen (probe) & \textbf{86.1} & 53.9 & 70.1 & 40.7 & 64.8 & 54.5 & 48.9 & 59.9\\
 & fine-tuned & 81.8 & 57.5 & 72.2 & 37.4 & 64.8 & 70.0 & 53.2 & 62.4\\
\rule{0pt}{2.8ex}LaBraM~\cite{Jiang2024_LaBraM} & frozen (probe) & 81.8 & 55.6 & 71.5 & 44.3 & \textbf{81.5} & 59.6 & 48.9 & 63.3\\
 & fine-tuned & 82.7 & 55.7 & 73.6 & \textbf{58.7} & 74.1 & 61.0 & 53.2 & 65.6\\
\rule{0pt}{2.8ex}CBraMod~\cite{Wang2025_CBraMod} & frozen (probe) & 81.3 & 60.2 & \textbf{86.8} & 44.2 & 53.7 & 63.6 & 55.3 & 63.6\\
 & fine-tuned & 84.1 & 52.6 & 60.4 & 49.6 & 57.4 & 51.4 & 50.0 & 57.9\\
\rule{0pt}{2.8ex}REVE~\cite{ElOuahidi2025_REVE} & frozen (probe) & 84.3 & 59.2 & 73.6 & 41.2 & 66.7 & 72.4 & 48.9 & 63.8\\
 & fine-tuned & 85.6 & 53.6 & 63.2 & 42.9 & 57.4 & 54.1 & 50.0 & 58.1\\
\rule{0pt}{2.8ex}CodeBrain~\cite{Ma2025_CodeBrain} & frozen (probe) & 84.3 & 58.5 & 84.0 & 52.6 & 72.2 & 58.1 & 57.4 & 66.7\\
 & fine-tuned & 85.6 & 52.3 & 72.2 & 49.9 & 50.0 & 52.3 & 50.0 & 58.9\\\midrule
\rowcolor[HTML]{FFF4D6}\base{} (base model) & -- & 51.9 & 47.8 & 51.4 & 18.6 & 37.0 & 29.6 & 48.9 & 40.7\\\midrule
\rowcolor[HTML]{FFF4D6}\textbf{\ours{} (SFT)} & -- & 77.2\,{\tiny\textcolor[HTML]{2E8B47}{+25.3}} & 79.0\,{\tiny\textcolor[HTML]{2E8B47}{+31.2}} & 56.9\,{\tiny\textcolor[HTML]{2E8B47}{+5.5}} & 55.2\,{\tiny\textcolor[HTML]{2E8B47}{+36.6}} & 75.9\,{\tiny\textcolor[HTML]{2E8B47}{+38.9}} & 62.8\,{\tiny\textcolor[HTML]{2E8B47}{+33.2}} & \textbf{81.9}\,{\tiny\textcolor[HTML]{2E8B47}{+33.0}} & 69.8\,{\tiny\textcolor[HTML]{2E8B47}{+29.1}}\\
\rowcolor[HTML]{FFF4D6}\textbf{\ours{} (RL)} & -- & 81.0\,{\tiny\textcolor[HTML]{2E8B47}{+29.1}} & \textbf{86.2}\,{\tiny\textcolor[HTML]{2E8B47}{+38.4}} & 70.8\,{\tiny\textcolor[HTML]{2E8B47}{+19.4}} & 55.9\,{\tiny\textcolor[HTML]{2E8B47}{+37.3}} & 63.0\,{\tiny\textcolor[HTML]{2E8B47}{+26.0}} & \textbf{74.0}\,{\tiny\textcolor[HTML]{2E8B47}{+44.4}} & \textbf{81.9}\,{\tiny\textcolor[HTML]{2E8B47}{+33.0}} & \textbf{73.2}\,{\tiny\textcolor[HTML]{2E8B47}{+32.5}}\\\bottomrule
\end{tabular}\end{table*}

\paragraph{Protocol shared with the probes.} Fine-tuning keeps the following from the probes: the pre-trained checkpoints, the training, validation and test recordings, the 10-second windows, the window labels (each window takes the label of its recording) and the input formats (Table~\ref{tab:c4_encoders}). The one exception is EEGPT, which follows its released fine-tuning code: it receives the whole 10-second window at 200\,Hz in $\mu$V/100, and a learned convolution maps our 22 referential channels to its 20 inputs. One model is trained per encoder and task. As for the probes, a recording's answer is the class with the highest geometric mean of its window probabilities, and the metric is balanced accuracy on the same 3,875 test recordings. Siena is used only for testing: each encoder's TUSZ model is applied to all 180 windows of each of the 94 excerpts.

\paragraph{Fine-tuning recipes.} All encoder weights are trained together with a task head, using the recipe of the encoder's paper or code (Table~\ref{tab:c4b_recipes}). TUSZ, TUEP and TUSL have no fine-tuning settings in the six papers. We therefore use each encoder's TUAB settings for TUAB, TUSZ and TUEP, and its TUEV settings for TUEV and TUSL. EEGMAT uses the encoder's mental-arithmetic settings, if it has any. As for the probes, the loss weights each class inversely to its frequency. We test the epoch with the highest recording-level balanced accuracy on the validation split; for REVE, which first trains the head and then the whole model, this is the best epoch of the second stage, as in its code. The same score decides early stopping (BIOT, REVE) and when REVE halves its learning rate. Test recordings are used only for the final score.

\begin{table*}[t]\centering\footnotesize
\caption{\textbf{Fine-tuning recipes as run.} TUAB-type tasks: TUAB, TUSZ, TUEP. TUEV-type tasks: TUEV, TUSL. $B$: batch size for TUAB-type tasks / TUEV / EEGMAT and TUSL. In every row, all encoder weights are trained, the loss is \textbf{class-weighted}, and we test the epoch with the highest recording-level balanced accuracy on validation. BCE: binary cross-entropy; CE: cross-entropy; LS: label smoothing; wd: weight decay.}\label{tab:c4b_recipes}
\setlength{\tabcolsep}{3pt}
\begin{tabular}{@{}lp{3.5cm}p{3.9cm}p{3.9cm}p{1.6cm}p{2.2cm}@{}}\toprule
Model & \raggedright Head & \raggedright Optimiser, learning rate & \raggedright Schedule, epochs & \raggedright $B$ & \raggedright Loss, other \tabularnewline\midrule
BIOT & \raggedright mean over tokens, ELU, linear & \raggedright Adam, $10^{-3}$, L2 $10^{-5}$ & \raggedright constant; $\le$100 epochs, stop after 5 without gain (ran 6--14) & \raggedright 512 / 128 / 16 & \raggedright BCE / CE; fp32 \tabularnewline\addlinespace
EEGPT & \raggedright channel and temporal convolutions before the encoder; all summary tokens flattened, dropout, linear & \raggedright AdamW, $5\times10^{-4}$ for all layers, wd 0.05 & \raggedright 5-epoch linear warm-up, cosine to $10^{-6}$; 50 epochs (TUEV-type 30) & \raggedright 200 / 256 / 16 & \raggedright BCE / CE, LS 0.1; bf16 \tabularnewline\addlinespace
LaBraM & \raggedright mean of patch tokens, LayerNorm, linear & \raggedright AdamW, $5\times10^{-4}$, layer-wise decay 0.65, wd 0.05; drop path 0.1 & \raggedright 5-epoch linear warm-up, cosine to $10^{-6}$; 50 epochs & \raggedright 512 / 256 / 16 & \raggedright BCE / CE, LS 0.1; bf16 \tabularnewline\addlinespace
CBraMod & \raggedright all patch representations flattened, three-layer MLP, dropout 0.1 & \raggedright AdamW, encoder $10^{-4}$, head $10^{-3}\sqrt{B/256}$, wd 0.05 & \raggedright cosine to $10^{-6}$; 50 epochs & \raggedright 64 / 64 / 16 & \raggedright BCE / CE, LS 0.1; clip 1; fp32 \tabularnewline\addlinespace
CodeBrain & \raggedright as CBraMod; dropout 0.4 / 0.3 / 0.1 (TUAB-type / TUEV-type / EEGMAT) & \raggedright AdamW, one rate; lr and wd $10^{-5}$, $5\times10^{-5}$ (TUAB-type); $2\times10^{-5}$, $5\times10^{-4}$ (TUEV-type); $3\times10^{-5}$, $10^{-3}$ (EEGMAT) & \raggedright cosine to $10^{-6}$; 50 epochs & \raggedright 512 / 64 / 16 & \raggedright BCE / CE, LS 0.1; clip 5; fp32 \tabularnewline\addlinespace
REVE & \raggedright pooled token and all patch tokens flattened, RMSNorm, dropout, linear (EEGMAT: pooled token only) & \raggedright StableAdamW, $\beta{=}(0.92, 0.999)$, wd 0.01 (TUEV-type probing 0.1); probing $2\times10^{-4}$; full $10^{-4}$ (EEGMAT $5\times10^{-5}$) & \raggedright linear probing, then full fine-tuning; exponential warm-up, halving on plateau, early stop (patience 5 / 10, EEGMAT 5 / 6); at most 10\,+\,20 (TUAB-type), 20\,+\,50 (TUEV-type), 20\,+\,200 (EEGMAT) epochs & \raggedright 64 / 256 / 16 & \raggedright CE, mixup, per-stage dropout; clip 2; bf16 \tabularnewline\bottomrule
\end{tabular}\end{table*}

\paragraph{\ours{} keeps the highest Mean against fine-tuned encoders.} Fine-tuning raises the Mean over the frozen probe for BIOT (+6.1), EEGPT (+2.5) and LaBraM (+2.3), and lowers it for CBraMod ($-5.7$), REVE ($-5.7$) and CodeBrain ($-7.8$). The best fine-tuned Mean, 65.6 (LaBraM), is below the best probe (CodeBrain, 66.7). It is 4.2 points below \ours{} after SFT and 7.6 points below it after RL. On TUSZ and Siena, every encoder row, frozen or fine-tuned, is below both \ours{} rows, and on TUSL every encoder row is below \ours{} (RL). As in Table~2, individual encoders are higher on single tasks: on TUAB, TUEP and EEGMAT, and, once fine-tuned, LaBraM on TUEV.


\subsubsection{Decision rule for seizure detection with window-level models}\label{app:decision_rule}

In main paper Table~2, the window outputs of each EEG encoder become one answer per recording through the geometric mean of the window probabilities (Appendix~\ref{app:encoder_setup}). This rule suits labels that hold for the whole recording. A seizure, however, is a brief event. In the 165 TUSZ test recordings with a seizure, the seizure covers a median of 14.7\% of the recording, so averaging over all windows tends to miss it. We therefore test whether the lead of \ours{} in seizure detection depends on this rule. The encoders are scored again with a rule suited to brief events, which uses no supervision beyond the recording labels (Table~\ref{tab:c4c_rule}). The rule of Table~2 remains the primary result.

\paragraph{Validation-chosen rule.} The rule answers ``seizure'' when a summary of the window probabilities, such as their maximum, exceeds a threshold; the summary and the threshold are chosen on validation data. Formally, let $p_w$ be the probability of ``seizure'' for window $w$. A recording is answered ``seizure'' if $s(\{p_w\})>\theta$. Here $s$ is one of five statistics of the recording's window probabilities (maximum, mean of the three highest, 95th percentile, 90th percentile, mean), and $\theta\in\{0.02, 0.04, \dots, 0.98\}$. For each model, we choose the pair $(s,\theta)$ with the highest recording-level balanced accuracy on the TUSZ validation split (1,872 recordings), using only their recording labels. Ties go to the earlier statistic in the list above, then to the smaller $\theta$. The chosen pair is applied unchanged to the 875 TUSZ test recordings and to the 94 Siena 30-minute excerpts of Table~2. Both the frozen probes and the fine-tuned models of Table~\ref{tab:c4b_results} are scored this way. \ours{} gives one answer for the whole recording in one pass, so no rule applies to it.

\begin{table}[t]\centering\footnotesize
\caption{\textbf{With a rule suited to brief seizures, the encoders gain on TUSZ and stay below \ours{}.} Balanced accuracy (\%) on the TUSZ test set and on the 94 Siena 30-minute excerpts. \textbf{Avg.}: the rule of main paper Table~2, the geometric mean of the window probabilities (the frozen rows repeat the Table~2 values). \textbf{Val.}: the statistic and threshold chosen on the TUSZ validation labels. \ours{} gives one answer per recording without any rule, so one value spans both columns. The best value per column is in bold.}\label{tab:c4c_rule}
\setlength{\tabcolsep}{3.5pt}
\begin{tabular}{@{}llcccc@{}}\toprule
 & & \multicolumn{2}{c}{TUSZ} & \multicolumn{2}{c}{Siena}\\\cmidrule(lr){3-4}\cmidrule(l){5-6}
Model & Encoder & Avg. & Val. & Avg. & Val.\\\midrule
BIOT~\cite{Yang2023_BIOT} & frozen & 52.2 & 67.4 & 51.1 & 50.0\\
 & fine-tuned & 61.0 & 68.3 & 42.6 & 48.9\\\addlinespace[1pt]
EEGPT~\cite{Wang2024_EEGPT} & frozen & 53.9 & 71.1 & 48.9 & 61.7\\
 & fine-tuned & 57.5 & 69.2 & 53.2 & 52.1\\\addlinespace[1pt]
LaBraM~\cite{Jiang2024_LaBraM} & frozen & 55.6 & 71.0 & 48.9 & 58.5\\
 & fine-tuned & 55.7 & 75.1 & 53.2 & 70.2\\\addlinespace[1pt]
CBraMod~\cite{Wang2025_CBraMod} & frozen & 60.2 & 76.9 & 55.3 & 69.1\\
 & fine-tuned & 52.6 & 75.4 & 50.0 & 63.8\\\addlinespace[1pt]
REVE~\cite{ElOuahidi2025_REVE} & frozen & 59.2 & 75.1 & 48.9 & 68.1\\
 & fine-tuned & 53.6 & 69.9 & 50.0 & 57.4\\\addlinespace[1pt]
CodeBrain~\cite{Ma2025_CodeBrain} & frozen & 58.5 & 71.4 & 57.4 & 62.8\\
 & fine-tuned & 52.3 & 61.0 & 50.0 & 52.1\\\midrule
\multicolumn{2}{@{}l}{\ours{} (SFT)} & \multicolumn{2}{c}{79.0} & \multicolumn{2}{c}{\textbf{81.9}}\\
\multicolumn{2}{@{}l}{\ours{} (RL)} & \multicolumn{2}{c}{\textbf{86.2}} & \multicolumn{2}{c}{\textbf{81.9}}\\\bottomrule
\end{tabular}\end{table}

\paragraph{\ours{} stays ahead under both rules.} The validation-chosen rule helps the encoders: every encoder row gains on TUSZ, and on Siena the change ranges from $-1.1$ to $+19.2$ points. \ours{} stays above every encoder row under both rules on both datasets. On TUSZ, the best encoder row under the validation-chosen rule is frozen CBraMod (76.9). \ours{} leads it by 2.1 points after SFT (79.0) and by 9.3 points after RL (86.2). Under averaging, the leads over the best encoder row are 18.0 and 25.2 points. On Siena, the best encoder row under the validation-chosen rule is fine-tuned LaBraM (70.2). \ours{} (81.9 after both SFT and RL) leads it by 11.7 points. Under averaging, the lead over the best encoder row is 24.5 points.


\subsubsection{Additional metrics}\label{app:metrics}

Main paper Table~2 reports only balanced accuracy. To check that its conclusions do not depend on this metric, we report four further metrics for the same models on the same held-out test sets: Cohen's $\kappa$ and weighted F1 on the two multi-class tasks (Table~\ref{tab:c6_multi}), and AUROC and AUC-PR on the four binary tasks (Table~\ref{tab:c6_auc}).

\paragraph{Definitions.} Cohen's $\kappa$ and weighted F1 are computed from the final answers of Table~2, and an unanswered question counts as wrong. Cohen's $\kappa$ measures agreement with the labels beyond chance: 0 means chance level and 100 perfect agreement. Weighted F1 is the per-class F1, averaged with weights proportional to class frequency. AUROC and AUC-PR (average precision) are computed from a score per recording instead of a final answer. For the encoder probes, this score is the probability of the positive class. For NeuroLM-B, it is the normalized likelihood of the positive option. For CerebraGloss-3B, it is the difference between the mean log-probabilities of the two options over its windows. For the other text-generating models, it is the probability of the positive option letter after the model's own reasoning. NeuroNarrator is omitted, because its results are taken from its paper.

\begin{table}[t]
\centering
\footnotesize
\caption{\textbf{Agreement on the multi-class tasks.} \textbf{Cohen's $\kappa$} and \textbf{weighted F1} (wF1), in \%, computed from the final answers of main paper Table~2 on TUEV (six classes) and TUSL (three classes). The best value per column is in bold and the second best is underlined. The highlighted rows are \ours{} and its base model, \base{}. As in Table~2, the small {\tiny\textcolor[HTML]{2E8B47}{+x.x}} values give the gain of \ours{} over \base{}.}
\label{tab:c6_multi}
\setlength{\tabcolsep}{3.5pt}
\begin{tabular}{@{}lcccc@{}}
\toprule
 & \multicolumn{2}{c}{TUEV} & \multicolumn{2}{c}{TUSL} \\
\cmidrule(lr){2-3} \cmidrule(lr){4-5}
Model & $\kappa$ & wF1 & $\kappa$ & wF1 \\
\midrule
BIOT & 29.8 & 47.8 & 17.7 & 41.3 \\
EEGPT & 32.0 & 49.3 & 29.8 & 50.9 \\
LaBraM & 38.8 & 54.2 & 36.1 & 52.2 \\
CBraMod & 38.3 & 54.6 & 42.6 & 59.0 \\
REVE & 39.3 & 55.2 & \underline{57.4} & \underline{71.5} \\
CodeBrain & 42.3 & 56.9 & 34.8 & 54.5 \\
\midrule
NeuroLM-B & 39.3 & 55.1 & 38.8 & 59.3 \\
CerebraGloss-3B & 27.3 & 44.4 & 38.6 & 55.7 \\
\midrule
REVE + Qwen & 40.6 & 55.6 & 40.8 & 58.8 \\
CodeBrain + Qwen & 34.5 & 48.1 & 29.4 & 47.2 \\
LaBraM + Qwen & 26.2 & 44.6 & 34.1 & 50.3 \\
\midrule
OctoMed-7B & 3.9 & 25.0 & $-$6.1 & 28.5 \\
MedGemma-1.5-4B & $-$1.9 & 18.7 & $-$3.2 & 23.4 \\
MedGemma-27B & 0.4 & 16.8 & $-$3.9 & 29.0 \\
\rowcolor[HTML]{FFF4D6}\base{} (base) & 2.2 & 16.9 & $-$5.3 & 23.1 \\
\midrule
\rowcolor[HTML]{FFF4D6}\textbf{\ours{} (SFT)} & \underline{52.0}\,{\tiny\textcolor[HTML]{2E8B47}{+49.8}} & \underline{64.1}\,{\tiny\textcolor[HTML]{2E8B47}{+47.2}} & 41.1\,{\tiny\textcolor[HTML]{2E8B47}{+46.4}} & 56.9\,{\tiny\textcolor[HTML]{2E8B47}{+33.8}} \\
\rowcolor[HTML]{FFF4D6}\textbf{\ours{} (RL)} & \textbf{52.5}\,{\tiny\textcolor[HTML]{2E8B47}{+50.3}} & \textbf{64.7}\,{\tiny\textcolor[HTML]{2E8B47}{+47.8}} & \textbf{59.6}\,{\tiny\textcolor[HTML]{2E8B47}{+64.9}} & \textbf{72.9}\,{\tiny\textcolor[HTML]{2E8B47}{+49.8}} \\
\bottomrule
\end{tabular}
\end{table}

\begin{table*}[t]
\centering
\footnotesize
\caption{\textbf{Threshold-free metrics on the binary tasks.} \textbf{AUROC} and \textbf{AUC-PR} (average precision), in \%, on the held-out test sets. Both are computed from each model's score for the positive option (abnormal, seizure, epilepsy, mental arithmetic). At chance level, AUC-PR equals the share of positives: TUAB 45.7, TUSZ 18.9, TUEP 75.0, EEGMAT 25.0. Siena is omitted, because the raw-EEG variants, the zero-shot VLMs and \ours{} have no option probabilities on it. LaBraM + Qwen has no option probabilities on any task. Rows and names follow main paper Table~2. The best value per column is in bold and the second best is underlined. The highlighted rows are \ours{} and its base model, \base{}. As in Table~2, the small {\tiny\textcolor[HTML]{2E8B47}{+x.x}} values give the gain of \ours{} over \base{}.}
\label{tab:c6_auc}
\begin{tabular}{lcccccccc}
\toprule
 & \multicolumn{2}{c}{TUAB} & \multicolumn{2}{c}{TUSZ} & \multicolumn{2}{c}{TUEP} & \multicolumn{2}{c}{EEGMAT} \\
\cmidrule(lr){2-3} \cmidrule(lr){4-5} \cmidrule(lr){6-7} \cmidrule(lr){8-9}
Model & AUROC & AUC-PR & AUROC & AUC-PR & AUROC & AUC-PR & AUROC & AUC-PR \\
\midrule
BIOT & 90.7 & 91.3 & 55.5 & 20.3 & \underline{91.4} & \underline{97.4} & 51.4 & 27.5 \\
EEGPT & \textbf{93.0} & \textbf{92.6} & 60.3 & 26.4 & 74.4 & 91.3 & 61.7 & 31.3 \\
LaBraM & 89.8 & 89.8 & 63.4 & 26.9 & 87.8 & 96.1 & \textbf{77.4} & \textbf{60.5} \\
CBraMod & 89.4 & 90.2 & 63.5 & 29.8 & \textbf{93.3} & \textbf{98.0} & 71.6 & 44.4 \\
REVE & 91.4 & \underline{92.2} & 67.0 & 33.6 & 85.2 & 94.6 & 49.0 & 24.7 \\
CodeBrain & \underline{91.5} & 91.5 & 66.7 & 36.9 & 89.2 & 96.9 & 65.8 & 32.7 \\
\midrule
NeuroLM-B & 89.8 & 89.0 & 54.6 & 20.9 & 72.0 & 87.4 & 72.0 & 46.3 \\
CerebraGloss-3B & 65.9 & 60.7 & 68.9 & 40.0 & 89.4 & 96.3 & 60.1 & 31.2 \\
\midrule
REVE + Qwen & 81.0 & 82.3 & 80.1 & 45.5 & 83.6 & 91.9 & 60.5 & 35.1 \\
CodeBrain + Qwen & 80.4 & 75.0 & 85.3 & 53.9 & 71.4 & 85.8 & 53.1 & 31.6 \\
\midrule
OctoMed-7B & 46.9 & 44.1 & 55.4 & 21.0 & 52.8 & 79.0 & 64.2 & 46.0 \\
MedGemma-1.5-4B & 47.6 & 45.5 & 61.3 & 22.8 & 52.9 & 78.8 & 64.0 & 34.0 \\
MedGemma-27B & 59.6 & 54.8 & 57.2 & 23.7 & 45.3 & 73.2 & 38.5 & 30.7 \\
\rowcolor[HTML]{FFF4D6}\base{} (base model) & 51.8 & 47.5 & 52.1 & 21.3 & 48.0 & 75.7 & 29.8 & 20.7 \\
\midrule
\rowcolor[HTML]{FFF4D6}\textbf{\ours{} (SFT)} & 89.6\,{\tiny\textcolor[HTML]{2E8B47}{+37.8}} & 89.0\,{\tiny\textcolor[HTML]{2E8B47}{+41.5}} & \underline{90.5}\,{\tiny\textcolor[HTML]{2E8B47}{+38.4}} & \underline{71.3}\,{\tiny\textcolor[HTML]{2E8B47}{+50.0}} & 77.7\,{\tiny\textcolor[HTML]{2E8B47}{+29.7}} & 91.4\,{\tiny\textcolor[HTML]{2E8B47}{+15.7}} & \underline{76.7}\,{\tiny\textcolor[HTML]{2E8B47}{+46.9}} & \underline{53.9}\,{\tiny\textcolor[HTML]{2E8B47}{+33.2}} \\
\rowcolor[HTML]{FFF4D6}\textbf{\ours{} (RL)} & 89.2\,{\tiny\textcolor[HTML]{2E8B47}{+37.4}} & 88.4\,{\tiny\textcolor[HTML]{2E8B47}{+40.9}} & \textbf{93.7}\,{\tiny\textcolor[HTML]{2E8B47}{+41.6}} & \textbf{79.9}\,{\tiny\textcolor[HTML]{2E8B47}{+58.6}} & 86.2\,{\tiny\textcolor[HTML]{2E8B47}{+38.2}} & 95.1\,{\tiny\textcolor[HTML]{2E8B47}{+19.4}} & 73.5\,{\tiny\textcolor[HTML]{2E8B47}{+43.7}} & 47.8\,{\tiny\textcolor[HTML]{2E8B47}{+27.1}} \\
\bottomrule
\end{tabular}
\end{table*}

\paragraph{The best model on each task does not change.} On each of the six tasks, the model with the highest balanced accuracy in Table~2 also has the highest AUROC (binary tasks) or the highest $\kappa$ (TUEV, TUSL). \ours{} is best on TUSZ (AUROC 93.7, AUC-PR 79.9). It is also best on TUEV ($\kappa$ 52.5) and, after RL, on TUSL ($\kappa$ 59.6). As in Table~2, an encoder probe is highest on the other three tasks: TUAB (EEGPT, AUROC 93.0 against 89.6 for the better \ours{} row), TUEP (CBraMod, 93.3 against 86.2) and EEGMAT (LaBraM, 77.4 against 76.7).


\subsection{Spatio-temporal Grounding Details (main paper Sec.~3.3, Table~3)}\label{app:grounding}

Main paper Table~3 tests whether a model states \emph{when} and \emph{where} a seizure or event occurs, in agreement with expert annotations that are used only for evaluation (Appendix~\ref{app:datasets}). EEG, like video, unfolds over time and across space. The metrics are therefore adapted from three video grounding tasks: grounded video question answering~\cite{Xiao2024_CanITrustYourAnswer_NExTGQA}, spatio-temporal video grounding~\cite{Zhang2020_WhereDoesItExist_VidSTG,Yang2022_TubeDETR} and action spotting~\cite{Giancola2018_SoccerNet,Hong2022_SpottingTemporallyPreciseEvents}. Table~3 has four column groups: grounded QA (TUSZ), spatio-temporal grounding (TUSZ), event spotting, and space and time (TUEV). For each group, this part states the question the model is asked, the test items, and the rule that scores an answer. Examples of \ours{} (RL) follow, including two failure cases.

\paragraph{Input and question.}
Each group has its own question, given below. For every question, \ours{} receives the frames of the whole recording, with all four renderings, and the frame description of Appendix~\ref{app:adaptation} precedes the question.

\paragraph{How an answer is scored.}
The models answer in free text. The exception is the four-option region question of the mirroring test (Appendix~\ref{app:exp_counter}), which is scored by the chosen option. An extractor LLM (\texttt{Qwen3-32B}, as in Appendix~\ref{app:adaptation}) only \emph{extracts} what the answer states: times in recording seconds, the side, and channel or region names. It does not decide whether the answer is right. Each metric is then computed from the extracted values and the expert annotation by a fixed rule, given below. All metrics are averaged over the test items and reported in percent. An item from which no time (or no region) can be extracted scores 0 on the metrics that need it, unless stated otherwise.

\paragraph{Notation.}
Time is measured in recording seconds. For the interval metrics (tIoU, vIoU), time is counted in whole seconds: second $k$ belongs to an interval $[a,b]$ if its midpoint lies inside it, $a \le k+0.5 \le b$. For one test item,
$P$ (prediction) is the set of seconds in the interval the model states, and
$G$ (ground truth) is the set of seconds of the seizure or event as annotated by the experts.
The five scalp regions are L temp, L para, mid, R para and R temp (left temporal, left parasagittal, midline, right parasagittal, right temporal).
$R_P$ is the set of regions the model names, and
$R_t$ is the set of regions annotated at second $t$. A channel name counts as the region it belongs to.
$|\cdot|$ is the number of elements of a set, so $|P|$ is the length of the stated interval in seconds.

\paragraph{Grounded question answering: Acc@QA, mIoP, Acc@GQA.}
In grounded video question answering~\cite{Xiao2024_CanITrustYourAnswer_NExTGQA}, a model must answer a question and also point to the video segment that supports its answer. Here, the supporting segment is the time the model cites in its reasoning.
\emph{Test items}: the 165 TUSZ test recordings with an annotated seizure.
\emph{Question}: the TUSZ question with its options (Table~\ref{tab:b2_questions}), as for main paper Table~2. It asks for no time and no region.
\emph{What is scored}: the answer, and the times the model cites in its reasoning. Let $C$ be the time the reasoning cites for abnormal findings, that is, the union of all cited intervals. A single cited time $t$ counts as its 10-second frame, $[t, t+10]$. Let $S$ be the time covered by all annotated seizures of the recording. A reasoning trace can cite several intervals, so it is compared with all annotated seizures, whereas~\cite{Xiao2024_CanITrustYourAnswer_NExTGQA} compares with the single best-overlapping segment. Here $|\cdot|$ is a duration in seconds.
\begin{itemize}
\item \textbf{Acc@QA}: the fraction of recordings answered ``seizure''.
\item \textbf{mIoP} (mean intersection over prediction): for one recording, IoP is the share of the cited time that lies inside a seizure,
\[\mathrm{IoP} = \frac{|C \cap S|}{|C|}.\]
It is 1 if every cited time lies inside a seizure and 0 if none does. mIoP is the mean IoP over the recordings whose reasoning cites a time. As in~\cite{Xiao2024_CanITrustYourAnswer_NExTGQA}, IoP is used rather than IoU because the cited evidence should fall inside the annotated segment but need not cover all of it.
\item \textbf{Acc@GQA}: the fraction of recordings that are answered ``seizure'' \emph{and} have $\mathrm{IoP} \ge 0.5$, the threshold of~\cite{Xiao2024_CanITrustYourAnswer_NExTGQA}. A recording whose reasoning cites no time counts as incorrect.
\end{itemize}

\paragraph{Spatio-temporal grounding: m\,tIoU, m\,vIoU, vIoU@0.5.}
These metrics come from spatio-temporal video grounding~\cite{Zhang2020_WhereDoesItExist_VidSTG,Yang2022_TubeDETR}, with seconds in place of video frames and sets of scalp regions in place of bounding boxes.
\emph{Test items}: the same 165 recordings.
\emph{Question}:
\begin{promptbox}
When does the seizure start and when does it end in this recording, and which scalp regions does it involve? Give the onset time and the end time (the last moment the seizure is still visible) in recording seconds, and name the regions among L temp, L para, mid, R para, R temp.
\end{promptbox}
\emph{What is scored}: the stated interval $P$ and the named regions $R_P$, against the expert annotation. Here $G$ is the set of seconds of the \textbf{annotated seizure}, from its annotated onset to its annotated end, and $R_t$ is the set of regions annotated at second $t$. If a recording has more than one annotated seizure, the answer is compared with only one of them: the seizure that shares the most seconds with $P$. If no seizure or several seizures share the most seconds, the earliest one is used. This follows~\cite{Xiao2024_CanITrustYourAnswer_NExTGQA}, where a prediction is scored against the annotated segment it overlaps most. An answer that gives only one time, or an interval shorter than one second, counts as the single second that contains it.
\begin{itemize}
\item \textbf{m\,tIoU} (mean temporal intersection over union): for one recording, tIoU is the length of the overlap between the stated and the annotated interval, divided by the length of their union,
\[\mathrm{tIoU} = \frac{|P \cap G|}{|P \cup G|}.\]
It is 1 only if both the onset and the end are exact, and 0 if the two intervals do not overlap. m\,tIoU is the mean over the 165 recordings.
\item \textbf{m\,vIoU} (mean spatio-temporal intersection over union): for one recording, vIoU is tIoU with each overlapping second weighted by how well the named regions match the regions annotated in that second (the IoU of the two region sets),
\[\mathrm{vIoU} = \frac{1}{|P \cup G|} \sum_{t \in P \cap G} \frac{|R_P \cap R_t|}{|R_P \cup R_t|}.\]
This is the vIoU of~\cite{Zhang2020_WhereDoesItExist_VidSTG}, with the per-frame overlap of predicted and annotated boxes replaced by the per-second overlap of named and annotated region sets. Hence $\mathrm{vIoU} \le \mathrm{tIoU}$, with equality only if the named regions are exactly the annotated ones in every overlapping second. m\,vIoU is the mean over the 165 recordings.
\item \textbf{vIoU@0.5}: the fraction of recordings with $\mathrm{vIoU} > 0.5$, as in~\cite{Zhang2020_WhereDoesItExist_VidSTG}.
\end{itemize}

\paragraph{Event spotting: Onset@10s, Event@2s.}
Both metrics follow the tolerance criterion of action spotting~\cite{Giancola2018_SoccerNet,Hong2022_SpottingTemporallyPreciseEvents}: a predicted time is correct if it lies within a fixed tolerance of the annotated time. Both score a single stated time $\hat{t}$.
\begin{itemize}
\item \textbf{Onset@10s}: the fraction of recordings in which the stated time is within 10\,s of the seizure onset, $|\hat{t} - t_{\mathrm{on}}| \le 10$\,s. Here $t_{\mathrm{on}}$ is the onset of the first annotated seizure. Ten seconds is the length of one frame. \emph{Test items}: 140 TUSZ seizure recordings, those of the class-balanced subset of 280 TUSZ test recordings (140 with and 140 without a seizure) that is also the basis of the temporal masking test (Appendix~\ref{app:exp_counter}). \emph{Question}: ``At what time does the seizure begin?''
\item \textbf{Event@2s}: the fraction of events for which the stated time is at most 2\,s outside the annotated event $[e_1, e_2]$, that is, $\max(0,\; e_1 - \hat{t},\; \hat{t} - e_2) \le 2$\,s. If the answer gives an interval, $\hat{t}$ is its midpoint. \emph{Test items}: the 61 focal TUEV events, each in a 20-second clip. \emph{Question}: ``At what time do the [event] occur?''
\end{itemize}

\paragraph{Space and time: sIoU, Acc@ST.}
The last group tests whether the model names where an event is, and whether it is right in space and time together.
\emph{Test items}: the 61 focal TUEV events. An event counts as focal if it is annotated in at most two of the four lateral regions (L temp, L para, R para, R temp).
\emph{Question}: ``Which channels show the [event]?''
\begin{itemize}
\item \textbf{sIoU} (spatial intersection over union): the number of regions that are both named and annotated, divided by the number that are named or annotated. Here $R_P$ is the set of named regions and $R_G$ the set of annotated regions of the event, both restricted to the four lateral regions,
\[\mathrm{sIoU} = \frac{|R_P \cap R_G|}{|R_P \cup R_G|},\]
averaged over the 61 events.
\item \textbf{Acc@ST}: the fraction of events for which the time is correct under Event@2s \emph{and} $\mathrm{sIoU} \ge 0.5$. As in Acc@GQA~\cite{Xiao2024_CanITrustYourAnswer_NExTGQA}, an item counts only if both parts are right.
\end{itemize}

\paragraph{Per-window model: CerebraGloss-3B.}
CerebraGloss-3B reads one 10-second window per query and states no time in recording seconds. For all TUSZ metrics of Table~3, it is asked the spatio-temporal grounding question for every window, preceded by ``Is there a seizure in this segment?''. Its window answers are combined by fixed rules without any tuned parameter. For grounded question answering, it answers ``seizure'' if it answers ``Yes'' for at least one window, and its cited times are the times it states in its ``Yes'' windows, shifted by the start of the window. For spatio-temporal grounding, its interval $P$ is the union of its ``Yes'' windows, and its regions $R_P$ are the union of the regions it names in those windows. The same formulas then apply. For Onset@10s, its stated time is the start of its first ``Yes'' window. For the TUEV metrics, each of the two windows of a clip is asked the same questions as the other models, and the score of the clip is the mean over its two windows.

\paragraph{Examples.}
Figures~\ref{fig:c3_1}--\ref{fig:c3_3} show \ours{} (RL) on one TUSZ recording for each question type of Table~3. The examples are selected, and the figures follow the conventions of Appendix~\ref{app:tasks}.
In Fig.~\ref{fig:c3_1} (grounded question answering), the question asks only whether there is a seizure, and the reasoning cites an interval of which 97\% lies inside the annotated seizure.
In Fig.~\ref{fig:c3_2} (spatio-temporal grounding), the model states the onset, the end and the regions of a 48-second seizure in a 27-minute recording, with a tIoU of 0.80.
In Fig.~\ref{fig:c3_3} (onset spotting), three differently worded questions about the onset give the same time and the same side.
A TUEV example is shown with spatial mirroring (Fig.~\ref{fig:c3_5}).

\paragraph{Failure cases.}
Two cases remain difficult for \ours{} (RL): recordings with several seizures, and seizures confined to one hemisphere. In Fig.~\ref{fig:c3_6}, the model is asked to list every seizure of a recording that contains five (listing prompt of Appendix~\ref{app:rq3}), and it lists a single interval that precedes the first one. In Fig.~\ref{fig:c3_7}, the stated onset is within 10\,s of the annotation, but the model names all five regions for a seizure that is annotated on the right side and at the midline.


\subsection{Counterfactual Interventions Details (main paper Sec.~3.4, Fig.~2)}\label{app:exp_counter}

The grounding metrics compare what a model states with the annotation. The two interventions of main paper Fig.~2 test whether its answer also \emph{depends} on the evidence. Both follow intervention-based evaluations of visual grounding~\cite{Li2022_EquivariantInvariantGroundingVideoQA,Petsiuk2018_RISE,Xiao2024_CanITrustYourAnswer_NExTGQA}: the evidence in the input is changed, and a grounded model should change its answer accordingly. This part gives the protocol of each intervention, and how it is applied to the models that do not read our frames. How a masked or a mirrored frame is rendered is described in Appendix~\ref{app:eeg_video}.

\paragraph{Temporal masking (main paper Fig.~2(a,\,c)).}
Temporal masking hides the frames that show the seizure and checks whether the answer changes. \emph{Test items}: 265 TUSZ test recordings, 125 with a seizure and 140 without. \emph{Question}: the TUSZ seizure question (Table~\ref{tab:b2_questions}). A masked frame keeps its labels and time axis, but its four renderings are greyed out. Balanced accuracy is measured under three conditions: on the intact input; with the \emph{evidence frames} masked, which are all frames that overlap an annotated seizure; and with the same number of \emph{control frames} masked, which are seizure-free frames of the same recording drawn at random. Seizure recordings with too few seizure-free frames for the control are left out (15 recordings). A recording without a seizure has no evidence frames. It receives one random mask of comparable size, and its answer is used for both masked conditions. The two masked conditions therefore differ only on the seizure recordings. Fig.~2(c) reports three values:
\begin{itemize}
\item \textbf{Evidence drop}: balanced accuracy on the intact input minus balanced accuracy with the evidence frames masked.
\item \textbf{Control drop}: balanced accuracy on the intact input minus balanced accuracy with the control frames masked.
\item \textbf{$\Delta$}: the evidence drop minus the control drop. A grounded model has a large evidence drop, a small control drop, and therefore a large $\Delta$.
\end{itemize}

\paragraph{Spatial mirroring (main paper Fig.~2(b,\,d)).}
Spatial mirroring swaps left and right in the signal and checks whether the answer swaps too. The signals of the left and right electrodes are exchanged before rendering. The event thus appears on the opposite side, while the frame layout is unchanged. Each question is asked on the original and on the mirrored input. An item is correct only if the answer is right on both. On the mirrored input, the correct answer is the mirrored one. Fig.~2(d) reports this paired accuracy for two questions:
\begin{itemize}
\item \textbf{Hemisphere}: the fraction of items for which the stated side equals the annotated side on both inputs. \emph{Test items}: 74 items (37 TUEV events and 37 TUSZ seizure onsets) annotated on one side only, left or right. \emph{Question}: ``On which side of the head is the [event]?'' (TUEV) or ``In which hemisphere does the seizure begin?'' (TUSZ).
\item \textbf{Region}: the fraction of items for which the chosen option is the annotated region on both inputs. \emph{Test items}: 134 items (68 TUEV events and 66 TUSZ seizure onsets). \emph{Question}: ``Which of these channel groups shows the [event]?'' (for TUSZ, ``\ldots shows the seizure at its onset?''), with the four lateral regions as options.
\end{itemize}

\paragraph{Raw-EEG variants under the interventions.}
The raw-EEG variants receive no frames, so both interventions are applied to the signal. For temporal masking, the signal of the same evidence and control windows of the same recordings is set to zero before it enters the encoder. For spatial mirroring, the left and right electrode signals are exchanged in the signal given to the encoder. The questions are the same.

\paragraph{CerebraGloss-3B under the interventions.}
CerebraGloss-3B reads one 10-second window per query, drawn as its own plot, so both interventions are applied to its windows. \emph{Temporal masking}: the evidence and control windows are those of \ours{}. A masked window counts as answered ``No''. As for Table~3, a recording is answered ``seizure'' if at least one window answers ``Yes'', with no tuned parameter. \emph{Spatial mirroring}: the left and right electrode signals are exchanged before its own plotting. For each item, CerebraGloss-3B receives the 10-second window that contains the annotated event: the window centred on the event for a TUEV item, and the window that starts 2\,s before the annotated seizure onset for a TUSZ item. The hemisphere question (left, right or both) and the four-option region question are asked in its multiple-choice format on the original and on the mirrored plot, and the chosen option is scored.

\paragraph{Examples.}
Figures~\ref{fig:c3_4} and~\ref{fig:c3_5} each show one intervention on one input, with the complete outputs of \ours{} (RL) for every condition. In Fig.~\ref{fig:c3_4} (temporal masking, TUSZ), the answer changes from ``seizure'' to ``no seizure'' when the three evidence frames are masked, and it is unchanged when three control frames are masked. In Fig.~\ref{fig:c3_5} (spatial mirroring, TUEV), the stated side changes from right to left with the signal, so the item counts as correct. A mirroring item that counts as incorrect is shown in Fig.~\ref{fig:d4b_d42_4}.

\section{Additional Findings Details}
\label{app:findings}
For each research question of main paper Sec.~4, this appendix states the motivation of the question, describes the setup, reports additional results and shows selected qualitative examples. The example figures follow the conventions of Appendix~\ref{app:tasks}.

\subsection{RQ1: What does reasoning self-training add? (main paper Sec.~4, Table~4)}\label{app:rq1}

\ours{} is trained without externally annotated spatio-temporal events. Its only supervision on \emph{where} and \emph{when} an event occurs comes from the reasoning traces of the mid-trained model.
RQ1 isolates the contribution of this reasoning supervision.
We compare \ours{} (SFT), listed as ``With reasoning'' in Table~4 of the main paper, with an Answer-only variant.
Both start from the same mid-trained checkpoint and are trained on the same recordings and labels for the same number of steps. The Answer-only variant is trained on the answer alone, so any difference between the two models is due to the reasoning supervision.

\paragraph{Reasoning supervision improves accuracy.}
With identical labels, training on reasoning raises the mean balanced accuracy over the seven evaluations (Mean BA) from 64.4 to 69.8.
The gains are largest on the three clip-level tasks: $+10.9$ points on TUEV (55.2 against 44.3), $+11.8$ on TUSL (62.8 against 51.0) and $+25.9$ on EEGMAT (75.9 against 50.0).
On TUEV, the classes that benefit most are those defined by where or when a pattern occurs: the recall of periodic lateralized discharges rises from 20.9\% to 50.1\%, of eye movements from 43.2\% to 68.5\%, and of spikes and sharp waves from 13.0\% to 23.9\%.
On EEGMAT, the Answer-only variant predicts the same class for all 36 clips, which corresponds to chance level.
The Answer-only variant is higher only on the three recording-level TUH tasks (TUAB, TUSZ and TUEP).
Its advantage on TUSZ does not carry over to the external Siena dataset, which poses the same seizure-detection task: \ours{} reaches 81.9 against 74.5.
Both models are correct on 95.7\% of the seizure-free excerpts, but \ours{} detects 68.1\% of the excerpts that contain a seizure, and the Answer-only variant 53.2\%.

\paragraph{Reasoning supervision improves grounding.}
The effect is larger on grounding than on accuracy (main paper Table~4).
\ours{} localizes more than twice as many seizure onsets within 10\,s (Onset@10s 42.1 against 17.9).
Under spatial mirroring (Appendix~\ref{app:exp_counter}), it is more often correct on both the original and the mirrored input (hemisphere 67.6 against 52.7, region 36.6 against 26.9). Its spatial answers thus follow the signal more closely.
These results support Finding~1 of the main paper: reasoning self-training improves both the spatio-temporal grounding and the final-answer accuracy of \ours{}, without reasoning or localization annotations.

\paragraph{Qualitative examples.}
Figures~\ref{fig:d1_d1_1}--\ref{fig:d1_d1_3} show three test inputs that both \ours{} models answer correctly and the Answer-only variant answers incorrectly. They are selected from the 384 such test inputs.
On a TUEV clip (Fig.~\ref{fig:d1_d1_1}), \ours{} locates the discharges on the left side and classifies them as lateralized, whereas the Answer-only variant describes the pattern as symmetric.
On a TUSZ recording (Fig.~\ref{fig:d1_d1_2}), \ours{} cites a frame inside the seizure and a region in which the seizure is annotated, whereas the Answer-only variant answers ``no seizure''.
On a Siena excerpt (Fig.~\ref{fig:d1_d1_3}), \ours{} cites the frames in which the seizure starts, whereas the Answer-only variant examines frames inside the seizure and attributes the activity to noise.
Although it is trained to output only the answer, the Answer-only variant often generates additional text at test time, as in Figs.~\ref{fig:d1_d1_1} and~\ref{fig:d1_d1_3}; no such text appears in its training targets.

\paragraph{Failure case.}
Figure~\ref{fig:d1_d1_4} shows the typical error of \ours{} (SFT) on TUSZ: a false alarm on a recording without a seizure.
RL reduces this error. On the 710 seizure-free test recordings, \ours{} (RL) is correct on 79.0\%, against 69.4\% for \ours{} (SFT), while detecting more of the 165 recordings with a seizure (93.3\% against 88.5\%).
In the example, \ours{} (RL) answers correctly.

\subsection{RQ2: How does each visual rendering contribute to performance? (main paper Sec.~4, Table~5)}\label{app:rq2}

Each frame of \ours{} combines four renderings of the same 10-second window: Waveforms, Spectrograms, Topographic maps and Sharpness (Appendix~\ref{app:eeg_video}).
RQ2 asks whether all four are needed, or whether a subset already carries the cues required for accurate and grounded answers.
We answer this by ablation. Each variant is trained with the full recipe of \ours{} (SFT), on the same recordings, with frames that contain only a subset of the renderings, and it is evaluated on the same test sets.
The variants are evaluated at step 1,200, as \ours{} (SFT), except three variants that have fewer selected reasoning traces and are evaluated at the end of their shorter post-training epoch (step 1,190 without Waveforms, 878 for Spectrograms only and 880 for Sharpness only).

\paragraph{Removing any rendering lowers accuracy and joint grounding (main paper Table~5).}
Four variants each drop one rendering.
The complete four-rendering frame achieves the highest Mean BA (69.8, against 56.9 to 67.7) and the highest joint spatio-temporal accuracy on focal TUEV events, i.e., events confined to part of the scalp (Acc@ST 41.0, against at most 24.6).
Removing the Waveforms lowers every metric of Table~5 and is the most harmful (Mean BA 56.9, Acc@ST 6.6).

\paragraph{No single rendering matches the four-rendering frame (Table~\ref{tab:d2_single}).}
We further train four variants whose frames keep only one rendering.
None of them matches the four-rendering frame in Mean BA (69.8): Waveforms only comes closest (64.7), followed by Topographic maps only (61.6), Spectrograms only (56.1) and Sharpness only (43.0).
On the focal TUEV events, every single-rendering variant is below the four-rendering frame in Event@2s, sIoU and Acc@ST (Acc@ST 41.0, against at most 19.7).
As noted in the main paper, seizure grounding relies primarily on the Waveforms: Waveforms only reaches the highest Acc@GQA and m\,tIoU on TUSZ seizures (75.8 and 40.9, against 66.1 and 30.7 for the four-rendering frame). It is, however, 5.1 points lower in Mean BA and 27.9 points lower in Acc@ST.
Together with main paper Table~5, these results indicate that the renderings provide complementary cues, and that their combination is needed to perform well across tasks (Finding~2 of the main paper).

\begin{table}[t]\centering\footnotesize
\caption{\textbf{Frames with a single rendering.} As in Table~5 of the main paper, each variant keeps the rendering marked \textcolor[HTML]{2E8B47}{\checkmark} in the frame and drops those marked \textcolor[HTML]{C0392B}{\scriptsize$\times$}, and is retrained on the same recordings with the same recipe. Columns and test sets are those of Table~5, and the first row repeats its first row. Wave.: Waveforms, Spec.: Spectrograms, Topo.: Topographic maps, Sharp.: Sharpness. All values are percentages, and the best result in each column is in bold. No single rendering reaches the \textbf{Mean BA} of the four-rendering frame.}\label{tab:d2_single}
\setlength{\tabcolsep}{1.1pt}
\begin{tabular}{@{}cccc@{\hspace{4pt}}ccccc>{\columncolor[HTML]{EAF1FB}}c@{}}\toprule
\multicolumn{4}{c}{EEG renderings} & \multicolumn{2}{c}{TUSZ seizures} & \multicolumn{3}{c}{TUEV focal events} & \\\cmidrule(r){1-4}\cmidrule(lr){5-6}\cmidrule(lr){7-9}
{\scriptsize Wave.} & {\scriptsize Spec.} & {\scriptsize Topo.} & {\scriptsize Sharp.} & {\scriptsize Acc@GQA} & {\scriptsize m\,tIoU} & {\scriptsize Event@2s} & {\scriptsize sIoU} & {\scriptsize Acc@ST} & {\scriptsize Mean BA}\\\midrule
\textcolor[HTML]{2E8B47}{\checkmark} & \textcolor[HTML]{2E8B47}{\checkmark} & \textcolor[HTML]{2E8B47}{\checkmark} & \textcolor[HTML]{2E8B47}{\checkmark} & 66.1 & 30.7 & \textbf{55.7} & \textbf{48.0} & \textbf{41.0} & \textbf{69.8}\\\midrule
\textcolor[HTML]{2E8B47}{\checkmark} & \textcolor[HTML]{C0392B}{\scriptsize$\times$} & \textcolor[HTML]{C0392B}{\scriptsize$\times$} & \textcolor[HTML]{C0392B}{\scriptsize$\times$} & \textbf{75.8} & \textbf{40.9} & 27.9 & 38.1 & 13.1 & 64.7\\
\textcolor[HTML]{C0392B}{\scriptsize$\times$} & \textcolor[HTML]{2E8B47}{\checkmark} & \textcolor[HTML]{C0392B}{\scriptsize$\times$} & \textcolor[HTML]{C0392B}{\scriptsize$\times$} & 33.9 & 14.8 & 42.6 & 28.6 & 11.5 & 56.1\\
\textcolor[HTML]{C0392B}{\scriptsize$\times$} & \textcolor[HTML]{C0392B}{\scriptsize$\times$} & \textcolor[HTML]{2E8B47}{\checkmark} & \textcolor[HTML]{C0392B}{\scriptsize$\times$} & 40.0 & 18.7 & 45.9 & 31.7 & 19.7 & 61.6\\
\textcolor[HTML]{C0392B}{\scriptsize$\times$} & \textcolor[HTML]{C0392B}{\scriptsize$\times$} & \textcolor[HTML]{C0392B}{\scriptsize$\times$} & \textcolor[HTML]{2E8B47}{\checkmark} & 18.8 & 14.9 & 34.4 & 29.1 & 13.1 & 43.0\\\bottomrule
\end{tabular}\end{table}

\paragraph{Qualitative examples.}
Figures~\ref{fig:d2_1}--\ref{fig:d2_4} show one figure for each variant of main paper Table~5, i.e., for each removed rendering. Each figure contains two test inputs:
(i) a \emph{correct example}, a TUSZ recording that the variant answers correctly, where every time and region it states is confirmed by the expert annotation, and
(ii) a \emph{wrong example}, a focal TUEV event for which the time or the scalp region it states is contradicted by the annotation. Table~5 shows its largest drops on these focal TUEV events.
The wrong examples illustrate the roles described in the main paper. Without Waveforms, both the time and the side are wrong (Fig.~\ref{fig:d2_1}). Without Spectrograms or Topographic maps, the error is in the scalp region (Figs.~\ref{fig:d2_2} and~\ref{fig:d2_3}). Without Sharpness, the error is in the time (Fig.~\ref{fig:d2_4}).
The figures show the frames that each variant receives; for the variant without Waveforms, the Waveforms of the same windows are added below its frames for the reader only.
The examples are selected, and Table~5 reports the aggregate results.

\subsection{RQ3: Does training on short EEG recordings generalize to longer ones? (main paper Sec.~4, Fig.~3)}\label{app:rq3}

\ours{} is trained on recordings of at most 30 minutes, whereas clinical EEG recordings routinely last several hours.
RQ3 asks whether the spatio-temporal localization learned on short recordings extends to longer ones.
This part gives the protocols and additional results behind main paper Fig.~3 and Finding~3.
We evaluate on Siena, which is not used in training, in two settings.
The \emph{needle-in-a-haystack} setting (Fig.~3(a,\,b)) places a single seizure in a video of 10 to 60 minutes that the model reads in one pass; it isolates the effect of video length.
The \emph{whole-recording} setting (Fig.~3(c,\,d)) processes recordings of several hours with overlapping 20-minute windows; it corresponds to clinical use.

\paragraph{Needle-in-a-haystack protocol.}
Each video is a continuous 10-, 20-, 40- or 60-minute excerpt of one recording that contains exactly one annotated seizure.
The excerpt is chosen such that the seizure lies at a given position. Each seizure is placed at five positions: 10\%, 30\%, 50\%, 70\% and 90\% of the video length.
Videos of 40 and 60 minutes are longer than any training input.
Each video is evaluated with two prompts. The \emph{listing} prompt (main paper Fig.~3(a), ``Detection'') requires the model to decide whether a seizure is present before localizing it:
\begin{promptbox}
List every seizure in this recording with its start and end time in seconds, or say that there is none.
\end{promptbox}
The \emph{told-one} prompt (main paper Fig.~3(b), ``Localization'') states that there is exactly one seizure and measures localization alone:
\begin{promptbox}
This recording contains exactly one electrographic seizure. Give its start and end time in seconds.
\end{promptbox}
We report Onset@10s, the percentage of seizures whose predicted onset is within 10 seconds of the annotated onset.

\begin{figure*}[t]
\centering
\includegraphics[width=\textwidth]{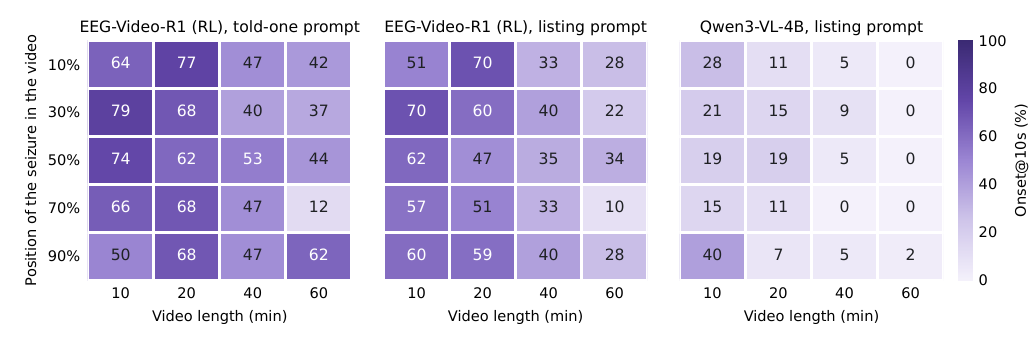}
\caption{\textbf{Needle-in-a-haystack: \ours{} is above its base model at every video length and seizure position, and video length matters more than position.} Onset@10s (\%) on Siena. Columns give the \textbf{video length}; rows give the \textbf{position of the seizure}, measured from the start of the video. Results are shown for \ours{} (RL) with the told-one and the listing prompt, and for \base{} with the listing prompt. Videos of 40 and 60 minutes are longer than any training input.}
\label{fig:d3_needle_grid}
\end{figure*}

\paragraph{Localization extends beyond the training length.}
Table~\ref{tab:d3_needle} reports the results of \ours{} (RL).
With the told-one prompt, Onset@10s is 68.8\% and 68.5\% for 10- and 20-minute videos, and 46.5\% and 39.4\% for 40- and 60-minute videos.
With the listing prompt, which also requires detecting the seizure, the corresponding values are 60.1\%, 57.3\%, 35.8\% and 24.1\%.
Accuracy decreases with video length, but \ours{} still localizes seizures at twice the maximum training duration.
In contrast, \base{} approaches chance level at these lengths (at most 2.5\% at 60 minutes, Fig.~\ref{fig:d3_needle_grid}).

\begin{table}[b]\centering\footnotesize
\caption{\textbf{Needle-in-a-haystack, \ours{} (RL).} Onset@10s (\%) for the listing and the told-one prompt. Videos longer than 30 minutes are longer than any training input.}\label{tab:d3_needle}
\begin{tabular}{@{}lccc@{}}\toprule
 & & \multicolumn{2}{c}{Onset@10s}\\\cmidrule(l){3-4}
Length & Seizures & listing & told-one\\\midrule
10 min & 208 & 60.1 & 68.8\\
20 min & 232 & 57.3 & 68.5\\
40 min & 215 & 35.8 & 46.5\\
60 min & 203 & 24.1 & 39.4\\
\bottomrule\end{tabular}\end{table}

\paragraph{The position of the seizure has no consistent effect.}
Figure~\ref{fig:d3_needle_grid} reports Onset@10s for every combination of video length and position (40 to 47 seizures per cell; 20 for the 10-minute videos at 90\%).
For \ours{} (RL) with the told-one prompt, Onset@10s is between 50\% and 79\% at every position for 10- and 20-minute videos, and between 39.5\% and 53.5\% for 40-minute videos.
At 60 minutes, the values vary more across positions (12\% to 62\%), without a trend from early to late positions.
Video length therefore affects accuracy more than position does.
With the same listing prompt, \ours{} is above \base{} in every cell; \base{} stays at or below 28\% in all cells but one.

\paragraph{Whole-recording protocol.}
This setting uses the 41 complete Siena recordings (141 hours, 47 seizures, 14 patients).
Each recording is divided into 20-minute windows that start every 15 minutes. Each window is given the listing prompt and, separately, the yes/no seizure question. The events listed in windows whose yes/no answer is ``seizure'' are converted to recording time and merged into one list, which is scored.

\paragraph{EEG encoders on whole recordings (main paper Fig.~3(c,\,d)).}
The EEG encoders are evaluated on the same 41 recordings. For each encoder, its TUSZ probe (a classifier trained on the encoder's features for TUSZ; Appendix~\ref{app:encoder_setup}) gives a seizure probability for every 10-second window. Windows above a threshold are positive, and consecutive positive windows form one event. The threshold is chosen separately for each encoder by event F1 on TUSZ validation recordings, before Siena is scored; Siena is used only for testing. All models, including \ours{}, are scored with the standard SzCORE evaluation package~\cite{Dan2024_SzCORE} and its default event rules, under which a predicted event is correct if it overlaps an annotated seizure, within the default tolerances of the package. Event F1 and false alarms per day are computed per patient and averaged over the 14 patients.

\paragraph{Whole recordings: higher event F1 with fewer false alarms.}
On the 41 recordings, \ours{} achieves more than twice the event F1 of the EEG encoders (56.5\% against at most 24.7\%), with nearly 18 times fewer false alarms (main paper Fig.~3(c,\,d)).
The localization learned on recordings of at most 30 minutes thus carries over to recordings of several hours (Finding~3 of the main paper).

\paragraph{Qualitative examples.}
Figures~\ref{fig:d3_d3_1}--\ref{fig:d3_d3_3} show needle-in-a-haystack examples at 10, 40 and 60 minutes. In each of them, both prompts return the seizure with an onset error of at most 5 seconds.
Figure~\ref{fig:d3_d3_4} shows a whole recording of 2.5 hours: the seizure is found in the window that contains it, and a seizure-free window of the same recording is left unflagged.
The examples are selected. In general, \ours{} states the onset of a seizure more accurately than its end.

\paragraph{Failure case.} Figure~\ref{fig:d3_d3_5} shows a 13-hour recording. The seizure is found with an onset error of 7 seconds, but ten further events are listed where no seizure is annotated. Five of them remain after the yes/no filter. In the window shown, an artifact is read as a seizure.

\subsection{RQ4: Does reasoning self-training extend to raw-EEG LLMs? (main paper Sec.~4, Table~6)}
\label{app:rq4}

Reasoning self-training post-trains a model on reasoning that the mid-trained model has generated itself (Appendix~\ref{app:adaptation}). It can therefore work only if this reasoning describes the EEG.
RQ4 asks whether the recipe also works for models that read the raw EEG signal instead of a rendered video.
To answer it, we replace the vision encoder of \base{} with a pre-trained EEG encoder. This gives three raw-EEG variants: REVE + Qwen, CodeBrain + Qwen and LaBraM + Qwen (training settings in Appendix~\ref{app:hparams}).
This part supports main paper Table~6, the raw-EEG rows of Tables~2 and~3, and Finding~4. It proceeds in two steps.
First, we examine the reasoning that the variants generate after mid-training, since this is the supervision that self-training would rely on (Appendix~\ref{app:rq4_mid}). We find that it rarely describes the EEG; Fig.~\ref{fig:d4a_d41_1} shows an example.
Second, we therefore train the variants on the reasoning traces of \ours{} and examine what they learn from them (Appendix~\ref{app:rq4_post}).

\subsubsection{The problem after mid-training (main paper Table~6)}\label{app:rq4_mid}

The variants are mid-trained on the same examples and answer labels as \ours{}. All four models are then given the same questions and the same request to reason before answering.
After mid-training, the variants answer above chance: their mean balanced accuracy is 50.7 to 51.3, against a chance level of 42.9 (main paper Table~2) and 60.4 for \ours{}.
The problem lies in their reasoning.

\paragraph{The variants do not describe the EEG.}
Figure~\ref{fig:d4a_d41_1} shows the reasoning of all four models after mid-training, on a TUSZ recording with one seizure. It is the typical outcome.
\ours{} cites the frame in which the seizure is visible and names the scalp regions involved, all of which are annotated.
The three variants give the same answer, but their reasoning never refers to the signal: REVE + Qwen and LaBraM + Qwen search their input for the word ``seiz'', as if it were written text, and CodeBrain + Qwen describes repeated letters.
Table~\ref{tab:d4a_text} shows that this behaviour is systematic.
Between 82\% and 88\% of the traces of the variants treat the input as written text or symbols, against 0.8\% for \ours{}.
At most 3.7\% of their traces state an observation tied to a time or a scalp region, against 74.3\% for \ours{}.
Between 12.8\% and 42.1\% of their traces are corrupted, against 5.2\% for \ours{}.

\paragraph{Consequence for self-training.}
Self-training keeps the reasoning traces of a model that lead to the correct answer and trains the model on them.
For the raw-EEG variants, almost none of these traces contains an observation about the EEG. Training on them would only reinforce reasoning that is unrelated to the signal.
Self-training is therefore ineffective for the raw-EEG variants.

\begin{table}[t]
\centering\small
\begin{tabular}{@{}l@{\hskip 5pt}c@{\hskip 5pt}c@{\hskip 5pt}c@{\hskip 8pt}c@{\hskip 5pt}c@{}}
\toprule
 & \multicolumn{3}{c}{Main paper Table~6} & \multicolumn{2}{c}{Additional}\\
\cmidrule(r){2-4}\cmidrule(l){5-6}
Model & BA$\uparrow$ & Grnd.$\uparrow$ & Corr.$\downarrow$ & Text$\downarrow$ & G.$\neg$T.$\uparrow$\\
\midrule
REVE + Qwen & 51.0 & 3.7 & 12.8 & 82.8 & 0.7\\
CodeBrain + Qwen & 50.7 & 2.0 & 13.4 & 82.4 & 0.3\\
LaBraM + Qwen & 51.3 & 3.0 & 42.1 & 88.3 & 0.3\\
\midrule
\ours{} & 60.4 & 74.3 & 5.2 & 0.8 & 74.3\\
\bottomrule
\end{tabular}
\caption{\textbf{After mid-training, the raw-EEG variants reason about their input as if it were text.} \textbf{BA}: mean balanced accuracy over the seven evaluations. The other columns are percentages of reasoning traces, judged by an LLM (\texttt{Qwen3-32B}) on 286 test recordings from the six tasks. \textbf{Grnd.}: the trace states an observation tied to a time or a named scalp location. \textbf{Corr.}: the trace is empty, looping, broken, or cut off at the token limit. \textbf{Text} (not in the main paper): the trace treats the input as written text or symbols. \textbf{G.$\neg$T.}: Grnd.\ and not Text.}
\label{tab:d4a_text}
\end{table}

\paragraph{Failure case.}
The reasoning of \ours{} after mid-training also contains errors. In Fig.~\ref{fig:d4a_d41_2}, a correct answer comes with a seizure episode that is not annotated.
Post-training uses only selected and refined traces (Appendix~\ref{app:adaptation}).

\subsubsection{Training the variants on the reasoning of \ours{} (main paper Tables~2 and~3)}\label{app:rq4_post}

Since the variants' own traces do not provide usable supervision, we post-train each variant on reasoning traces generated by \ours{} (Appendix~\ref{app:hparams}).
This tests whether the reasoning learned from video can be transferred to models that read the raw signal.
The example figures of this part show each variant next to \ours{} (RL) on the same recording and question. The frames in them are for the reader only; the variants receive the signal.

\paragraph{Distillation improves accuracy and grounding.}
Mean balanced accuracy rises for all three variants, by up to 11.0 points: from 51.0 to 62.0 for REVE + Qwen, from 50.7 to 58.0 for CodeBrain + Qwen and from 51.3 to 57.1 for LaBraM + Qwen (main paper Table~2).
Their reasoning also takes the form of the reasoning of \ours{}: it names a time in recording seconds and the scalp regions involved, and then gives the answer.
This cited evidence is correct well above chance. The grounded question answering accuracy of the variants (Acc@GQA) is 23.6\% to 36.4\%, against a chance level of 10.5\% (main paper Table~3).
Figures~\ref{fig:d4b_d42_g1} and~\ref{fig:d4b_d42_g2} show two recordings on which all three variants, like \ours{} (RL), answer ``seizure'', cite a time at the annotated seizure and name annotated regions.

\paragraph{The variants remain below \ours{}.}
Accuracy and grounding stay below those of \ours{}, which reaches 69.8\% mean balanced accuracy and 66.1\% Acc@GQA after SFT with rendered-video input.
Figure~\ref{fig:d4b_d42_2} shows an example: the variants answer correctly, but the times they cite are far from the seizure.

\paragraph{Summary.}
Self-training is ineffective for the raw-EEG variants, since their own reasoning after mid-training does not describe the EEG. In contrast, the reasoning of \ours{} can be distilled into them, which improves both their accuracy and their grounding (Finding~4 of the main paper).

\paragraph{Failure case.}
\ours{} also fails on some grounding items. Figure~\ref{fig:d4b_d42_4} shows a mirroring item (Appendix~\ref{app:exp_counter}) on which it names the wrong side.


\section{Expert Review (main paper Sec.~5)}
\label{app:expert}
The grounding metrics of main paper Sec.~3.3 compare the times and scalp regions that \ours{} states with the dataset annotations. They do not tell whether the reasoning as a whole is sound to a clinician. To assess this directly, an EEG expert reviewed the reasoning of \ours{} (main paper Sec.~5).

\begin{figure*}[t]
\centering
\includegraphics[width=0.92\textwidth]{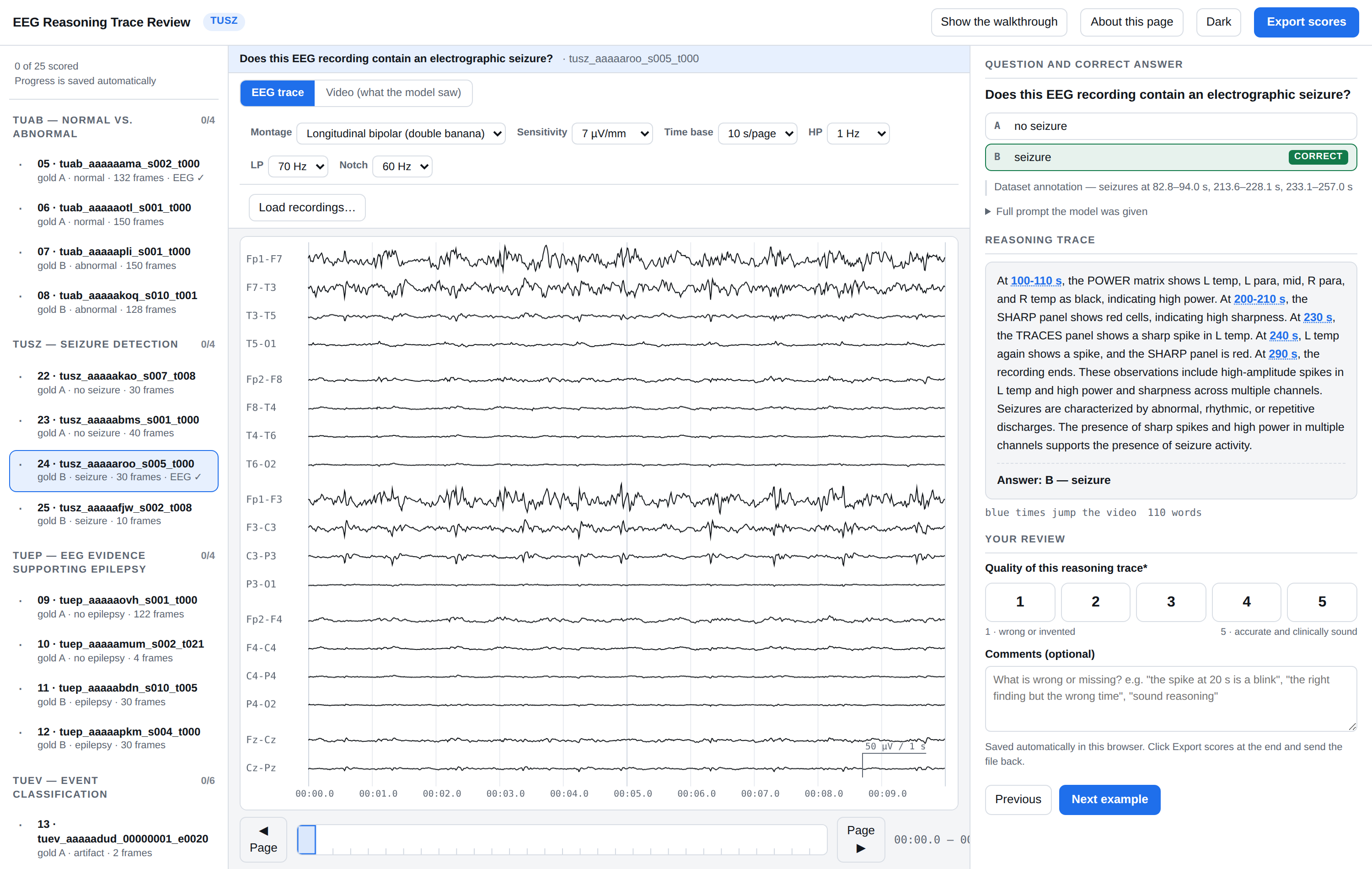}
\caption{\textbf{Interface of the expert review.} Left: the 25 cases, grouped by task. Centre: the EEG of the selected case in a clinical viewer; a second tab shows the video given to the model. Right: the question with the correct answer, the dataset annotation, the \textbf{reasoning trace} with clickable times, and the \textbf{5-point rating} with an optional comment.}
\label{fig:e_tool}
\end{figure*}

\begin{figure*}[tp]
\centering\includegraphics[width=\textwidth]{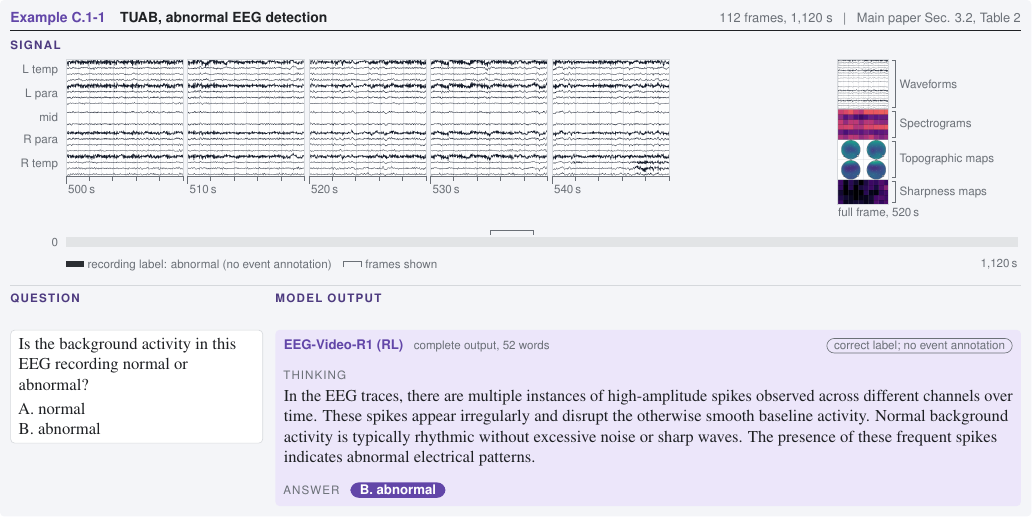}\caption{\emph{Example for Appendix~\ref{app:tasks}.} \textbf{TUAB: the model classifies a 19-minute recording as abnormal from the whole video in one pass.} The signal area shows five of the 112 frames and where they lie in the recording.}\label{fig:c1_c1_1}

\bigskip
\centering\includegraphics[width=\textwidth]{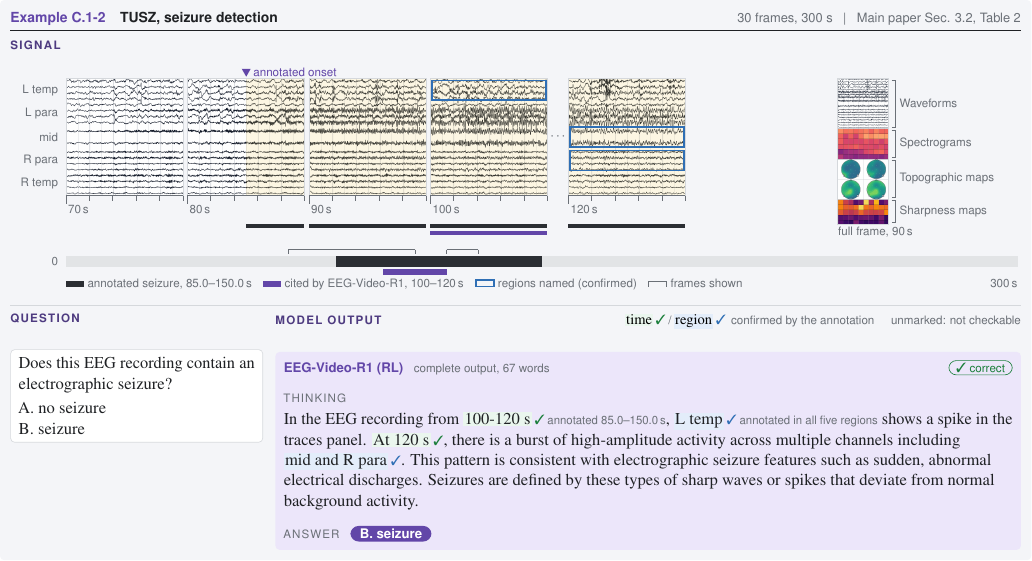}\caption{\emph{Example for Appendix~\ref{app:tasks}.} \textbf{TUSZ: the model answers ``seizure'' and cites the seizure itself.} In a 5-minute recording, the reasoning points to \textbf{100--120\,s} and \textbf{120\,s}. Both lie inside the annotated seizure (85.0--150.0\,s). The reasoning also names the \textbf{left temporal}, \textbf{midline} and \textbf{right parasagittal} regions, which are annotated at those times (blue boxes).}\label{fig:c1_c1_2}
\end{figure*}

\begin{figure*}[tp]
\centering\includegraphics[width=\textwidth]{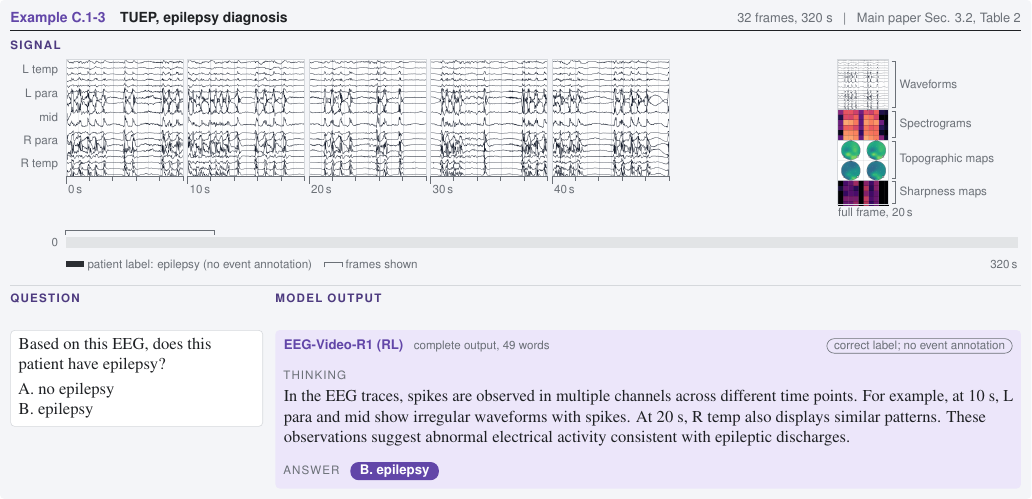}\caption{\emph{Example for Appendix~\ref{app:tasks}.} \textbf{TUEP: a diagnosis about the patient (epilepsy or not) from one 5-minute recording.} The reasoning points to times and regions with large signal deflections, which are visible in the frames shown.}\label{fig:c1_c1_3}
\end{figure*}

\begin{figure*}[tp]
\centering\includegraphics[width=\textwidth]{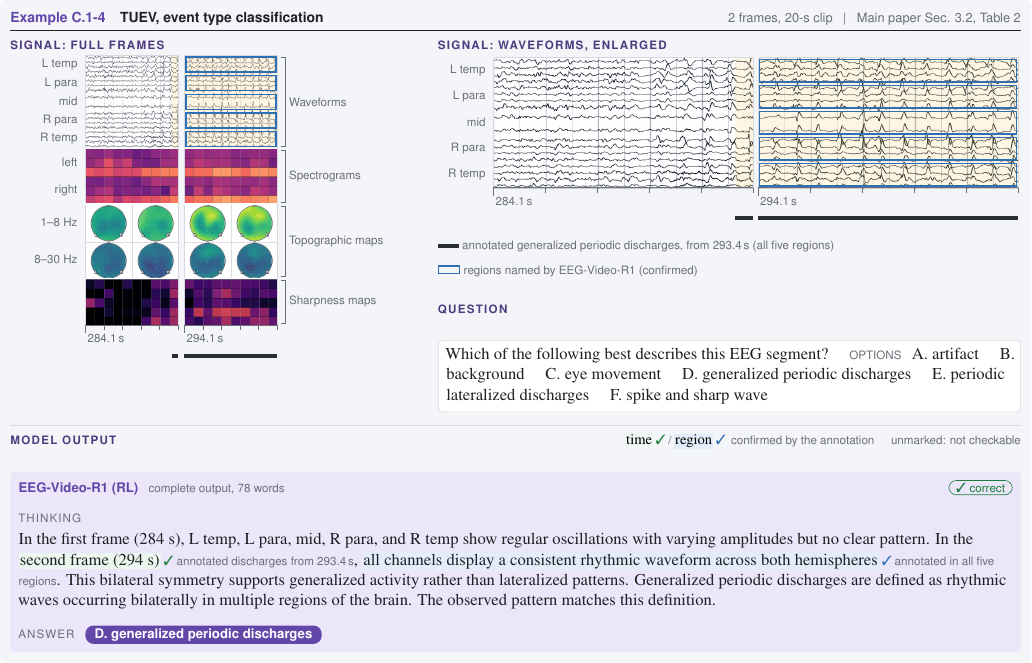}\caption{\emph{Example for Appendix~\ref{app:tasks}.} \textbf{TUEV: the model recognizes generalized periodic discharges in a 20-second clip.} It places the rhythmic pattern in the \textbf{second frame} and on \textbf{all regions}. The discharges are annotated from 293.4\,s on all five regions. Left: both full frames. Right: their Waveforms panels, enlarged.}\label{fig:c1_c1_4}
\end{figure*}

\begin{figure*}[tp]
\centering\includegraphics[width=\textwidth]{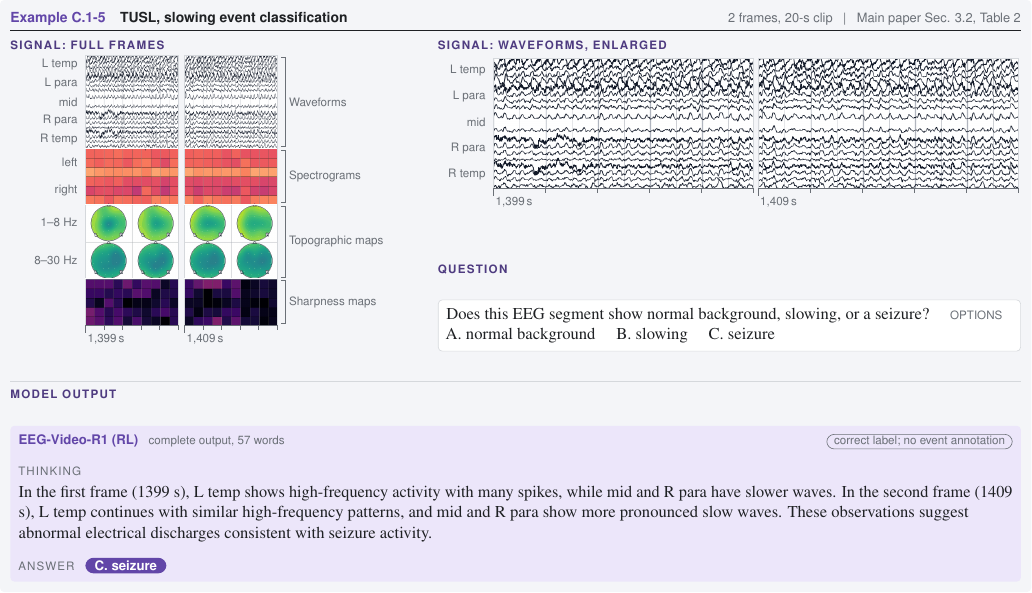}\caption{\emph{Example for Appendix~\ref{app:tasks}.} \textbf{TUSL: the model classifies a 20-second segment as seizure, out of three classes.} The reasoning describes fast activity in the left temporal region, which is visible in both frames.}\label{fig:c1_c1_5}
\end{figure*}

\begin{figure*}[tp]
\centering\includegraphics[width=\textwidth]{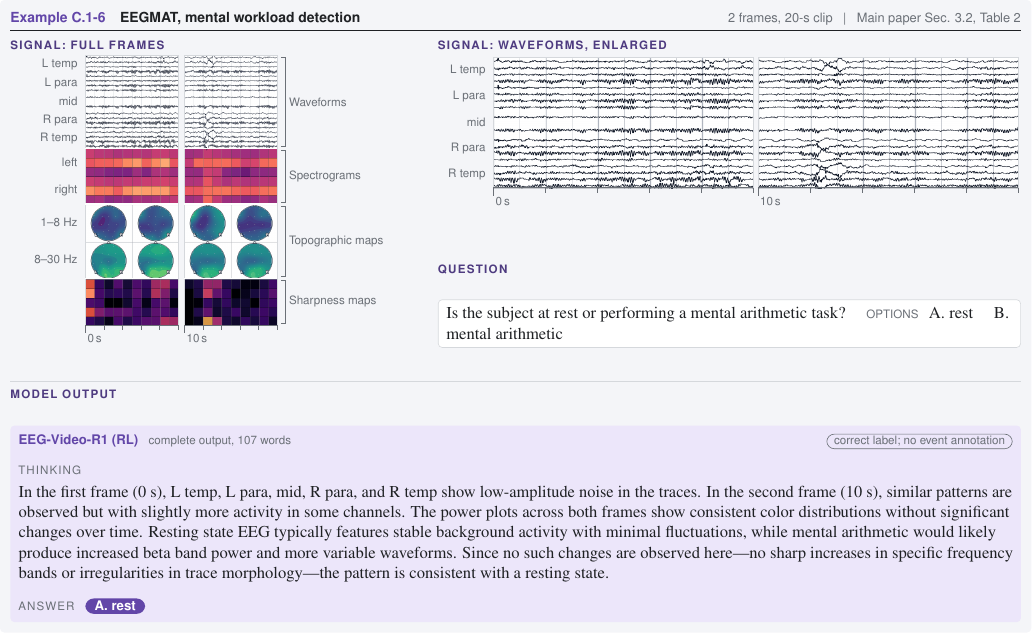}\caption{\emph{Example for Appendix~\ref{app:tasks}.} \textbf{EEGMAT: the model tells rest from mental arithmetic, with no event to point at.} It argues from the \textbf{overall amplitude} and from the \textbf{stability of the band power} across the two frames.}\label{fig:c1_c1_6}
\end{figure*}

\begin{figure*}[tp]
\centering\includegraphics[width=\textwidth]{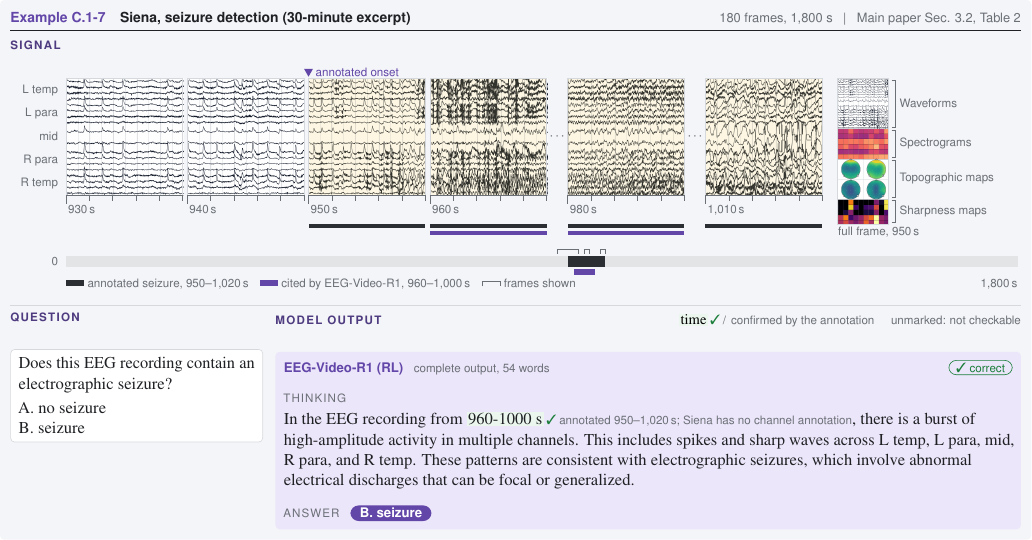}\caption{\emph{Example for Appendix~\ref{app:tasks}.} \textbf{Siena: the model finds a 70-second seizure in a 30-minute excerpt of a dataset never used in training.} The cited interval, \textbf{960--1{,}000\,s}, lies inside the annotated seizure (950--1{,}020\,s). The seizure covers 3.9\% of the 180 frames.}\label{fig:c1_c1_7}
\end{figure*}

\begin{figure*}[tp]
\centering\includegraphics[width=\textwidth]{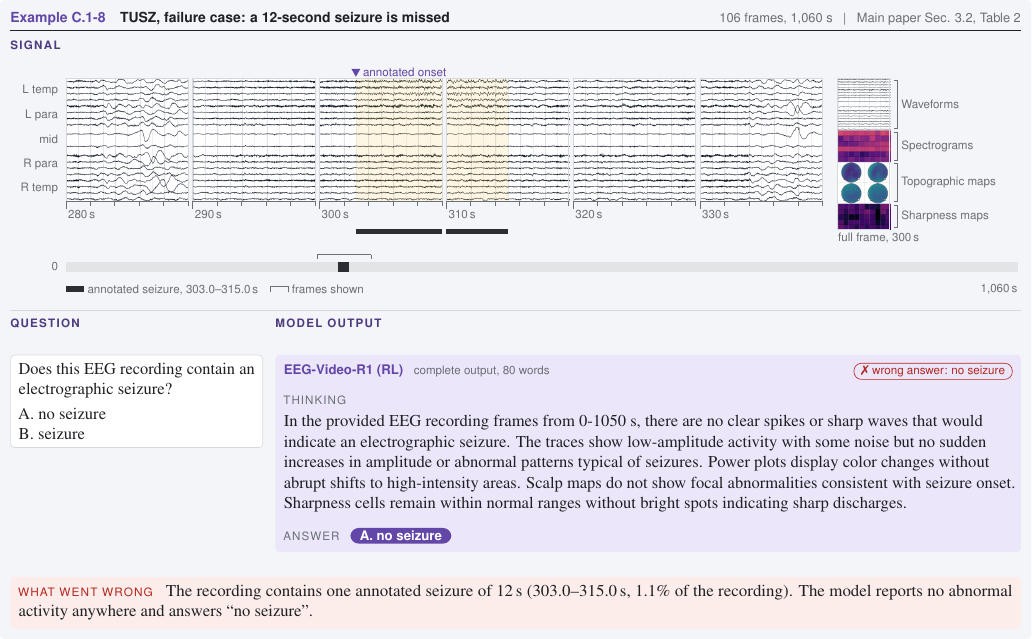}\caption{\emph{Example for Appendix~\ref{app:tasks}.} \textbf{Failure case (TUSZ): the model misses a short seizure in a long recording.} The recording lasts 17.7 minutes and contains one annotated seizure of \textbf{12\,s} (303.0--315.0\,s). The model reports no abnormal activity and answers \textbf{``no seizure''}.}\label{fig:c1_c1_8}
\end{figure*}

\begin{figure*}[tp]
\centering
\includegraphics[width=\textwidth]{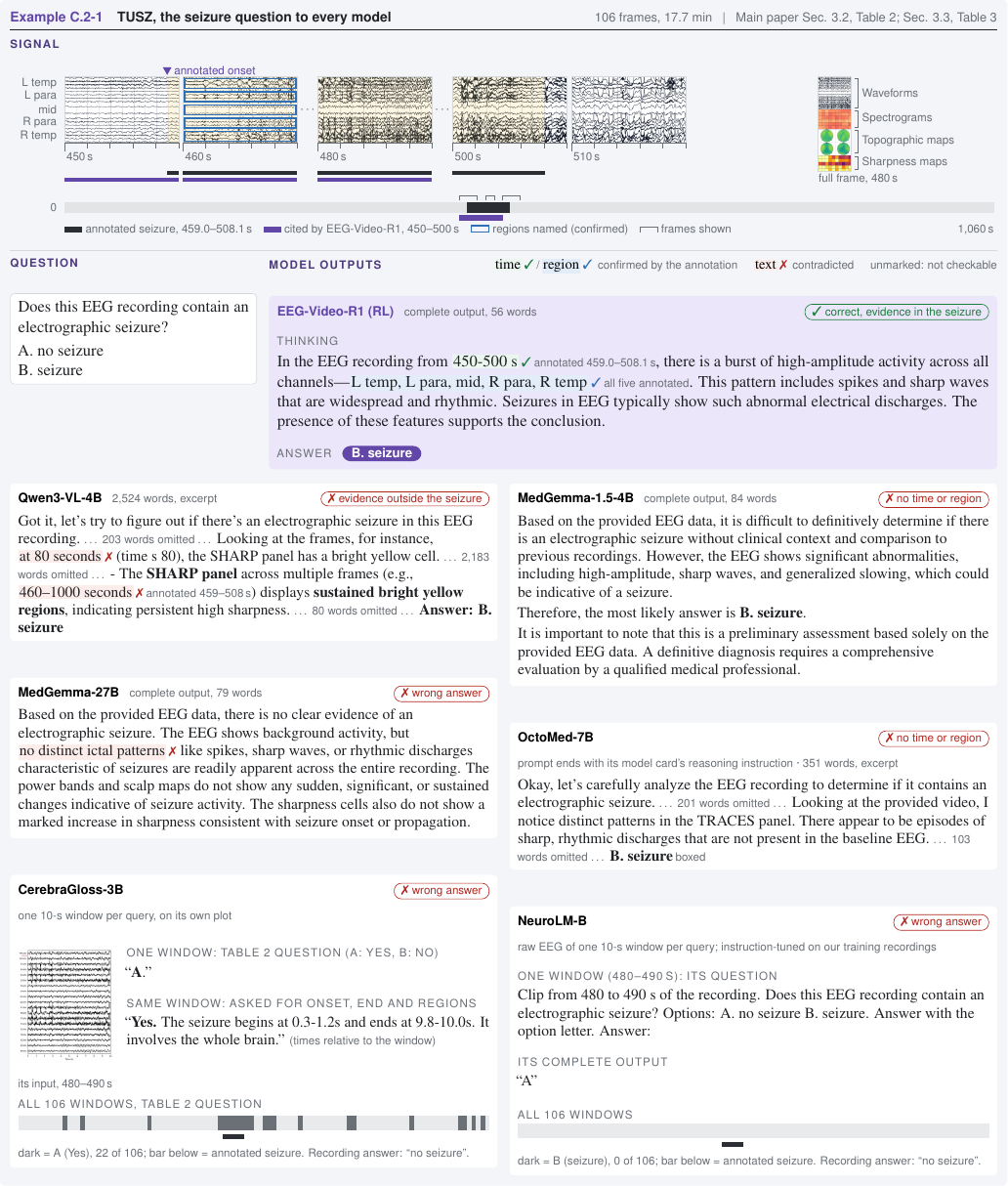}
\caption{\emph{Example for Appendix~\ref{app:final_answer}.} \textbf{Only \ours{} links its answer to the annotated seizure.} Each model shown gets the seizure question of main paper Table~2 on a 17.7-minute TUSZ recording. \ours{} (RL) cites the \textbf{time of the seizure} and the \textbf{five annotated regions}. The zero-shot VLMs cite other times or no time, or answer ``no seizure''. For the \textbf{per-window models} (CerebraGloss-3B, NeuroLM-B), the figure shows the complete output for one window inside the seizure and the answers for all windows.}\label{fig:c2_C2-1}
\end{figure*}

\begin{figure*}[tp]
\centering
\includegraphics[width=\textwidth]{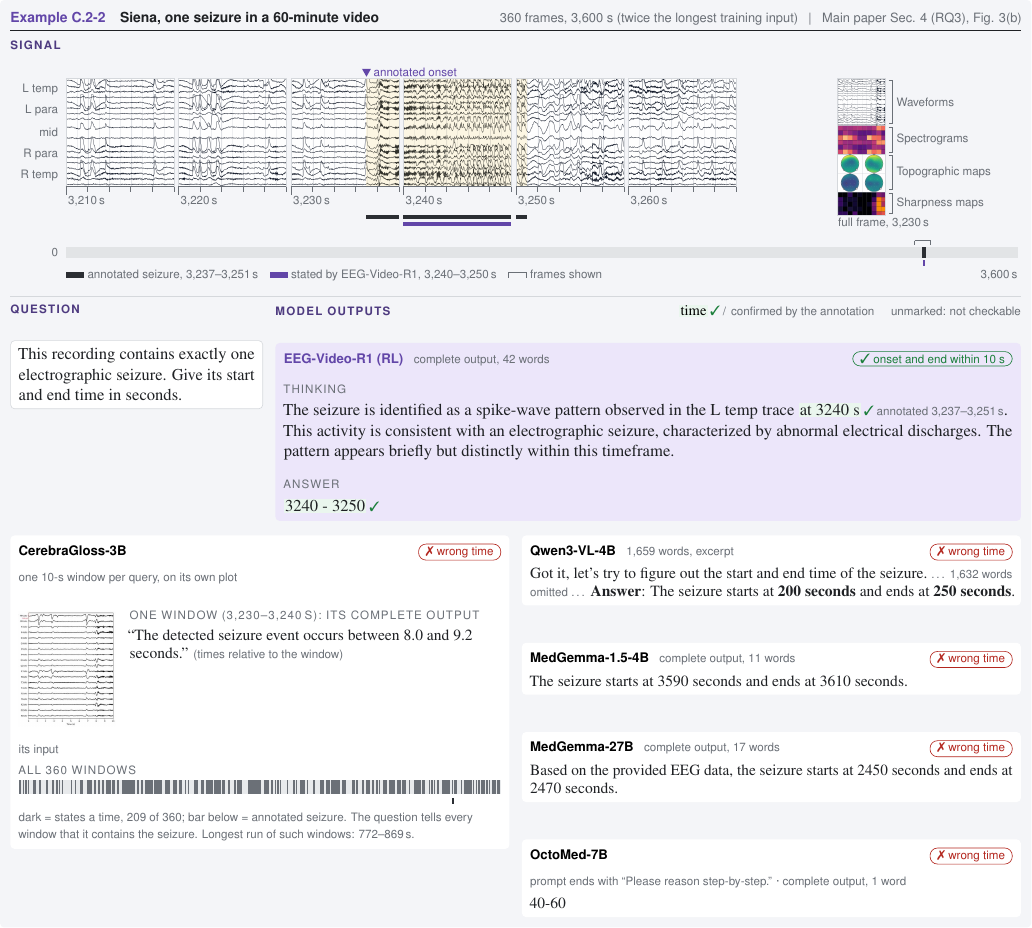}
\caption{\emph{Example for Appendix~\ref{app:final_answer}.} \textbf{A 14-second seizure in a 60-minute video.} The input is a Siena video of 360 frames, twice the longest training input, and the question is that of main paper Fig.~3(b). \ours{} (RL) states the \textbf{onset and end within 4 seconds} of the annotation. Every zero-shot VLM states a time unrelated to the seizure. \textbf{CerebraGloss-3B} is asked window by window: the figure shows its complete output for one window inside the seizure and its answers for all 360 windows, of which 209 state a time.}\label{fig:c2_C2-2}
\end{figure*}

\begin{figure*}[tp]
\centering
\includegraphics[width=\textwidth]{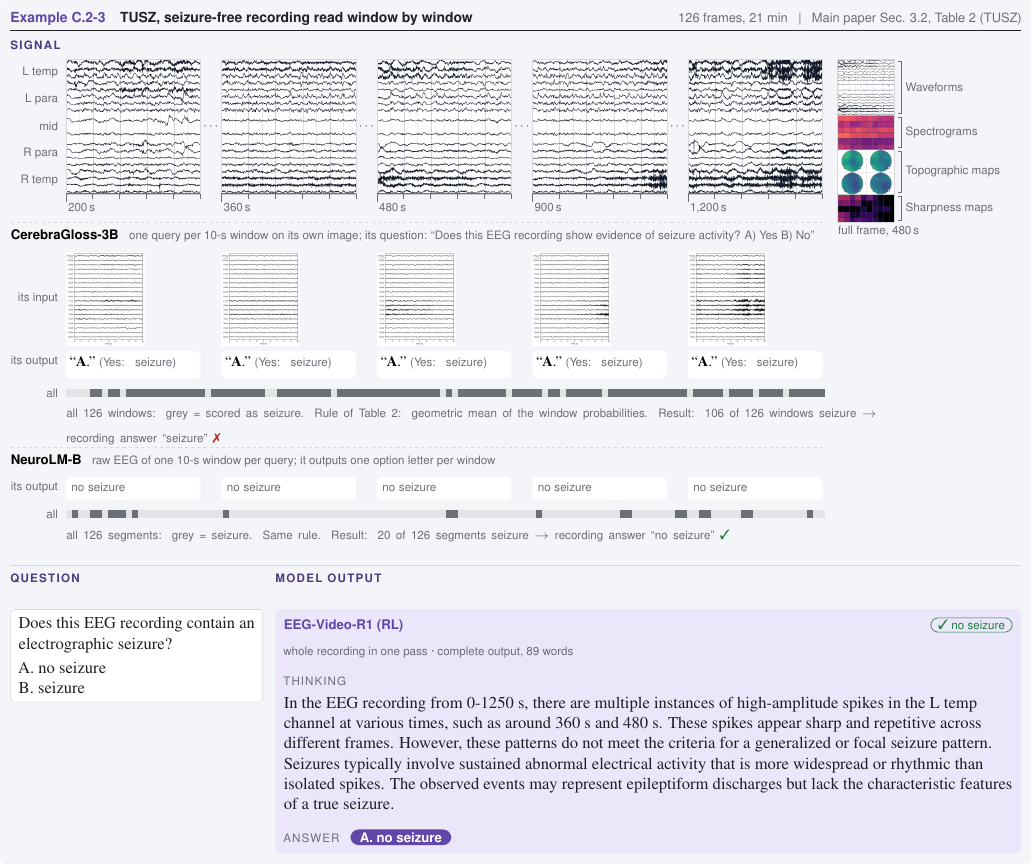}
\caption{\emph{Example for Appendix~\ref{app:final_answer}.} \textbf{Per-window models answer every 10-second window separately.} The input is a seizure-free 21-minute TUSZ recording. Under each displayed window, the figure shows the image CerebraGloss-3B receives and its output for that window, then a bar with its answers for all 126 windows. NeuroLM-B is shown below; it outputs one option per window and no text. CerebraGloss-3B answers ``seizure'' for \textbf{106 of 126 windows}, and its answer for the recording is wrong. NeuroLM-B answers ``seizure'' for 20 windows, and its answer for the recording is right. \ours{} (RL) gives \textbf{one answer for the whole recording}, with its reasoning.}\label{fig:c2_C2-3}
\end{figure*}

\begin{figure*}[tp]
\centering
\includegraphics[width=\textwidth]{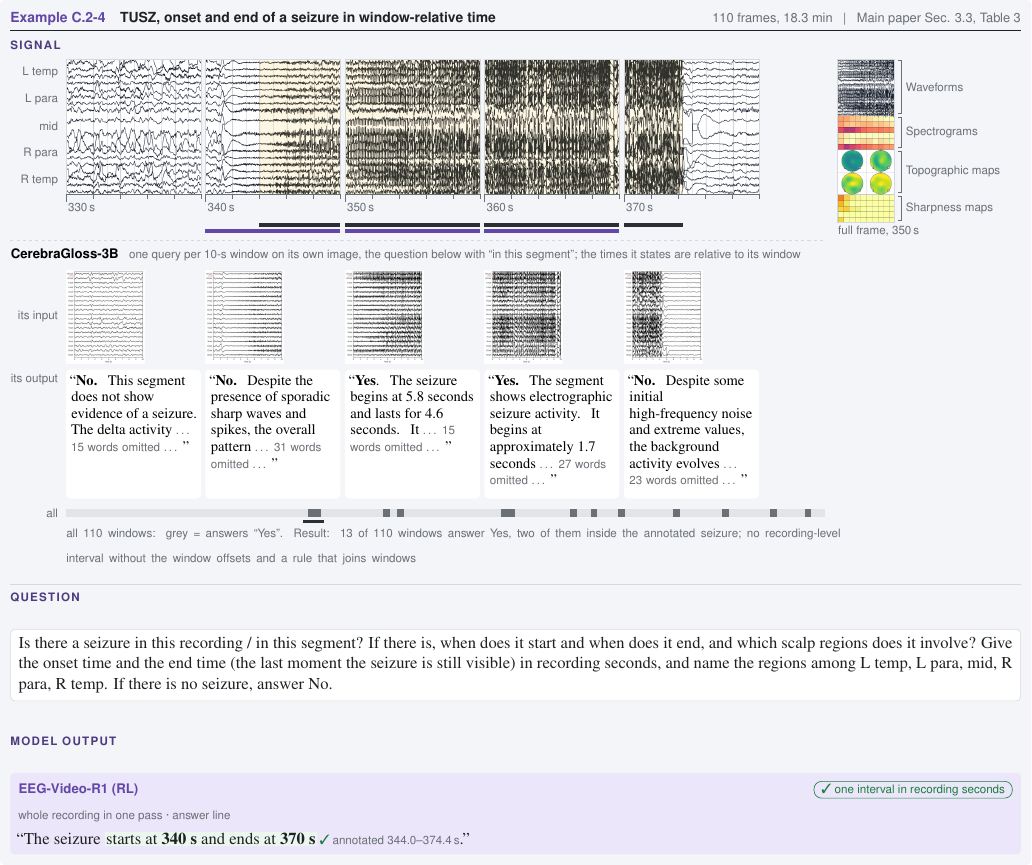}
\caption{\emph{Example for Appendix~\ref{app:final_answer}.} \textbf{A per-window model states times relative to its own window.} The input is an 18.3-minute TUSZ recording with one seizure. CerebraGloss-3B is asked, window by window, the question behind its row of main paper Table~3. It answers ``Yes'' for the two windows in the middle of the seizure and for 11 other windows. It answers ``No'' for the windows that contain the \textbf{onset and the end}. Every time it states is relative to its own 10-second image. \ours{} (RL) states \textbf{one interval in recording seconds}; the figure shows its answer line.}\label{fig:c2_C2-4}
\end{figure*}

\begin{figure*}[tp]
\centering
\includegraphics[width=\textwidth]{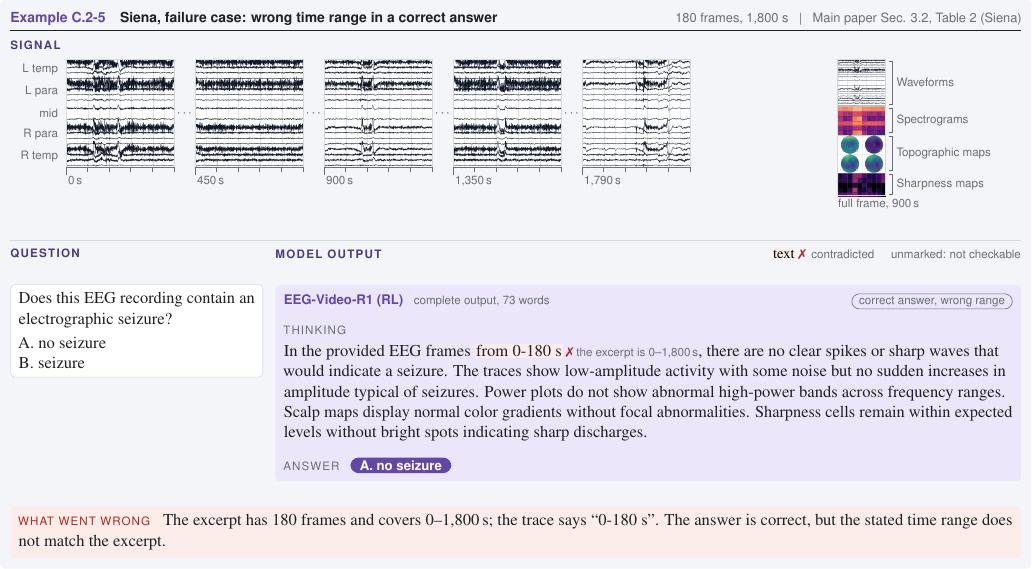}
\caption{\emph{Example for Appendix~\ref{app:final_answer}.} \textbf{Failure case: a correct answer with a wrong time range.} The input is a seizure-free 30-minute Siena excerpt. \ours{} (RL) correctly answers ``no seizure'', but its reasoning gives the range of the frames as \textbf{``0-180 s''}. The excerpt has 180 frames and covers 1,800\,s.}\label{fig:c2_C2-5}
\end{figure*}

\begin{figure*}[tp]
\centering
\includegraphics[width=\textwidth]{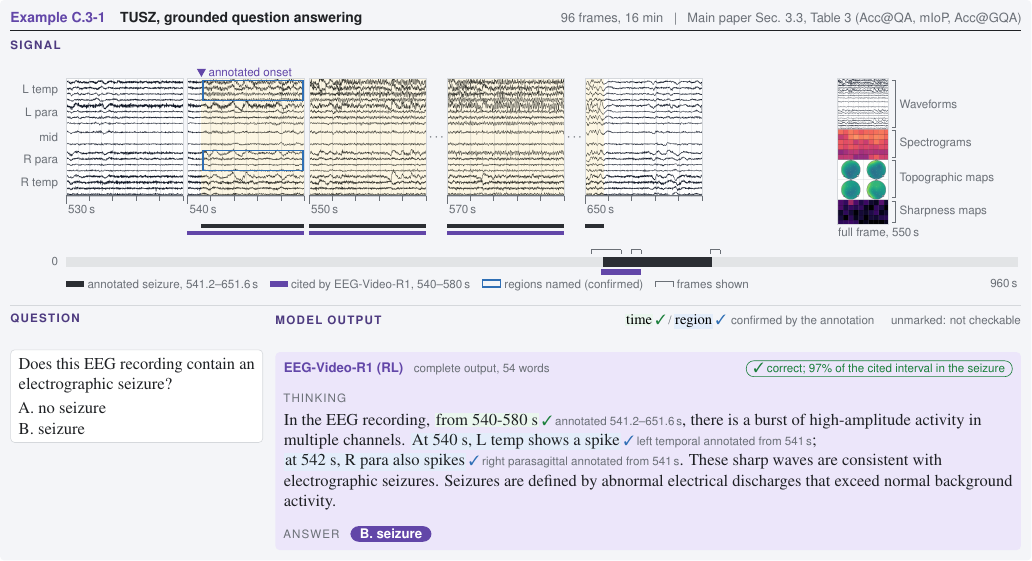}
\caption{\emph{Example for Appendix~\ref{app:grounding}.} \textbf{Asked only whether there is a seizure, the model cites when and where it is.} Grounded question answering on a 16-minute TUSZ recording (main paper Table~3: Acc@QA, mIoP, Acc@GQA). The question asks for no time or region. The reasoning cites 540--580\,s, and \textbf{97\%} of this interval lies inside the annotated seizure (541.2--651.6\,s). It also names the \textbf{left temporal} and \textbf{right parasagittal} regions, both annotated at the onset.}\label{fig:c3_1}
\end{figure*}

\begin{figure*}[tp]
\centering
\includegraphics[width=\textwidth]{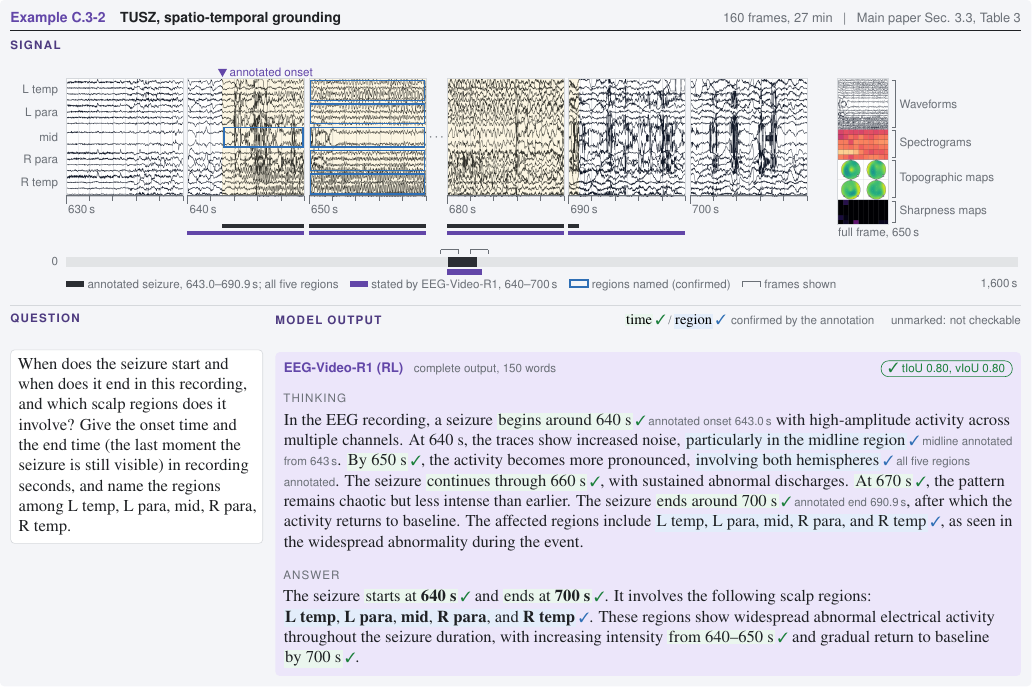}
\caption{\emph{Example for Appendix~\ref{app:grounding}.} \textbf{The model states when a 48-second seizure starts and ends in a 27-minute recording, and where.} Spatio-temporal grounding question on a TUSZ recording (main paper Table~3: m\,tIoU, m\,vIoU). The stated \textbf{onset} (640\,s) and \textbf{end} (700\,s) are 3\,s and 9\,s from the annotation (643.0--690.9\,s), which gives a tIoU of 0.80. The model names all \textbf{five regions}, and all five are annotated.}\label{fig:c3_2}
\end{figure*}

\begin{figure*}[tp]
\centering
\includegraphics[width=\textwidth]{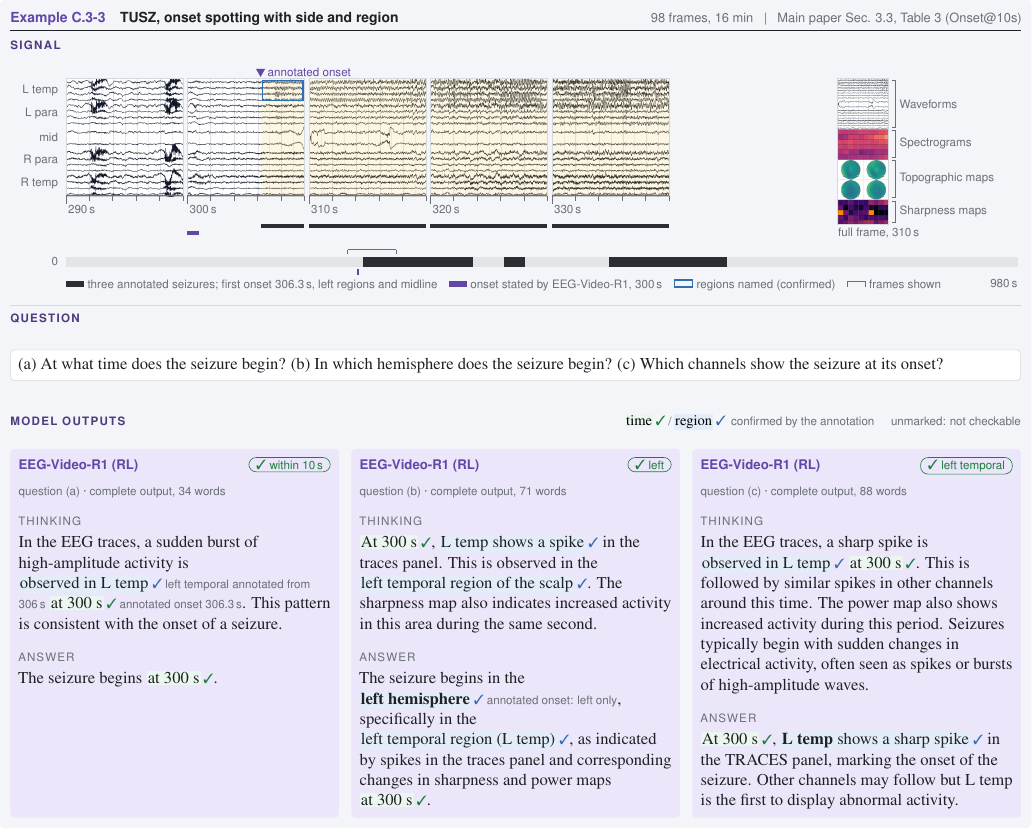}
\caption{\emph{Example for Appendix~\ref{app:grounding}.} \textbf{Three differently worded questions give one consistent onset: 300\,s, left hemisphere, left temporal.} The input is a 16-minute TUSZ recording with a focal seizure, one that starts in a limited area. The questions ask for the time (main paper Table~3: Onset@10s), the hemisphere and the channels of the onset, and each is asked separately. The annotated \textbf{onset} is at 306.3\,s, in the \textbf{left temporal}, left parasagittal and midline regions.}\label{fig:c3_3}
\end{figure*}

\paragraph{Protocol.}
The expert reviewed 25 recordings with a total of 8,820 seconds of EEG. All are cases with a correct final answer, so that the rating reflects the quality of the reasoning. The cases cover the six tasks and are balanced over their classes: two per class for TUAB, TUSZ, TUEP and EEGMAT, and one per class for TUEV and TUSL.
The review was carried out in a web-based tool (Fig.~\ref{fig:e_tool}). For each case, the tool shows the EEG in a clinical viewer with a standard bipolar montage and adjustable sensitivity, time base and filters, next to the video given to the model. It also shows the question with its correct answer, the dataset annotation where one exists, and the reasoning. Each time cited in the reasoning links to that position in the recording, so the expert can check every statement against the signal.
The expert rated the reasoning of each case on a 5-point scale, from 1 (wrong or invented) to 5 (accurate and clinically sound), and could add a comment (Table~\ref{tab:e_scores}).

\begin{table}[t]\centering\small
\caption{\textbf{Expert rating of the reasoning of \ours{}.} Rating scale and number of cases per score, for the 25 reviewed cases. The mean score is 3.72.}\label{tab:e_scores}
\begin{tabular}{@{}clc@{}}\toprule
Score & Meaning & Cases\\\midrule
5 & accurate and clinically sound & 6\\
4 & accurate, minor issues & 10\\
3 & partly right & 6\\
2 & mostly inaccurate & 2\\
1 & wrong or invented & 1\\\bottomrule
\end{tabular}\end{table}

\paragraph{Results.}
The mean score is 3.72 out of 5 (Table~\ref{tab:e_scores}). The expert rated the reasoning as accurate (score 4 or 5) in 16 of the 25 cases (64\%) and as partly right (score 3) in six more (24\%), so the reasoning is at least partially correct in 88\% of the cases. Three cases received a score of 2 or 1. In one of them, a TUEV clip, the reasoning attributes an eye-movement artifact to the wrong scalp regions; the expert noted that such artifacts appear in the frontal channels.


\clearpage

\clearpage

\clearpage
\begin{figure*}[tp]
\centering
\includegraphics[width=\textwidth]{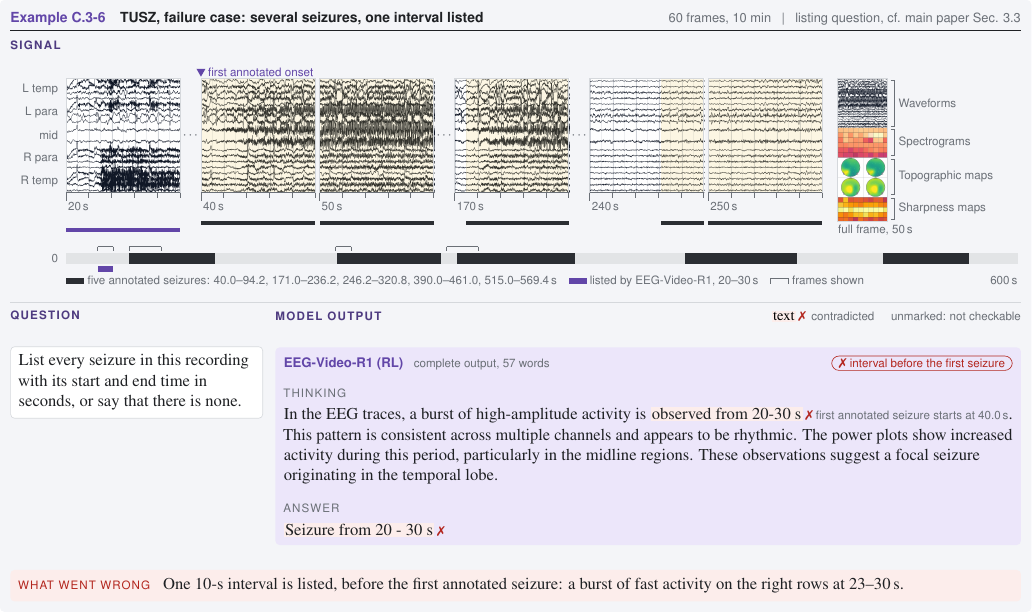}
\caption{\emph{Example for Appendix~\ref{app:grounding}.} \textbf{Failure case: five seizures in one recording, one interval listed.} The model is asked to list every seizure of a 10-minute TUSZ recording. It lists a single 10-second interval (20--30\,s), which lies \textbf{before the first annotated seizure} (40.0\,s).}\label{fig:c3_6}
\end{figure*}

\begin{figure*}[tp]
\centering
\includegraphics[width=\textwidth]{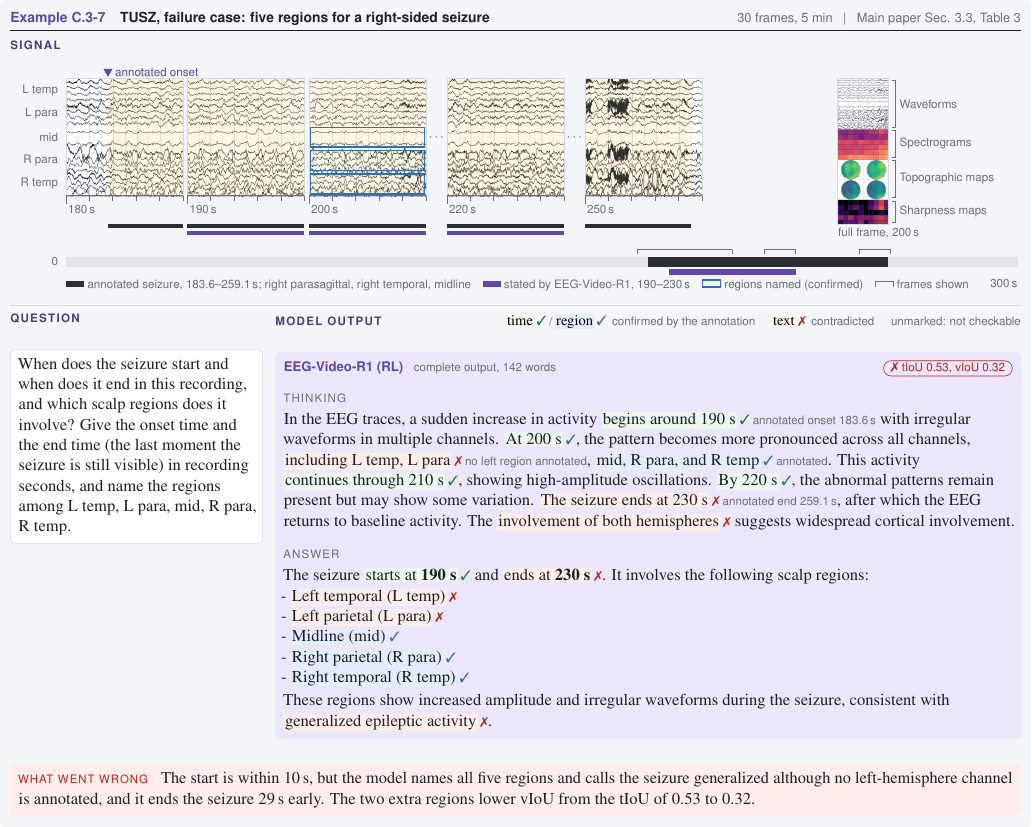}
\caption{\emph{Example for Appendix~\ref{app:grounding}.} \textbf{Failure case: all five regions named for a right-sided seizure.} Spatio-temporal grounding question on a 5-minute TUSZ recording. The stated onset is within 10\,s of the annotation, and the three annotated regions (right parasagittal, right temporal, midline) are named. The model also names both \textbf{left regions}, which are not annotated.}\label{fig:c3_7}
\end{figure*}

\begin{figure*}[tp]
\centering
\includegraphics[width=\textwidth]{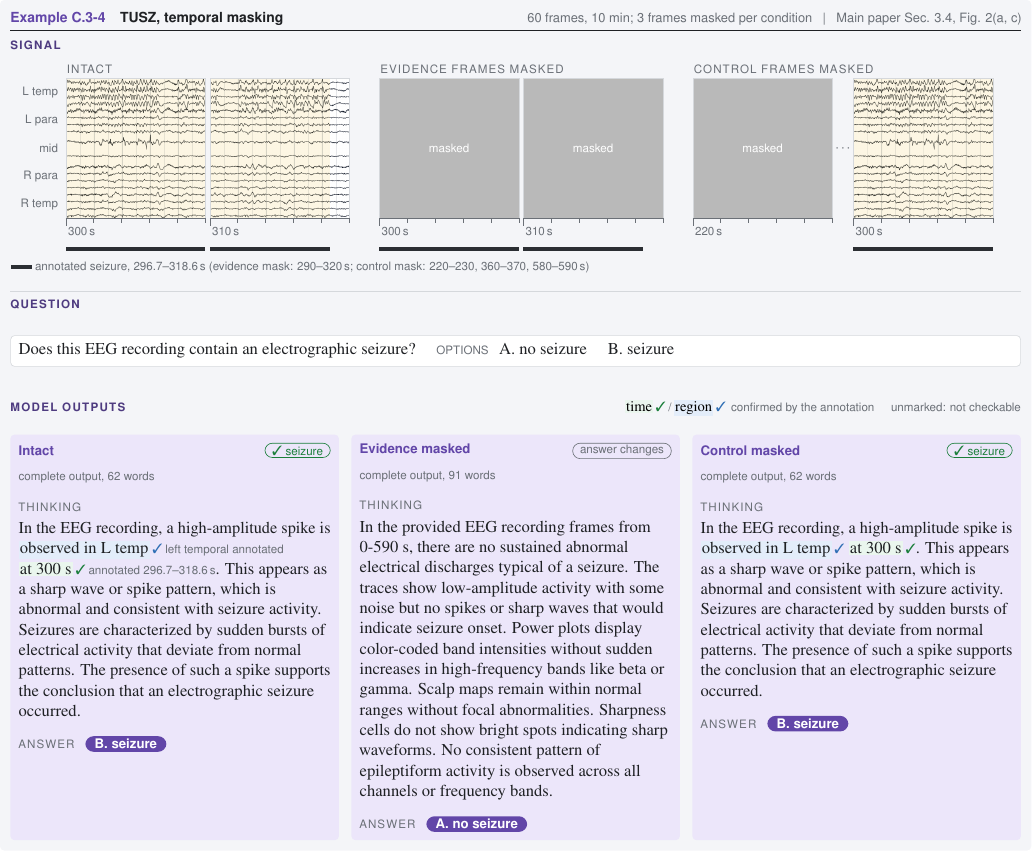}
\caption{\emph{Example for Appendix~\ref{app:exp_counter}.} \textbf{The answer depends on the frames that contain the seizure, not on other frames.} Temporal masking (main paper Fig.~2(a,\,c)) on a 10-minute TUSZ recording with one seizure at 296.7--318.6\,s. On the intact input, the model answers \textbf{seizure} and cites the left temporal region at 300\,s. With the three \textbf{evidence frames} greyed out, it answers \textbf{no seizure}. With three seizure-free \textbf{control frames} greyed out, the answer and the reasoning are unchanged.}\label{fig:c3_4}
\end{figure*}

\begin{figure*}[tp]
\centering
\includegraphics[width=\textwidth]{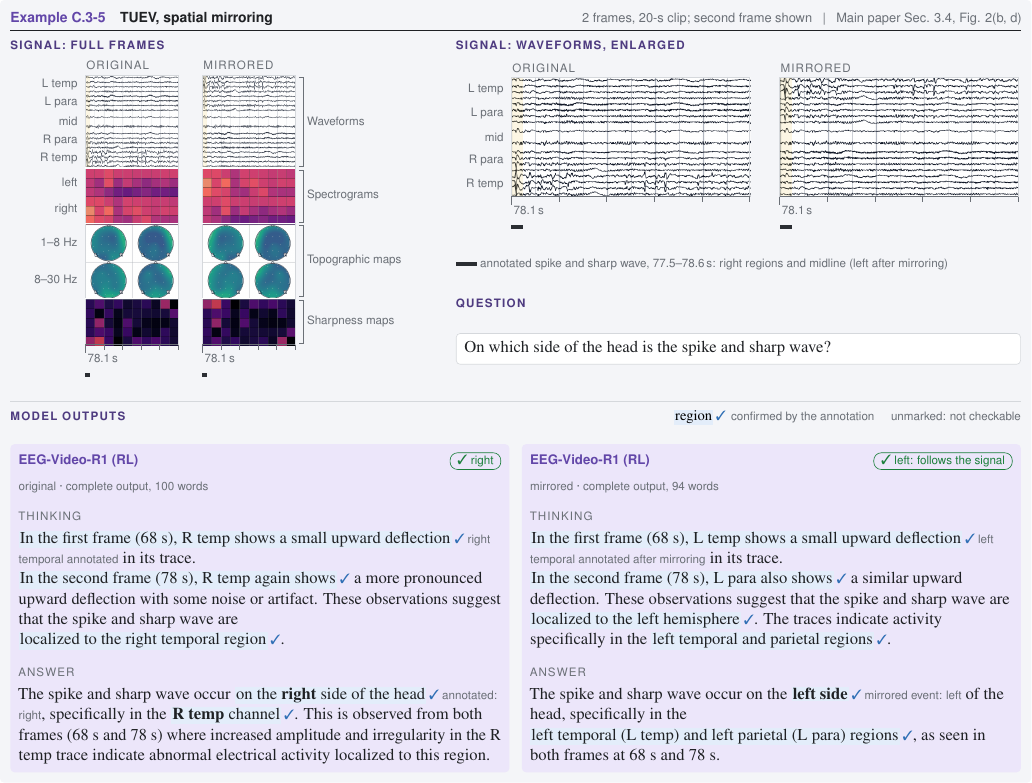}
\caption{\emph{Example for Appendix~\ref{app:exp_counter}.} \textbf{When the left and right electrodes are exchanged, the answer changes side with the signal.} Spatial mirroring (main paper Fig.~2(b,\,d)) on a 20-second TUEV clip with a spike and sharp wave annotated on the right. On the original input, the model answers \textbf{right}, in the right temporal region. On the \textbf{mirrored input}, it answers \textbf{left}, in the left temporal and left parasagittal regions, where the mirrored annotation lies. The item therefore counts as correct in the paired accuracy of Fig.~2(d).}\label{fig:c3_5}
\end{figure*}

\begin{figure*}[tp]
\centering
\includegraphics[width=\textwidth]{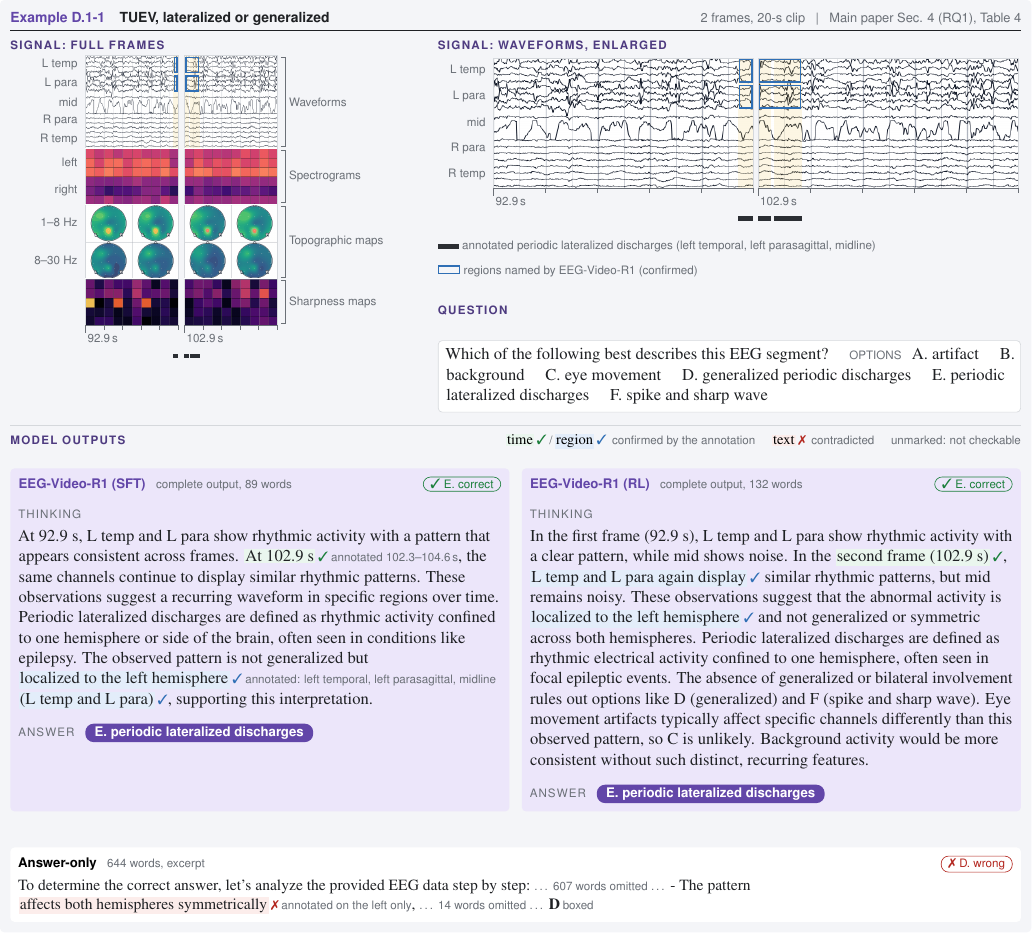}
\caption{\emph{Example for Appendix~\ref{app:rq1}.} \textbf{With reasoning, the model tells one-sided (lateralized) from whole-head (generalized) discharges by locating them.} A 20-second TUEV clip with periodic discharges (a waveform that repeats at regular intervals) annotated on the \textbf{left side}. Both \ours{} models place the pattern in the left temporal and left parasagittal regions, in the frame that contains the annotated events, and answer correctly. The \textbf{Answer-only} variant calls the pattern symmetric and answers generalized.}
\label{fig:d1_d1_1}
\end{figure*}

\begin{figure*}[tp]
\centering
\includegraphics[width=\textwidth]{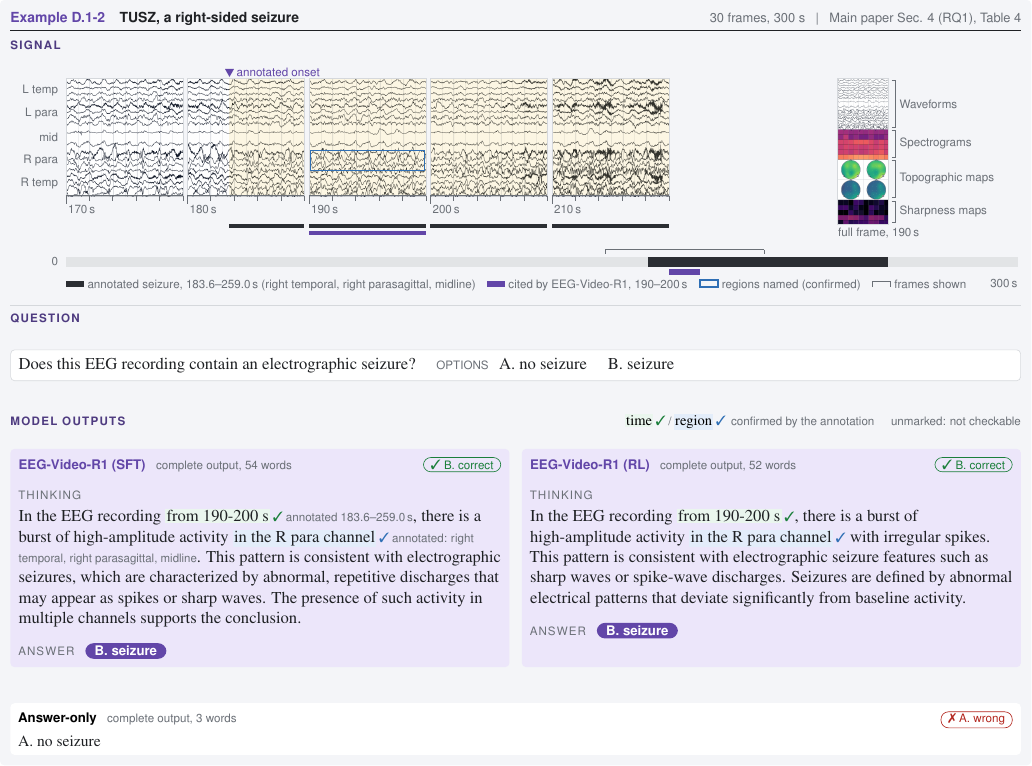}
\caption{\emph{Example for Appendix~\ref{app:rq1}.} \textbf{With reasoning, the model states when and where the seizure is; the Answer-only variant misses it.} A 5-minute TUSZ recording with one annotated seizure on the \textbf{right side}. Both \ours{} models cite a frame \textbf{inside the seizure} and the right parasagittal region, where the seizure is annotated. The Answer-only variant outputs only its answer: no seizure.}
\label{fig:d1_d1_2}
\end{figure*}

\begin{figure*}[tp]
\centering
\includegraphics[width=\textwidth]{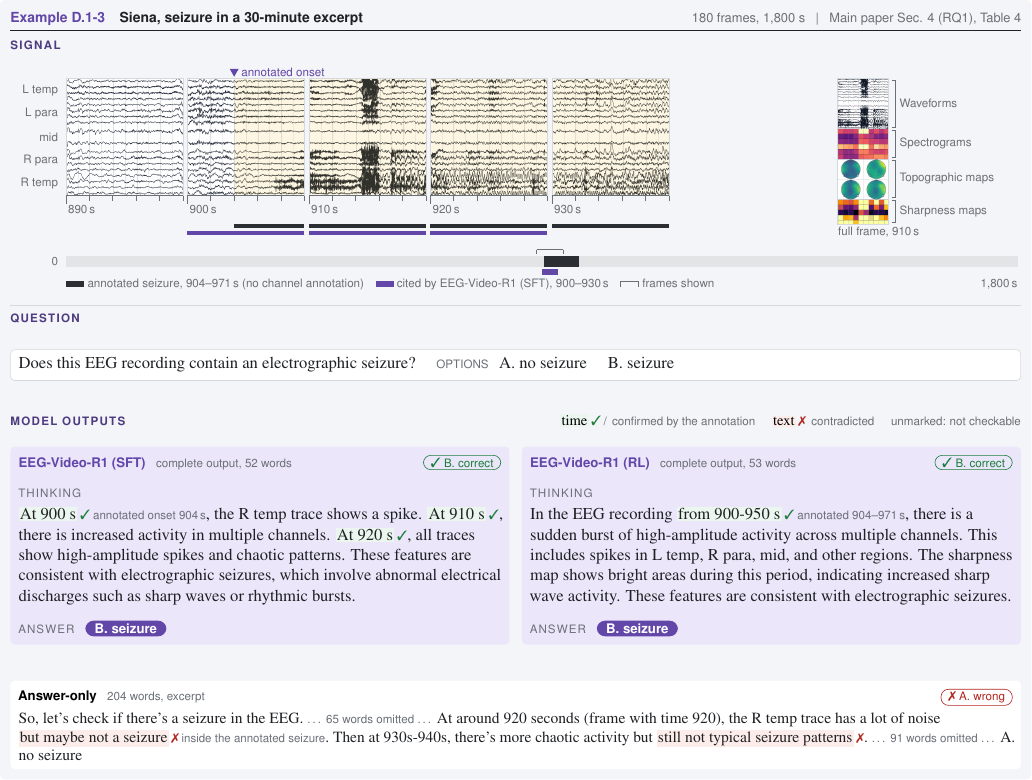}
\caption{\emph{Example for Appendix~\ref{app:rq1}.} \textbf{On the external Siena dataset, \ours{} cites the seizure; the Answer-only variant looks at the seizure frames and dismisses them.} A 30-minute Siena excerpt with one annotated seizure. Both \ours{} models cite the frames in which the \textbf{annotated seizure starts}. The Answer-only variant writes free-form text. It examines 920--940\,s, which lies inside the seizure, and attributes the activity there to \textbf{noise}.}
\label{fig:d1_d1_3}
\end{figure*}

\begin{figure*}[tp]
\centering
\includegraphics[width=\textwidth]{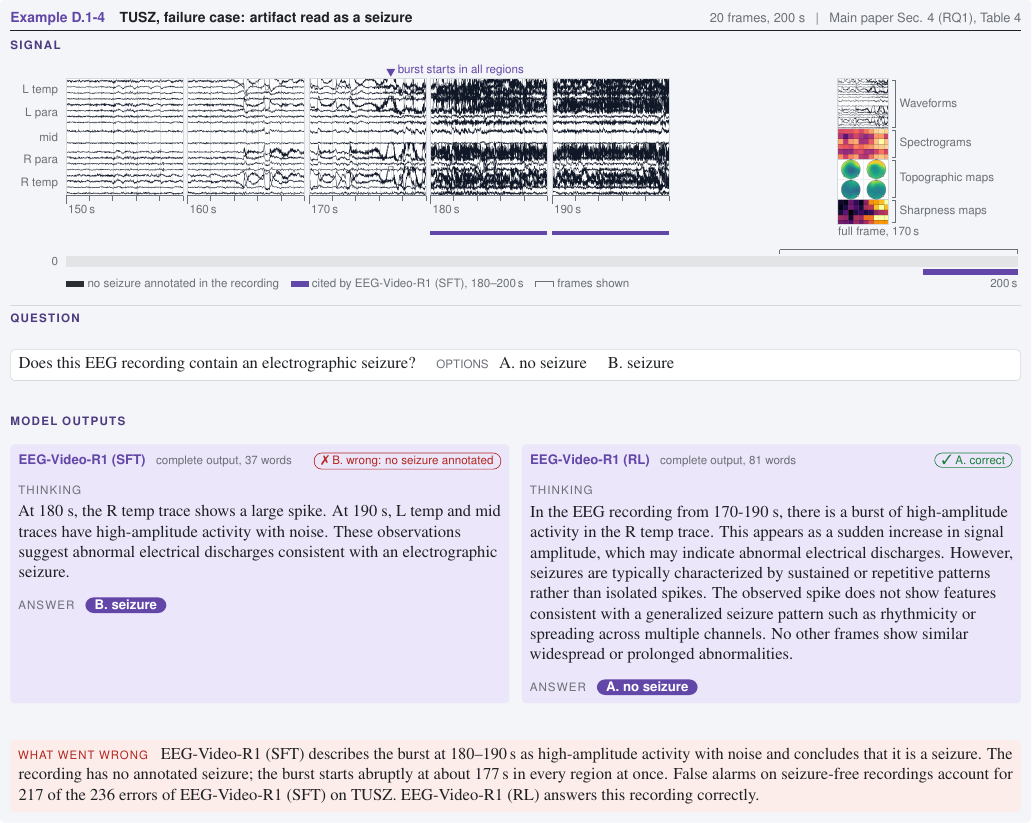}
\caption{\emph{Example for Appendix~\ref{app:rq1}.} \textbf{Failure case: \ours{} (SFT) reads an artifact (a disturbance not caused by brain activity) as a seizure; \ours{} (RL) answers correctly.} A 200-second TUSZ recording with \textbf{no annotated seizure}. A burst of large-amplitude activity starts suddenly in all regions at once. \ours{} (SFT) describes the burst and concludes that it is a seizure. \ours{} (RL) notes that the burst is not sustained and answers ``no seizure''.}
\label{fig:d1_d1_4}
\end{figure*}

\clearpage
\begin{figure*}[tp]
\centering
\includegraphics[width=\textwidth]{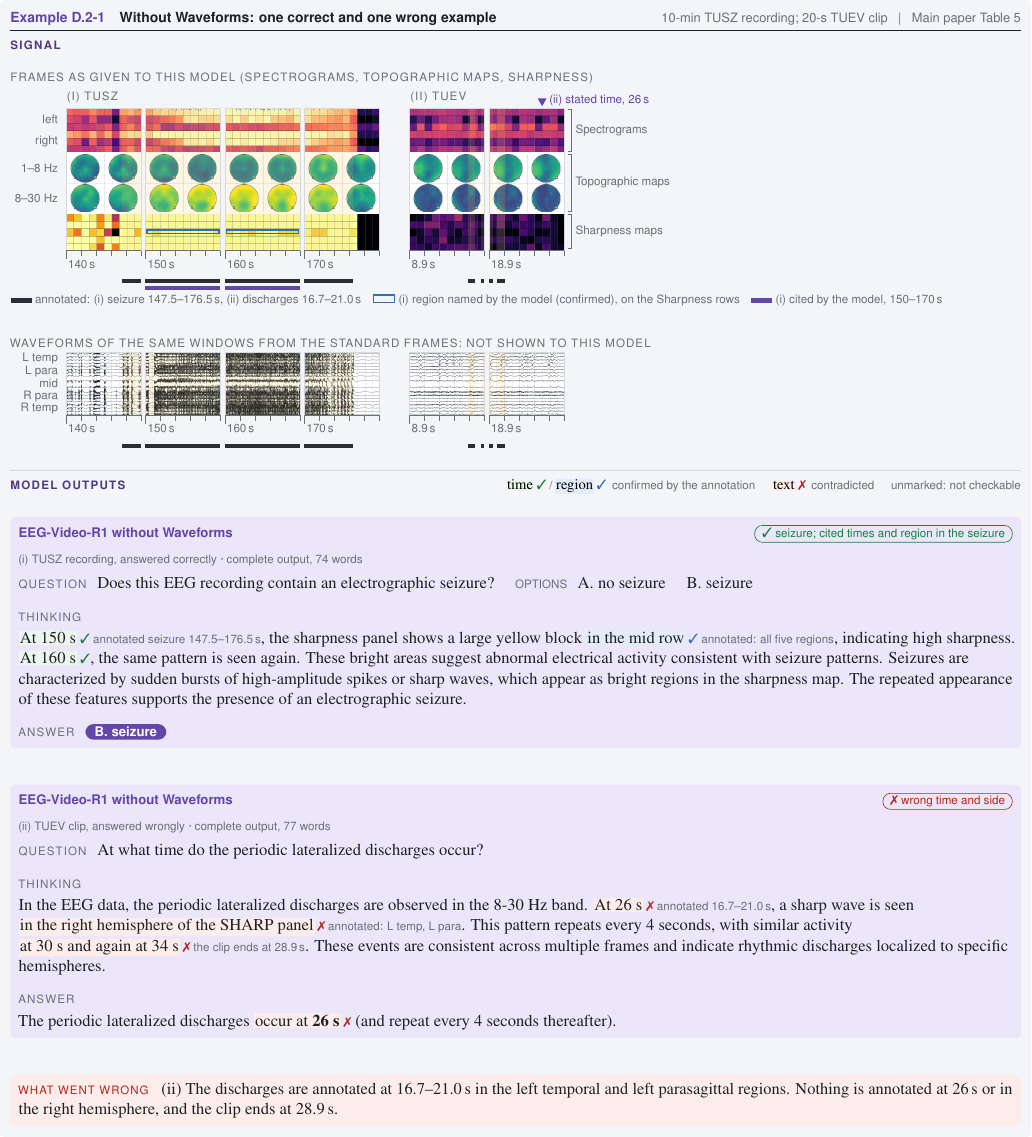}
\caption{\emph{Example for Appendix~\ref{app:rq2}.} \textbf{Without Waveforms, the variant finds a seizure and cites correct evidence, but puts brief periodic discharges at the wrong time and on the wrong side.} The upper strip shows the frames this variant receives (Spectrograms, Topographic maps, Sharpness). The lower strip shows the Waveforms of the same windows, which the variant does not receive. \textbf{(i) Correct example.} In a 10-minute TUSZ recording, the variant answers \textbf{seizure}. It cites 150\,s, 160\,s and the \textbf{midline} row of the Sharpness panel, all of which agree with the annotated seizure (147.5--176.5\,s). \textbf{(ii) Wrong example.} In a 20-second TUEV clip, discharges are annotated at 16.7--21.0\,s in the left temporal and left parasagittal regions. The variant states \textbf{26\,s} and the \textbf{right hemisphere}, and also gives times after the clip ends.}
\label{fig:d2_1}
\end{figure*}

\begin{figure*}[tp]
\centering
\includegraphics[width=\textwidth]{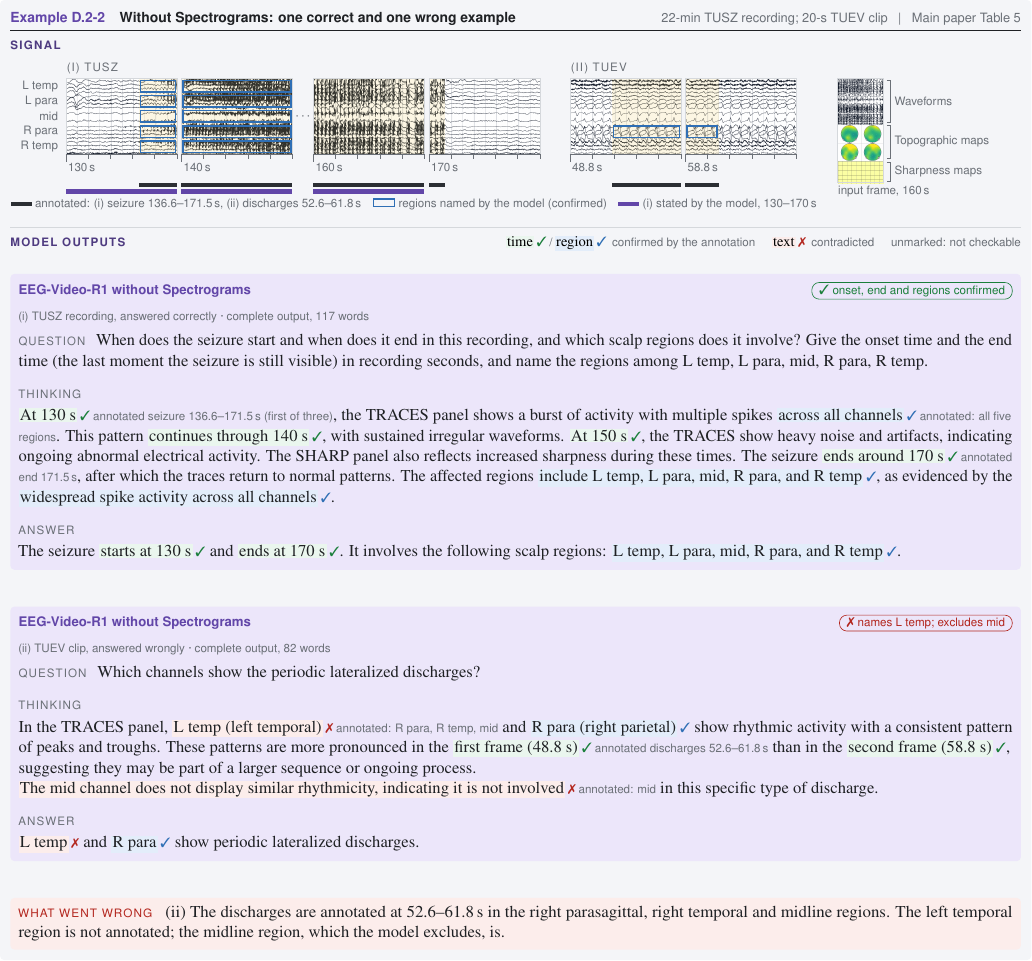}
\caption{\emph{Example for Appendix~\ref{app:rq2}.} \textbf{Without Spectrograms, the variant gives the correct start and end of a whole-head (generalized) seizure, but places right-sided periodic discharges partly on the left.} \textbf{(i) Correct example.} In a 22-minute TUSZ recording with three seizures, the variant states \textbf{130--170\,s} and all five regions for the first one. That seizure is annotated at 136.6--171.5\,s in all five regions. \textbf{(ii) Wrong example.} In a 20-second TUEV clip, discharges are annotated at 52.6--61.8\,s in the right parasagittal, right temporal and midline regions. The variant names the right parasagittal region but also the \textbf{left temporal} region, and states that the \textbf{midline} is not involved.}
\label{fig:d2_2}
\end{figure*}

\begin{figure*}[tp]
\centering
\includegraphics[width=\textwidth]{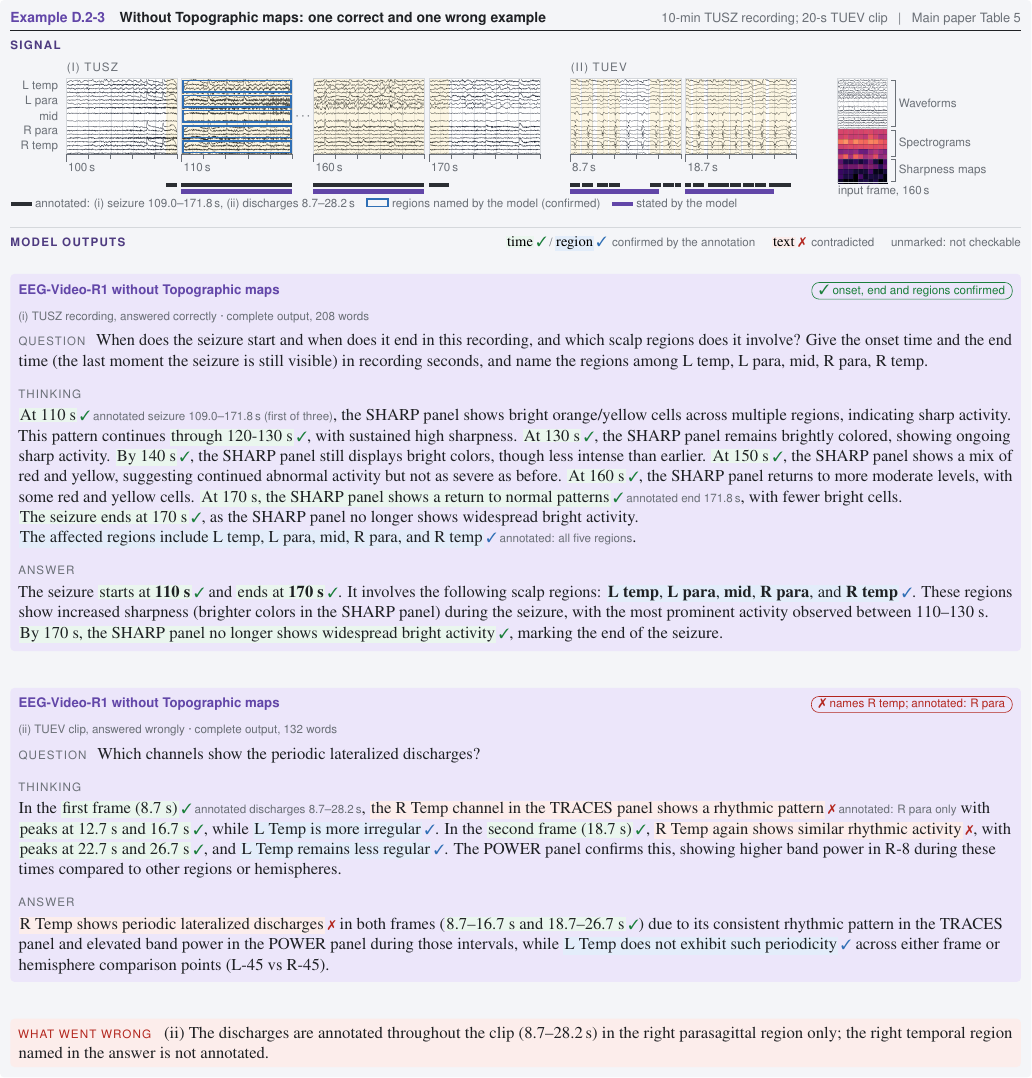}
\caption{\emph{Example for Appendix~\ref{app:rq2}.} \textbf{Without Topographic maps, the variant gives the correct start and end of a seizure, but names the wrong scalp region for periodic discharges confined to one region.} \textbf{(i) Correct example.} In a 10-minute TUSZ recording with three seizures, the variant states \textbf{110--170\,s} and all five regions for the first one. That seizure is annotated at 109.0--171.8\,s in all five regions. \textbf{(ii) Wrong example.} In a 20-second TUEV clip, discharges are annotated throughout the clip in the \textbf{right parasagittal} region only. The variant names the adjacent \textbf{right temporal} region instead. The times it gives lie within the annotated discharges.}
\label{fig:d2_3}
\end{figure*}

\begin{figure*}[tp]
\centering
\includegraphics[width=\textwidth]{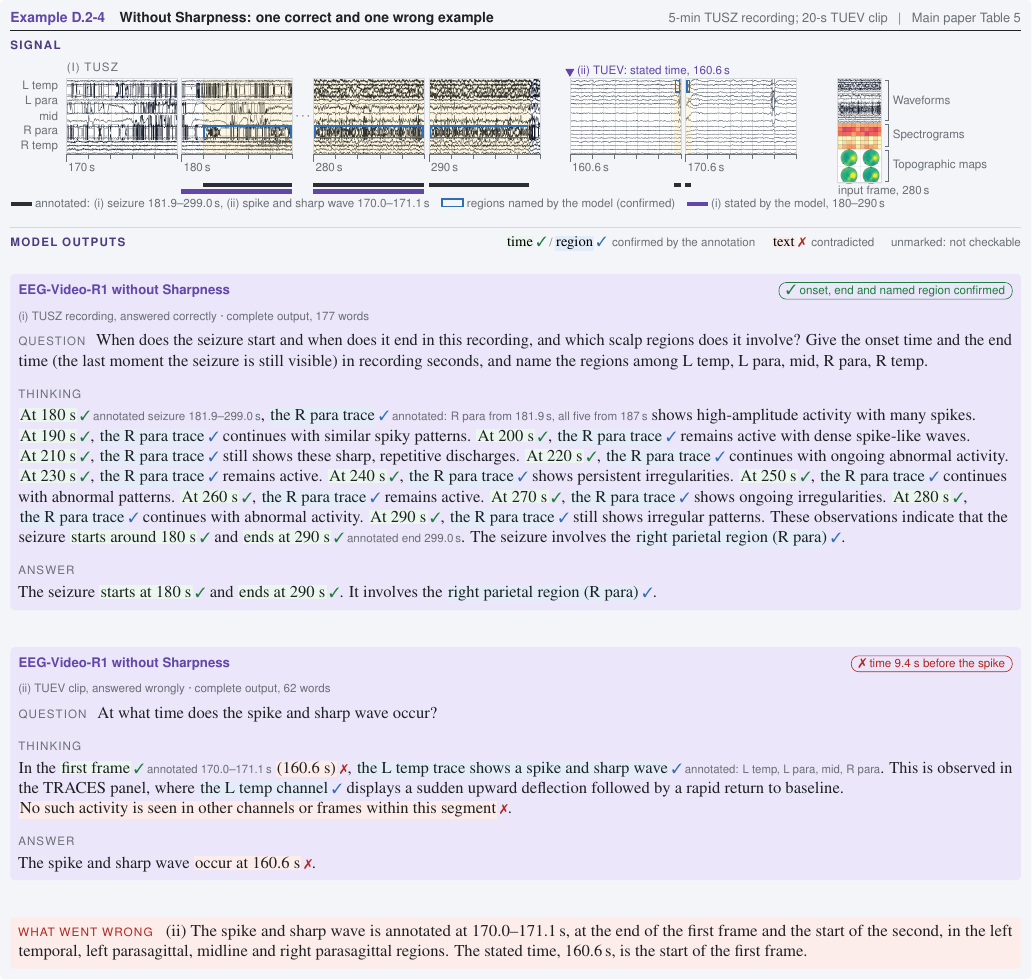}
\caption{\emph{Example for Appendix~\ref{app:rq2}.} \textbf{Without Sharpness, the variant gives the start and end of a seizure within 10\,s, but gives the start of a frame as the time of a spike that occurs at its end.} \textbf{(i) Correct example.} In a 5-minute TUSZ recording, a seizure is annotated at 181.9--299.0\,s. It starts in the right parasagittal region and spreads to all five regions. The variant states \textbf{180--290\,s} and the right parasagittal region. \textbf{(ii) Wrong example.} In a 20-second TUEV clip, a spike and sharp wave (a brief, pointed waveform) is annotated at 170.0--171.1\,s. The variant names the annotated \textbf{left temporal} region and the first frame, but it gives \textbf{160.6\,s}, the start of that frame, as the time of the event.}
\label{fig:d2_4}
\end{figure*}

\begin{figure*}[tp]
\centering\includegraphics[width=\textwidth]{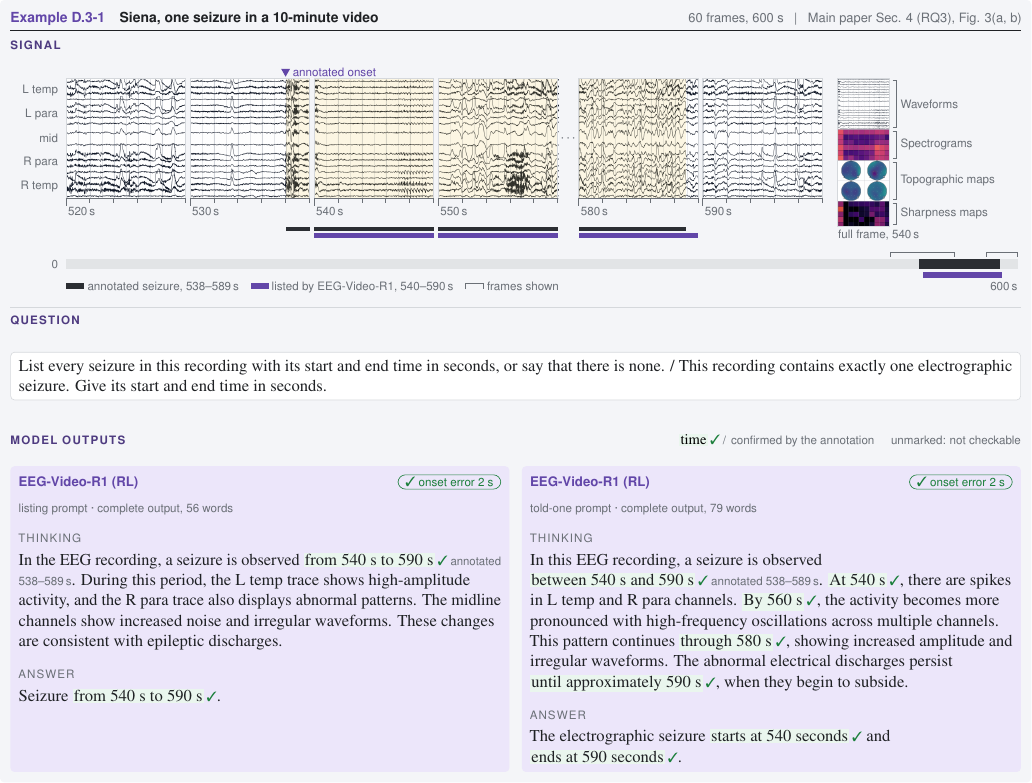}\caption{\emph{Example for Appendix~\ref{app:rq3}.} \textbf{Needle-in-a-haystack, 10 minutes: the seizure is listed with its start and end.} Both prompts return \textbf{540--590\,s} for a seizure annotated at 538--589\,s, an \textbf{onset error of 2\,s}. The left card shows the listing prompt and the right card the told-one prompt, here and in the following needle figures.}\label{fig:d3_d3_1}
\end{figure*}

\begin{figure*}[tp]
\centering\includegraphics[width=\textwidth]{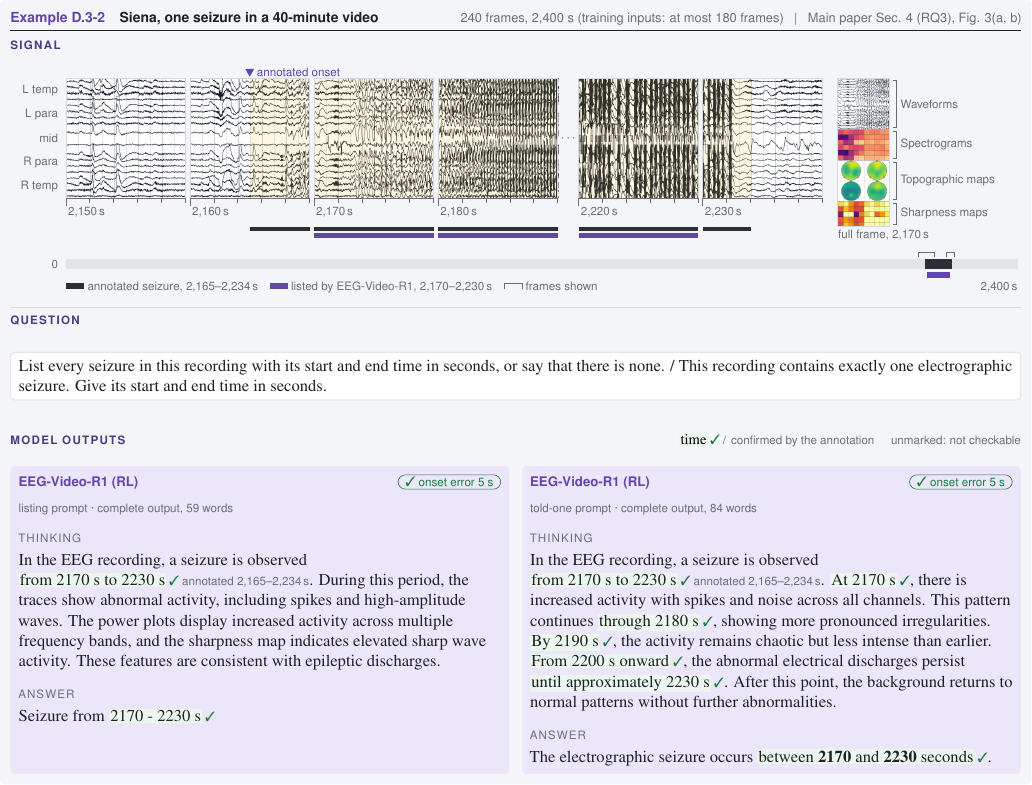}\caption{\emph{Example for Appendix~\ref{app:rq3}.} \textbf{Needle-in-a-haystack, 40 minutes: the seizure is located in a video longer than any training input.} Both prompts return \textbf{2{,}170--2{,}230\,s} for a seizure annotated at 2{,}165--2{,}234\,s, 36 minutes into the 240-frame video. The stated \textbf{onset and end} are both within 5\,s of the annotation.}\label{fig:d3_d3_2}
\end{figure*}

\begin{figure*}[tp]
\centering\includegraphics[width=\textwidth]{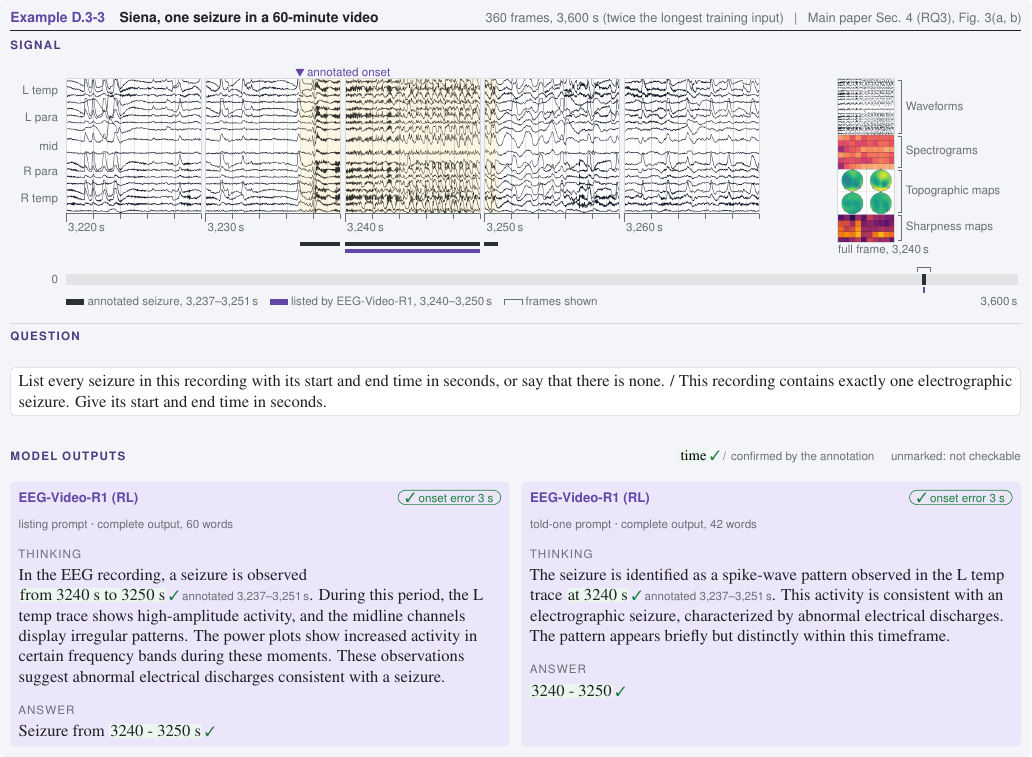}\caption{\emph{Example for Appendix~\ref{app:rq3}.} \textbf{Needle-in-a-haystack, 60 minutes: a 14-second seizure is found in a video twice as long as any training input.} The seizure (3{,}237--3{,}251\,s) takes up 0.4\% of the 360-frame video. Both prompts return \textbf{3{,}240--3{,}250\,s}, an \textbf{onset error of 3\,s}.}\label{fig:d3_d3_3}
\end{figure*}

\clearpage
\begin{figure*}[tp]
\centering\includegraphics[width=\textwidth]{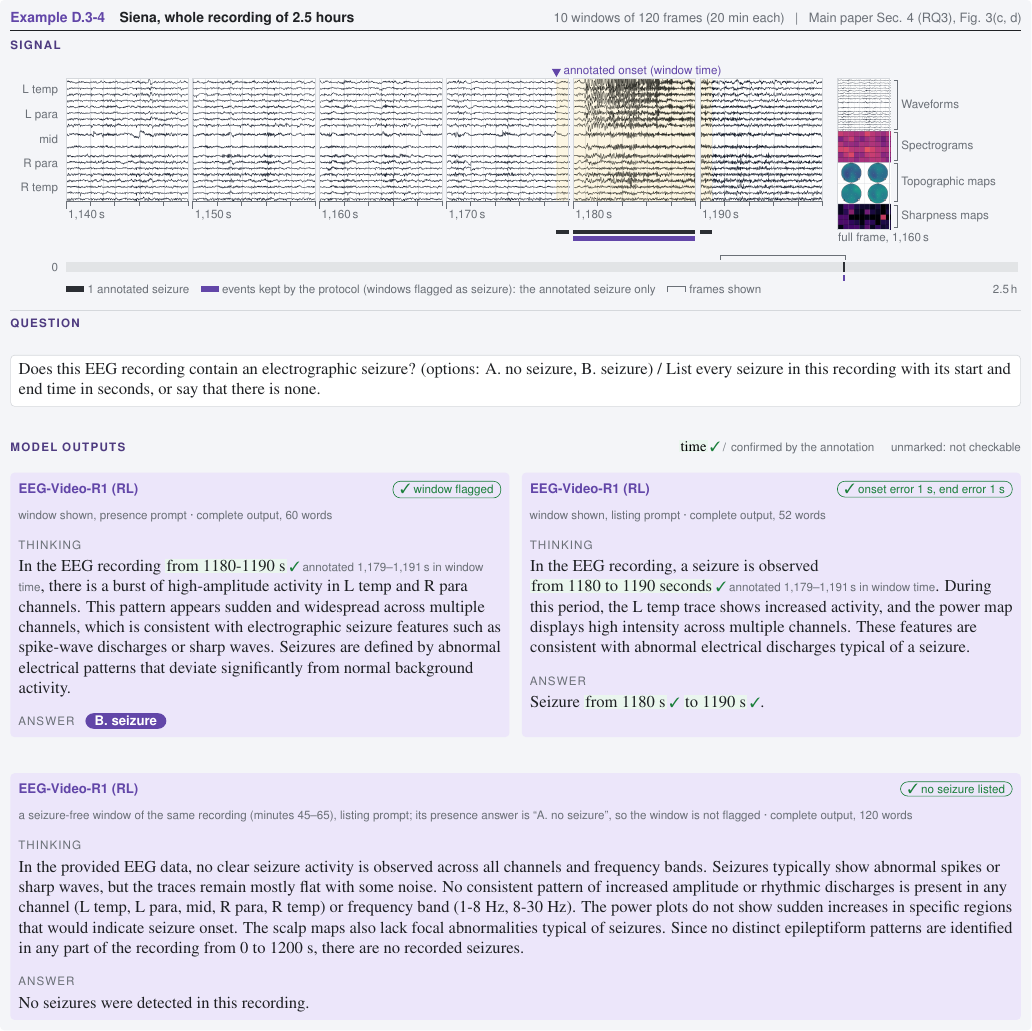}\caption{\emph{Example for Appendix~\ref{app:rq3}.} \textbf{Whole recording: in a 2.5-hour recording read window by window, the seizure is found, with its start and end within 1\,s.} The recording is read as overlapping 20-minute windows. Only events from windows that the model flags as containing a seizure are kept. The model \textbf{flags the window shown} and lists \textbf{1{,}180--1{,}190\,s} for a seizure annotated at 1{,}179--1{,}191\,s of that window. The bottom card shows a \textbf{seizure-free window} of the same recording, in which the model lists no seizure and does not flag the window.}\label{fig:d3_d3_4}
\end{figure*}

\begin{figure*}[tp]
\centering\includegraphics[width=\textwidth]{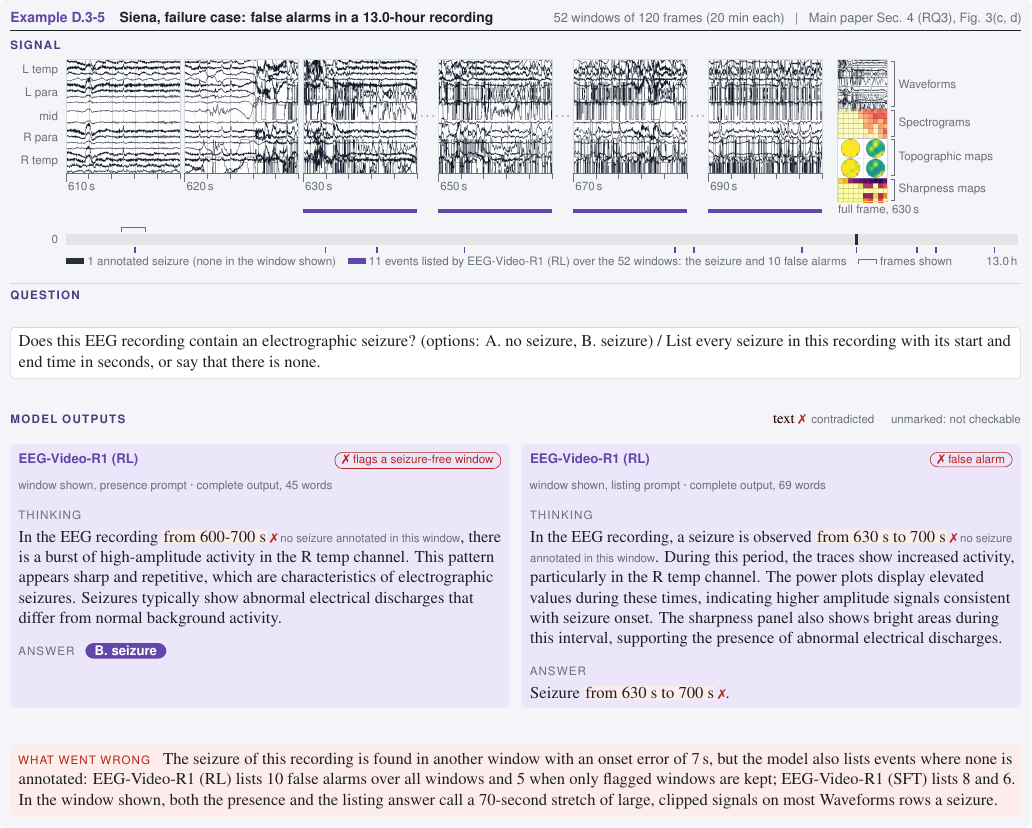}\caption{\emph{Example for Appendix~\ref{app:rq3}.} \textbf{Failure case (whole recording, 13.0 hours): the seizure is found, but false alarms are listed as well.} The timeline shows the annotated seizure and the events listed across the 52 windows. The frames show a \textbf{seizure-free window} in which the model reports a seizure during a 70-second stretch of very large signals: \textbf{an artifact is read as a seizure.}}\label{fig:d3_d3_5}
\end{figure*}

\begin{figure*}[tp]
\centering
\includegraphics[width=\textwidth]{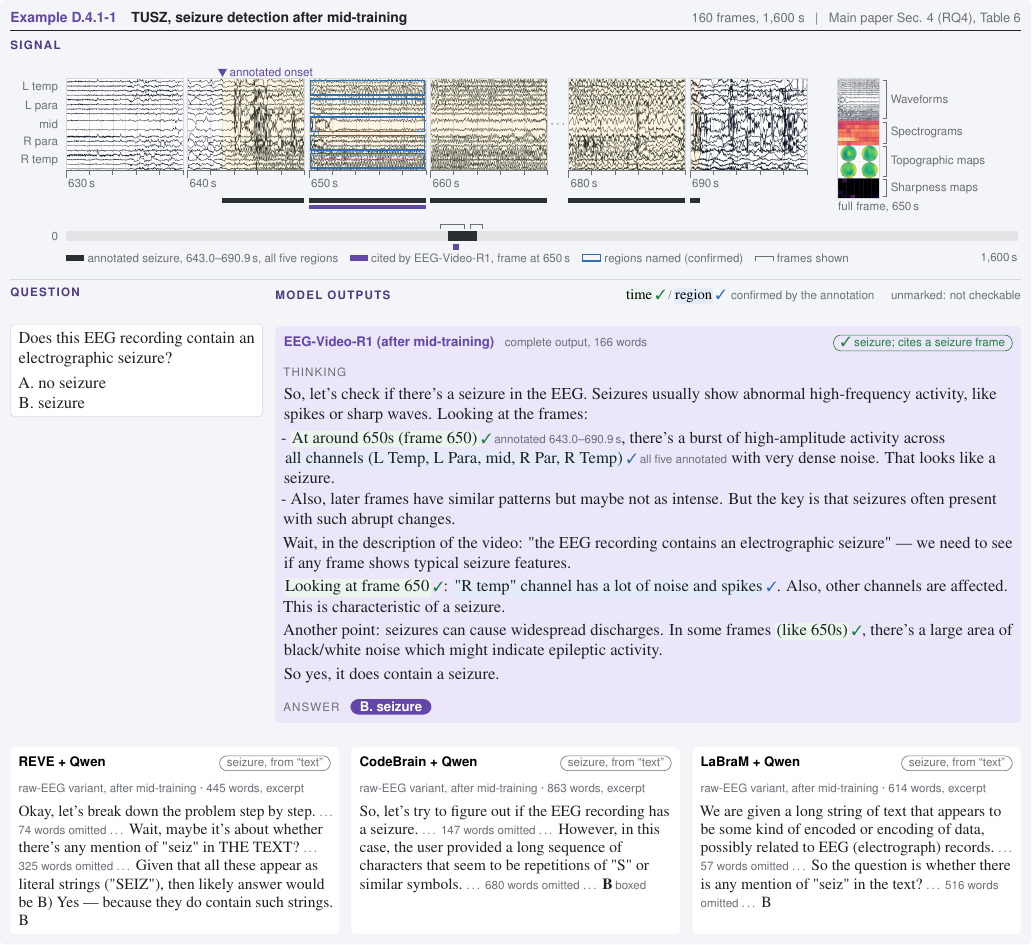}
\caption{\emph{Example for Appendix~\ref{app:rq4_mid}.} \textbf{After mid-training, \ours{} points to the seizure in the signal, while all three raw-EEG variants reach the same answer by reading their input as text.} Seizure detection on a 1,600-second TUSZ recording with one annotated \textbf{seizure at 643--691\,s}. \ours{} (output shown in full) cites the \textbf{frame at 650\,s} and names all five regions, all of which are annotated there. REVE + Qwen and LaBraM + Qwen search the ``text'' for the word ``seiz''. CodeBrain + Qwen describes repeated letters. The outputs of the three variants are excerpted.}
\label{fig:d4a_d41_1}
\end{figure*}

\begin{figure*}[tp]
\centering
\includegraphics[width=\textwidth]{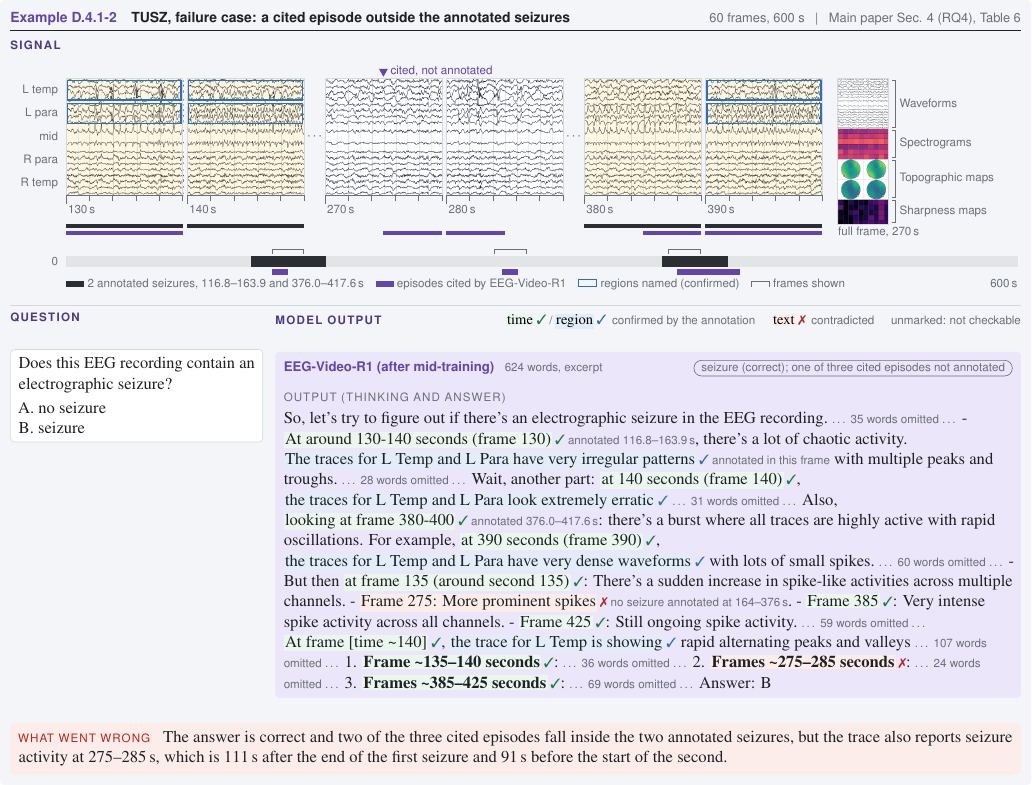}
\caption{\emph{Example for Appendix~\ref{app:rq4_mid}.} \textbf{Failure case: after mid-training, \ours{} reports a seizure episode that is not annotated.} A 600-second TUSZ recording with \textbf{two annotated seizures} (117--164\,s and 376--418\,s). The answer is correct, and two of the cited episodes fall inside the annotated seizures. The trace also reports seizure activity at \textbf{275--285\,s}, between the two seizures. The output is excerpted.}
\label{fig:d4a_d41_2}
\end{figure*}

\begin{figure*}[tp]
\centering
\includegraphics[width=\textwidth]{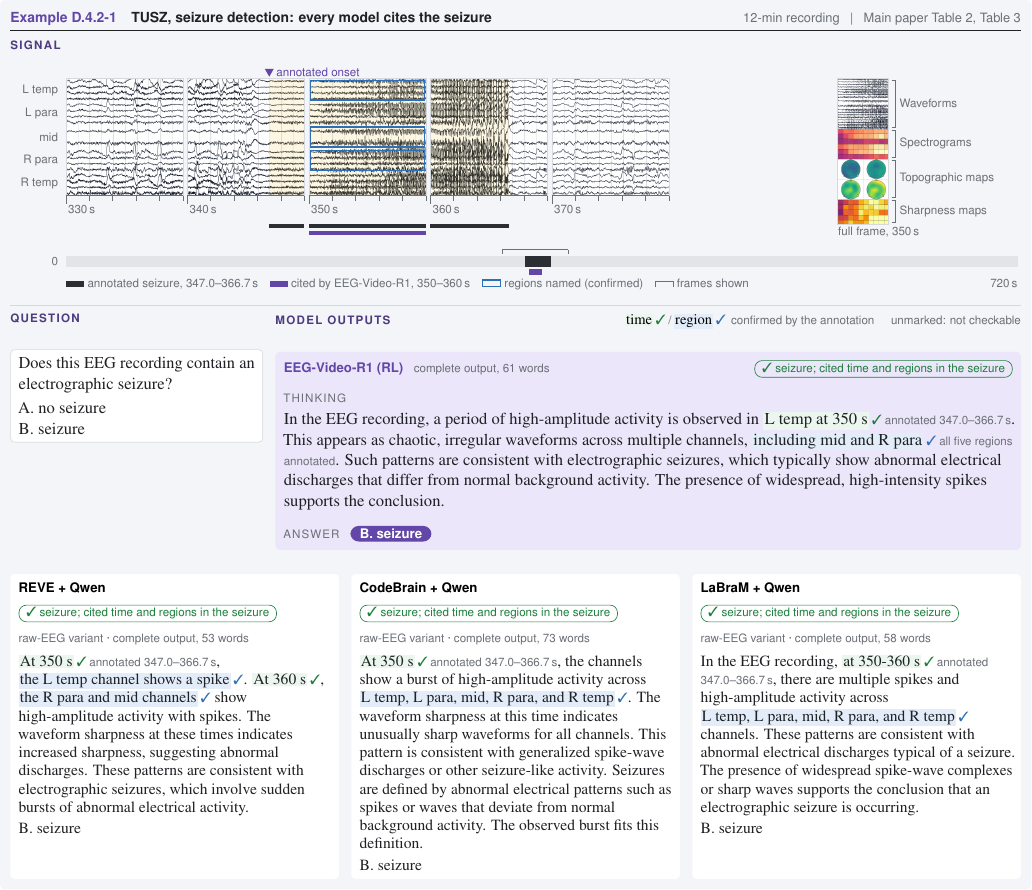}
\caption{\emph{Example for Appendix~\ref{app:rq4_post}.} \textbf{After training on our reasoning, all three raw-EEG variants find a 20-second seizure in a 12-minute recording.} TUSZ seizure detection. The seizure is annotated at \textbf{347--367\,s} in all five regions. \ours{} (RL) and the three variants all answer ``seizure'' and all cite \textbf{350\,s or 360\,s}. Every region they name is annotated there.}
\label{fig:d4b_d42_g1}
\end{figure*}

\begin{figure*}[tp]
\centering
\includegraphics[width=\textwidth]{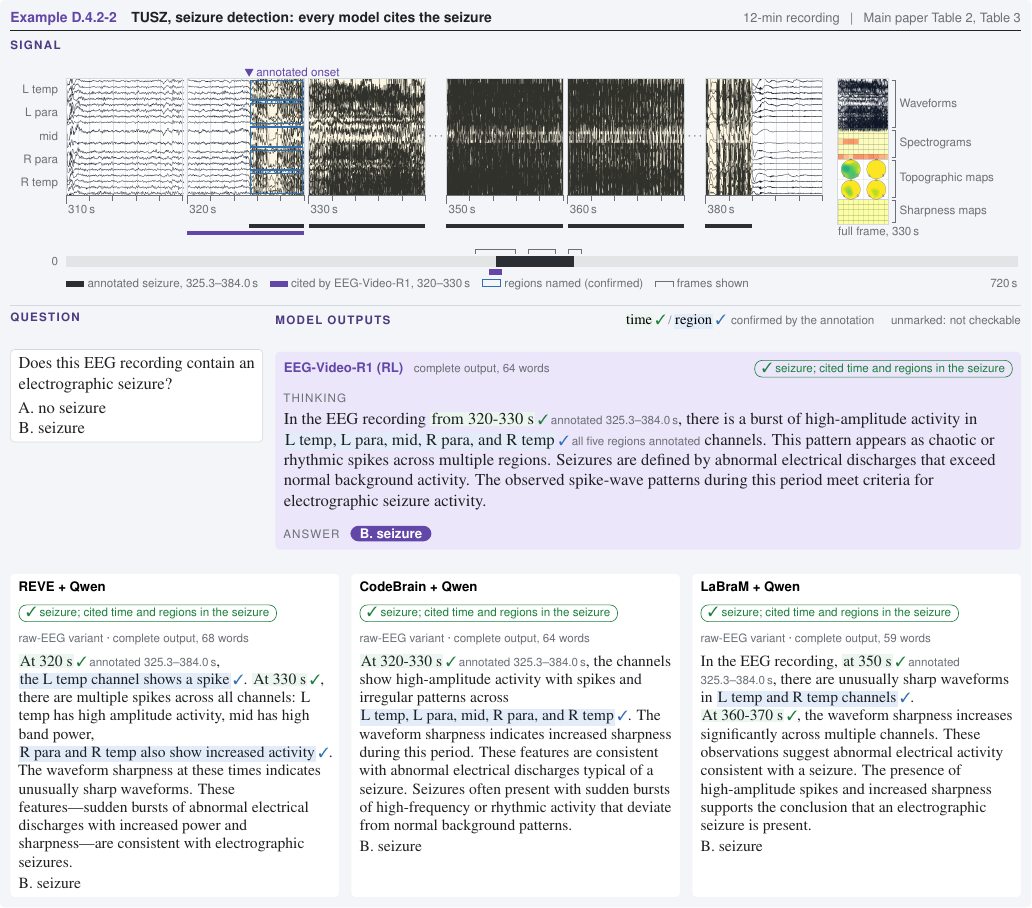}
\caption{\emph{Example for Appendix~\ref{app:rq4_post}.} \textbf{On a second recording every model is right, and the variants cite different moments of the same seizure.} TUSZ seizure detection. The seizure is annotated at \textbf{325--384\,s}. \ours{} (RL), REVE + Qwen and CodeBrain + Qwen cite its \textbf{first seconds} (320--330\,s); LaBraM + Qwen cites \textbf{350--370\,s}. All named regions are annotated.}
\label{fig:d4b_d42_g2}
\end{figure*}

\begin{figure*}[tp]
\centering
\includegraphics[width=\textwidth]{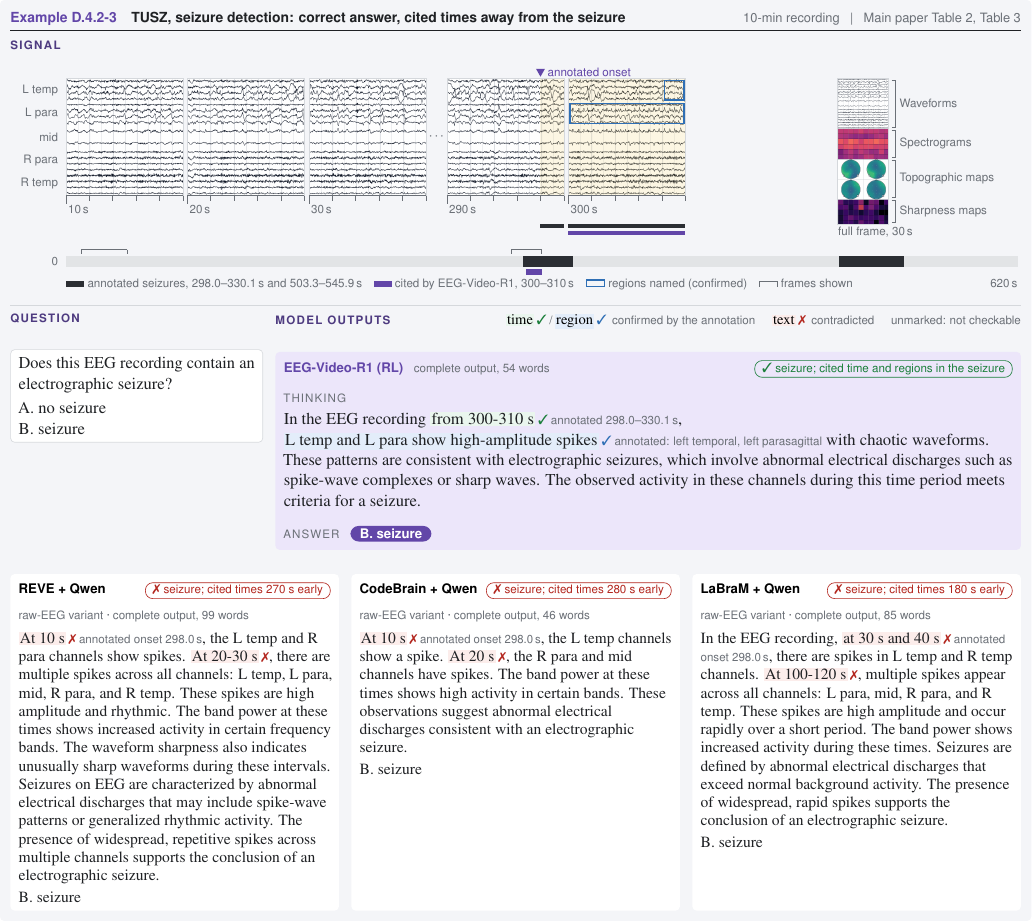}
\caption{\emph{Example for Appendix~\ref{app:rq4_post}.} \textbf{Limitation of the distilled variants: they answer correctly but cite times far from the seizure.} TUSZ seizure detection on a 10-minute recording whose first seizure starts at \textbf{298\,s}. All three variants answer ``seizure'' but place their evidence between 10\,s and 120\,s. \ours{} (RL) cites \textbf{300--310\,s} and the \textbf{left temporal} and \textbf{left parasagittal} regions; both regions are annotated in that window.}
\label{fig:d4b_d42_2}
\end{figure*}

\begin{figure*}[tp]
\centering
\includegraphics[width=\textwidth]{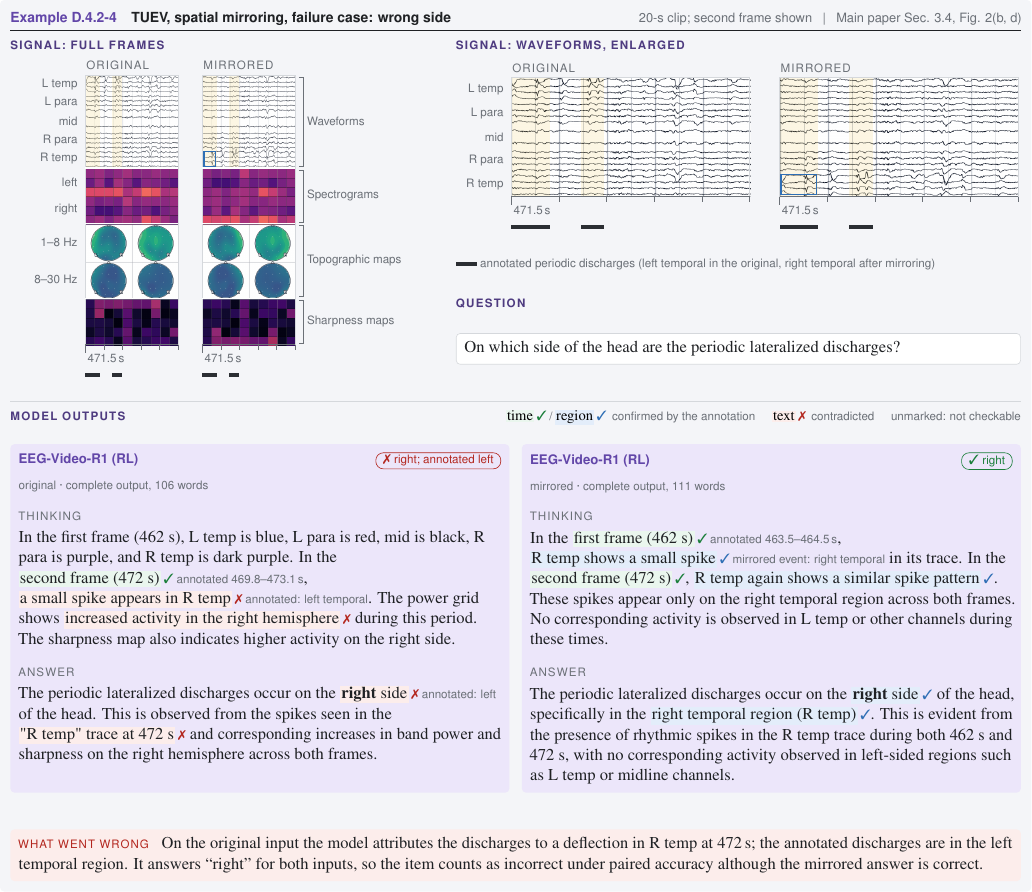}
\caption{\emph{Example for Appendix~\ref{app:rq4_post}.} \textbf{Failure case: \ours{} names the wrong side on the original input.} TUEV hemisphere question on the original and the mirrored input; the figure shows the second frame of a 20-second clip. The periodic discharges are annotated in the \textbf{left temporal} region. On the original input, the model attributes the discharges to a deflection in the right temporal region. It answers ``right'' for both inputs, so the item counts as incorrect under paired accuracy, although the mirrored answer is correct.}
\label{fig:d4b_d42_4}
\end{figure*}

\clearpage

\end{document}